%% file: tmlr_submission.tex
\documentclass[10pt]{article} %
\usepackage[accepted]{tmlr}

\input{math_commands.tex}

\usepackage{url}

\usepackage{graphicx}
\usepackage{tikz}
\usepackage{booktabs}
\usepackage{multirow}
\usepackage{subcaption}
\usepackage{amssymb}
\usepackage[pagebackref,breaklinks,colorlinks]{hyperref}

\definecolor{red1}{rgb}{0.8431 0.1882 0.1529}
\newcommand{\red}[1]{{\color{red1}#1}}

\RequirePackage{xspace}
\makeatletter
\DeclareRobustCommand\onedot{\futurelet\@let@token\@onedot}
\def\@onedot{\ifx\@let@token.\else.\null\fi\xspace}

\def\eg{\emph{e.g}\onedot} 

\def\ie{\emph{i.e}\onedot}

\makeatother

\title{Bigger is not Always Better: \\Scaling Properties of Latent Diffusion Models}

\author{\name Kangfu Mei\thanks{This work was done during an internship at Google.} \email kmei1@jhu.edu \\ \addr Johns Hopkins University \\
\AND
Zhengzhong Tu\thanks{This work was done while at Google.} \email tzz@tamu.edu \\ \addr Texas A\&M University
\AND
Mauricio Delbracio \email mdelbra@google.com \\ \addr Google
\AND
Hossein Talebi \email 
htalebi@google.com \\ \addr Google
\AND
Vishal M. Patel \email vpatel36@jhu.edu \\ \addr Johns Hopkins University
\AND
Peyman Milanfar \email  milanfar@google.com \\ 
\addr Google
}

\def\month{12}  %
\def\year{2024} %
\def\openreview{\url{https://openreview.net/forum?id=0u7pWfjri5}} %

\newcommand{\imageWithNote}[3]{%
  \begin{tikzpicture}
    \node[anchor=south west, inner sep=0] (image) at (0,0) {\includegraphics[width=#1]{#3}};
    \node[anchor=south east, inner sep=2pt, fill=white] at (current bounding box.south east) {\parbox{.5cm}{#2}};
  \end{tikzpicture}%
}

\newcommand{\imageWithMoreNote}[3]{%
  \begin{tikzpicture}
    \node[anchor=south west, inner sep=0] (image) at (0,0) {\includegraphics[width=#1]{#3}};
    \node[anchor=south east, inner sep=2pt, fill=white] at (current bounding box.south east) {\parbox{1.4cm}{#2}};
  \end{tikzpicture}%
}

\begin{document}

\maketitle

\begin{abstract}
We study the scaling properties of latent diffusion models (LDMs) with an emphasis on their sampling efficiency.
While improved network architecture and inference algorithms have shown to effectively boost sampling efficiency of diffusion models, the role of model size---a critical determinant of sampling efficiency---has not been thoroughly examined.
Through empirical analysis of established text-to-image diffusion models, we conduct an in-depth investigation into how model size influences sampling efficiency across varying sampling steps.
Our findings unveil a surprising trend: when operating under a given inference budget, smaller models frequently outperform their larger equivalents in generating high-quality results.
Moreover, we extend our study to demonstrate the generalizability of the these findings by applying various diffusion samplers, exploring diverse downstream tasks, evaluating post-distilled models, as well as comparing performance relative to training compute.
These findings open up new pathways for the development of LDM scaling strategies which can be employed to enhance generative capabilities within limited inference budgets. 
\end{abstract}

\section{Introduction}
\label{sec:intro}

Latent diffusion models (LDMs)~\citep{rombach2022high}, and diffusion models in general, trained on large-scale, high-quality data~\citep{lin2014microsoft, schuhmann2022laion} have emerged as a powerful and robust framework for generating impressive results in a variety of tasks, including image synthesis and editing~\citep{rombach2022high, podell2023sdxl, delbracio2023inversion, ren2023multiscale, qi2023tip}, video creation~\citep{mei2023vidm, mei2023t1, wu2023tune, singer2022make}, audio production~\citep{liu2023audioldm}, and 3D synthesis~\citep{lin2023magic3d, liu2023zero}.
Despite their versatility, the major barrier against wide deployment in real-world applications~\citep{du2023exploring, choi2023squeezing} comes from their low \emph{sampling efficiency}.
The essence of this challenge lies in the inherent reliance of LDMs on multi-step sampling~\citep{song2020score, ho2020denoising} to produce high-quality outputs, where the total cost of sampling is the product of sampling steps and the cost of each step.
Specifically, the go-to approach involves using the 50-step DDIM sampling~\citep{song2020denoising, rombach2022high}, a process that, despite ensuring output quality, still requires a relatively long latency for completion on modern mobile devices with post-quantization.
In contrast to single shot generative models (e.g., generative-adversarial networks (GANs)~\citep{goodfellow2020generative}) which bypass the need for iterative refinement~\citep{goodfellow2020generative,karras2019style}, the operational latency of LDMs calls for a pressing need for efficiency optimization to further facilitate their practical applications.

Recent advancements in this field~\citep{li2023snapfusion, zhao2023mobilediffusion, peebles2023scalable, kim2023bk, kim2023architectural, choi2023squeezing} have primarily focused on developing faster network architectures with comparable model size to reduce the inference time per step, along with innovations in improving sampling algorithms that allow for using less sampling steps~\citep{song2020denoising, dockhorn2022genie, karras2022elucidating, lu2022dpm, liu2023instaflow, xu2023ufogen}.
Further progress has been made through diffusion-distillation techniques~\citep{luhman2021knowledge, salimans2022progressive, song2023consistency, sauer2023adversarial, gu2023boot, mei2023conditional}, which simplifies the process by learning multi-step sampling results in a single forward pass, and then broadcasts this single-step prediction multiple times.
These distillation techniques leverage the redundant learning capability in LDMs, enabling the distilled models to assimilate additional distillation knowledge.
Despite these efforts being made to improve diffusion models, the sampling efficiency of smaller, less redundant models has not received adequate attention.
A significant barrier to this area of research is the scarcity of available modern accelerator clusters~\citep{jouppi2023tpu}, as training high-quality text-to-image (T2I) LDMs from scratch is both time-consuming and expensive---often requiring several weeks and hundreds of thousands of dollars.

In this paper, we empirically investigate the scaling properties of LDMs, with a particular focus on understanding how their scaling properties impact the sampling efficiency across various model sizes.
We trained a suite of 12 text-to-image LDMs from scratch, ranging from 39 million to 5 billion parameters, under a constrained budget. Example results are depicted in Fig.~\ref{fig:t2i_results}.
All models were trained on TPUv5 using internal data sources with about 600 million aesthetically-filtered text-to-image pairs.
Our study reveals that there exist a scaling trend within LDMs, notably that smaller models may have the capability to surpass larger models under an equivalent sampling budget.
Furthermore, we investigate how the size of pre-trained text-to-image LDMs affects their sampling efficiency across diverse downstream tasks, such as real-world super-resolution~\citep{saharia2022image, sahak2023denoising} and subject-driven text-to-image synthesis (i.e., Dreambooth)~\citep{ruiz2023dreambooth}.

\subsection{Summary}
Our key findings for scaling latent diffusion models in text-to-image generation and various downstream tasks are as follows:

\vspace{.5em}
\noindent \textbf{Pretraining performance scales with training compute.}
We demonstrate a clear link between compute resources and LDM performance by scaling models from 39 million to 5 billion parameters.  This suggests potential for further improvement with increased scaling. See Section~\ref{sec:scalingt2i} for details.\vspace{.5em}

\vspace{.5em}
\noindent \textbf{Downstream performance scales with pretraining.}
We demonstrate a strong correlation between pretraining performance and success in downstream tasks. Smaller models, even with extra training, cannot fully bridge the gap created by the pretraining quality of larger models. 
This is explored in detail in Section~\ref{sec:scalingsr}.

\vspace{.5em}
\noindent \textbf{Smaller models sample more efficient.} 
Smaller models initially outperform larger models in image quality for a given sampling budget, but larger models surpass them in detail generation when computational constraints are relaxed. %
This is further elaborated in Section~\ref{sec:optimalparams} and Section~\ref{sec:optimalmodelsizes}.

\vspace{.5em}
\noindent \textbf{Sampler does not change the scaling efficiency.}
Smaller models consistently demonstrate superior sampling efficiency, regardless of the diffusion sampler used. This holds true for deterministic DDIM~\citep{song2020denoising}, stochastic DDPM~\citep{ho2020denoising}, and higher-order DPM-Solver++~\citep{lu2022dpm2}. For more details, see Section~\ref{sec:samplerscaling}.

\vspace{.5em}
\noindent \textbf{Smaller models sample more efficient on the downstream tasks with fewer steps.}
The advantage of smaller models in terms of sampling efficiency extends to the downstream tasks when using less than 20 sampling steps.
This is further elaborated in Section~\ref{sec:scalingsamplingsr}.

\vspace{.5em}
\noindent \textbf{Diffusion distillation does not change scaling trends.}
Even with diffusion distillation, smaller models maintain competitive performance against larger distilled models when sampling budgets are constrained. This suggests distillation does not fundamentally alter scaling trends. See Section~\ref{sec:scalingdistill} for in-depth analysis.

\section{Related Work}
\paragraph{Scaling laws.}
Recent Large Language Models (LLMs) including GPT~\citep{brown2020language},  PaLM~\citep{anil2023palm}, and LLaMa~\citep{touvron2023llama} have dominated language generative modeling tasks.
The foundational works~\citep{kaplan2020scaling, brown2020language, hoffmann2022training} for investigating their scaling behavior have shown the capability of predicting the performance from the model size. They also investigated the factors that affect the scaling properties of language models, including training compute, dataset size and quality, learning rate schedule, etc.
Those experimental clues have effectively guided the later language model development, which have led to the emergence of several parameter-efficient LLMs~\citep{hoffmann2022training, touvron2023llama, zhou2023brainformers, alabdulmohsin2024getting}.
However, scaling generative text-to-image models are relatively unexplored, and existing efforts have only investigated the scaling properties on small datasets or small models, like scaling UNet~\citep{nichol2021improved} to 270 million parameters and DiT~\citep{peebles2023scalable} on ImageNet (14 million), or less-efficient autoregressive models~\citep{chen2020generative}.
Different from these attempts, our work investigates the scaling properties by scaling down the efficient and capable diffusion models, \ie LDMs~\citep{rombach2022high}, on internal data sources that have about 600 million aesthetics-filtered text-to-image pairs for featuring the sampling efficiency of scaled LDMs.
We also scale LDMs on various scenarios such as finetuning LDMs on downstream tasks~\citep{wang2021real,ruiz2023dreambooth} and distilling  LDMs~\citep{mei2023conditional} for faster sampling to demonstrate the generalizability of the scaled sampling-efficiency.

\paragraph{Efficient diffusion models.}
Nichol et al.~\citep{nichol2021improved} show that the generative performance of diffusion models improves as the model size increases.
Based on this preliminary observation, the model size of widely used LDMs, \eg, Stable Diffusion~\citep{rombach2022high}, has been empirically increased to billions of parameters~\citep{ramesh2022hierarchical, podell2023sdxl}.
However, such a large model makes it impossible to fit into the common inference budget of practical scenarios.
Recent work on improving the sampling efficiency focus on improving network architectures~\citep{li2023snapfusion, zhao2023mobilediffusion, peebles2023scalable, kim2023bk, kim2023architectural, choi2023squeezing, mei2024latent} or the sampling procedures~\citep{song2020denoising, dockhorn2022genie, karras2022elucidating, lu2022dpm, liu2023instaflow, xu2023ufogen, mei2025improving}.
We explore sampling efficiency by training smaller, more compact LDMs. Our analysis involves scaling down the model size, training from scratch, and comparing performance at equivalent inference cost.

\begin{figure*}[htbp]
    \centering
    \begin{subfigure}[b]{0.32\textwidth}
    \centering
    \includegraphics[width=\textwidth]{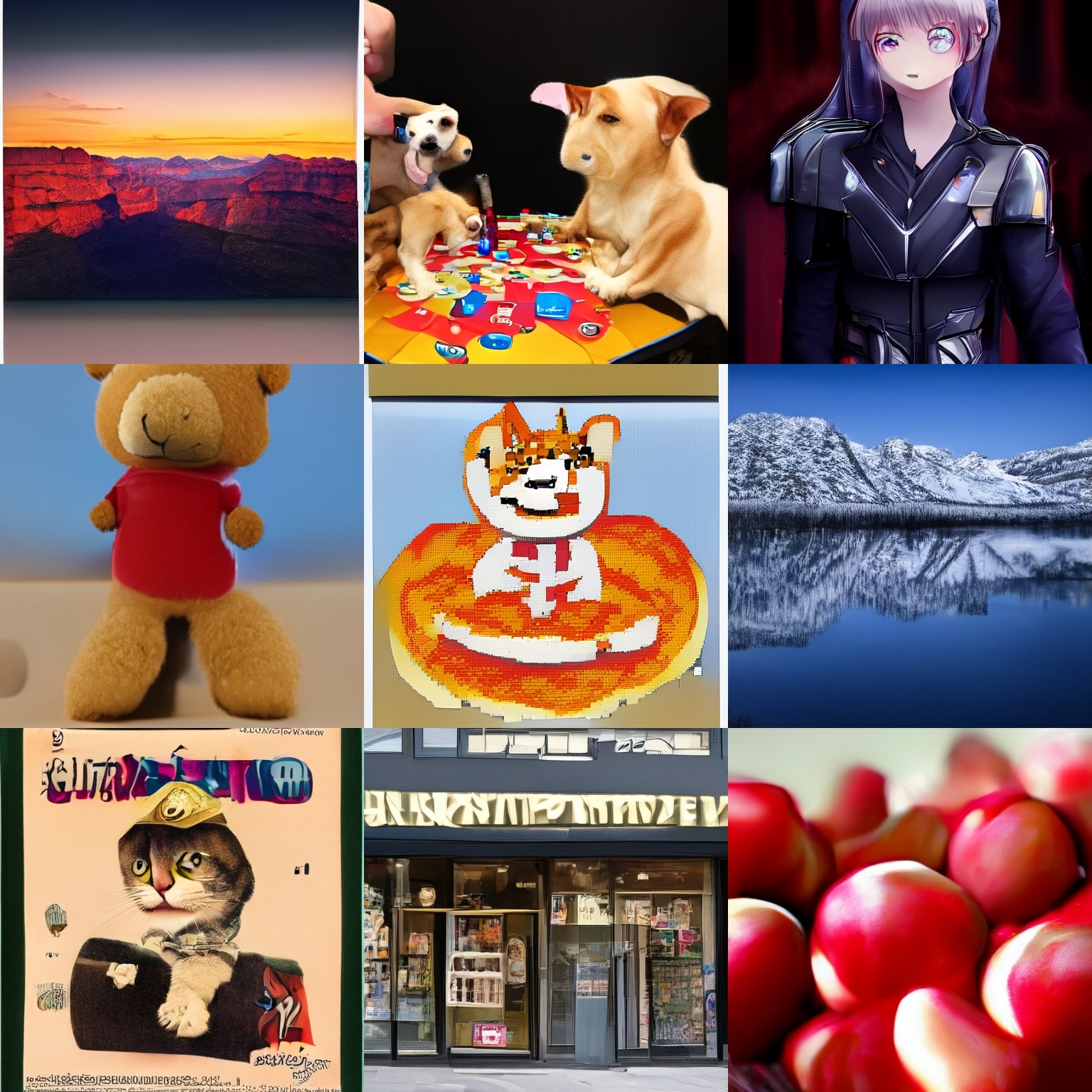}
    \caption{\texttt{39M} model}
    \end{subfigure}
    \begin{subfigure}[b]{0.32\textwidth}
    \centering
    \includegraphics[width=\textwidth]{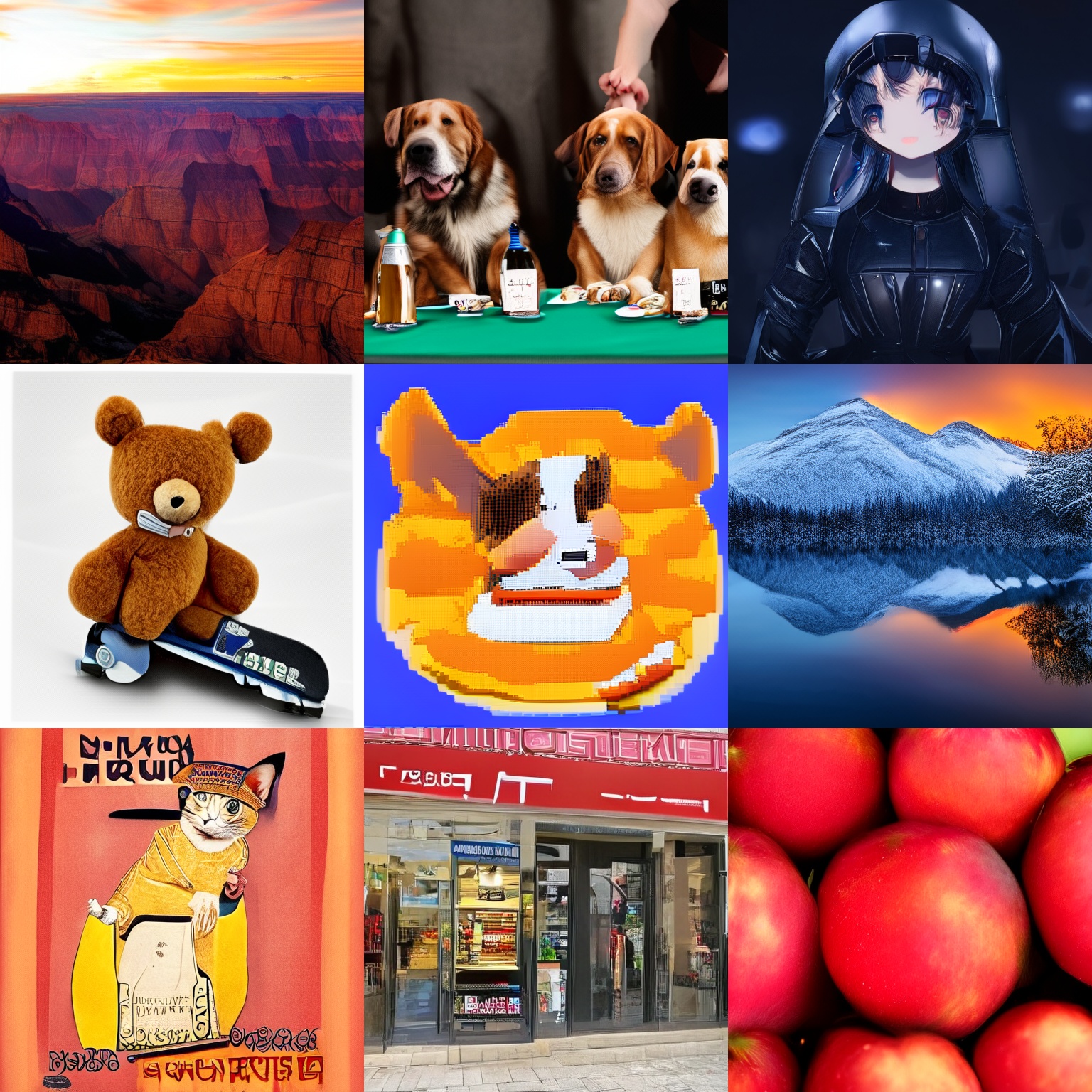}
    \caption{\texttt{83M} model}
    \end{subfigure}
    \begin{subfigure}[b]{0.32\textwidth}
    \centering
    \includegraphics[width=\textwidth]{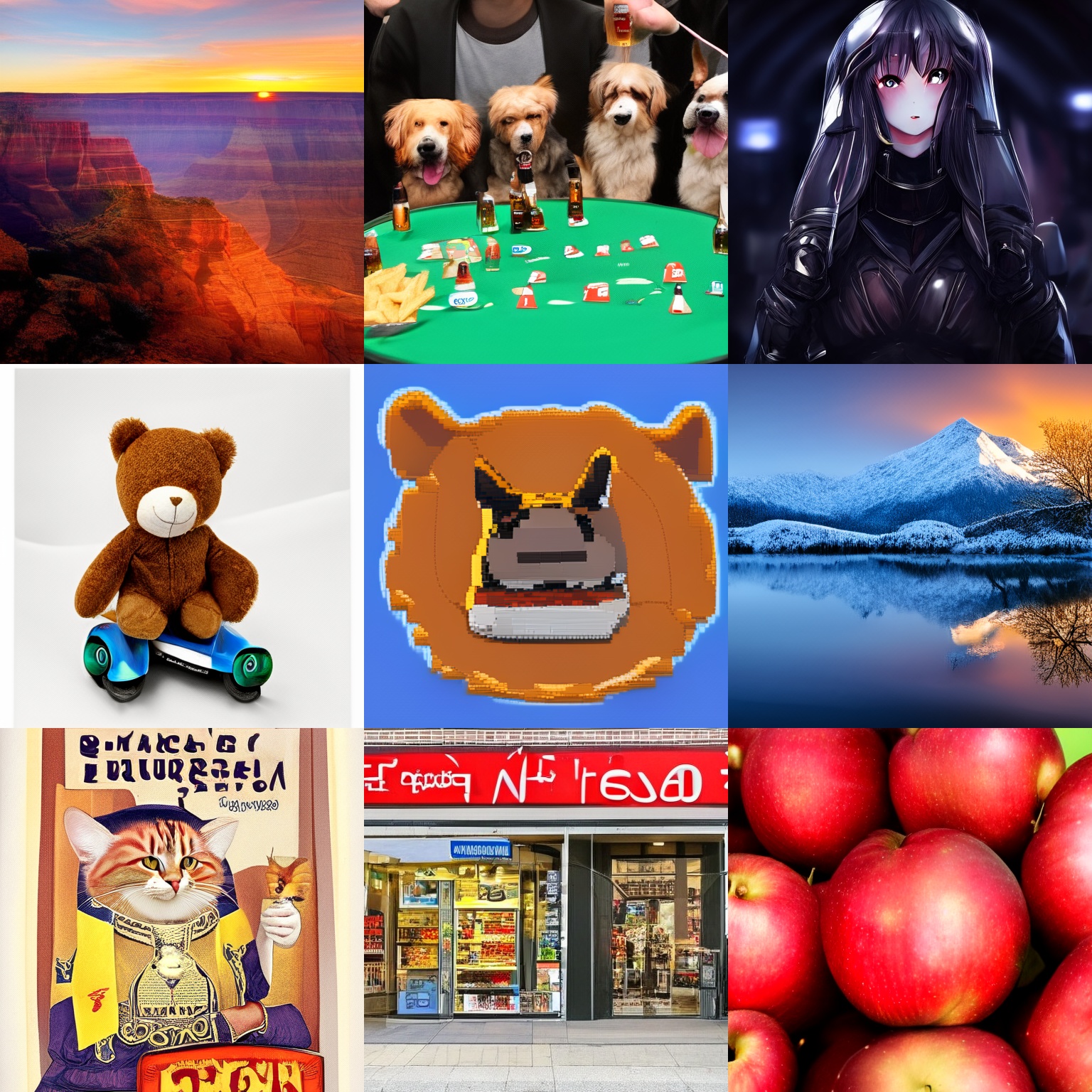}
    \caption{\texttt{145M} model}
    \end{subfigure}
    \hfill

    \begin{subfigure}[b]{0.32\textwidth}
    \centering
    \includegraphics[width=\textwidth]{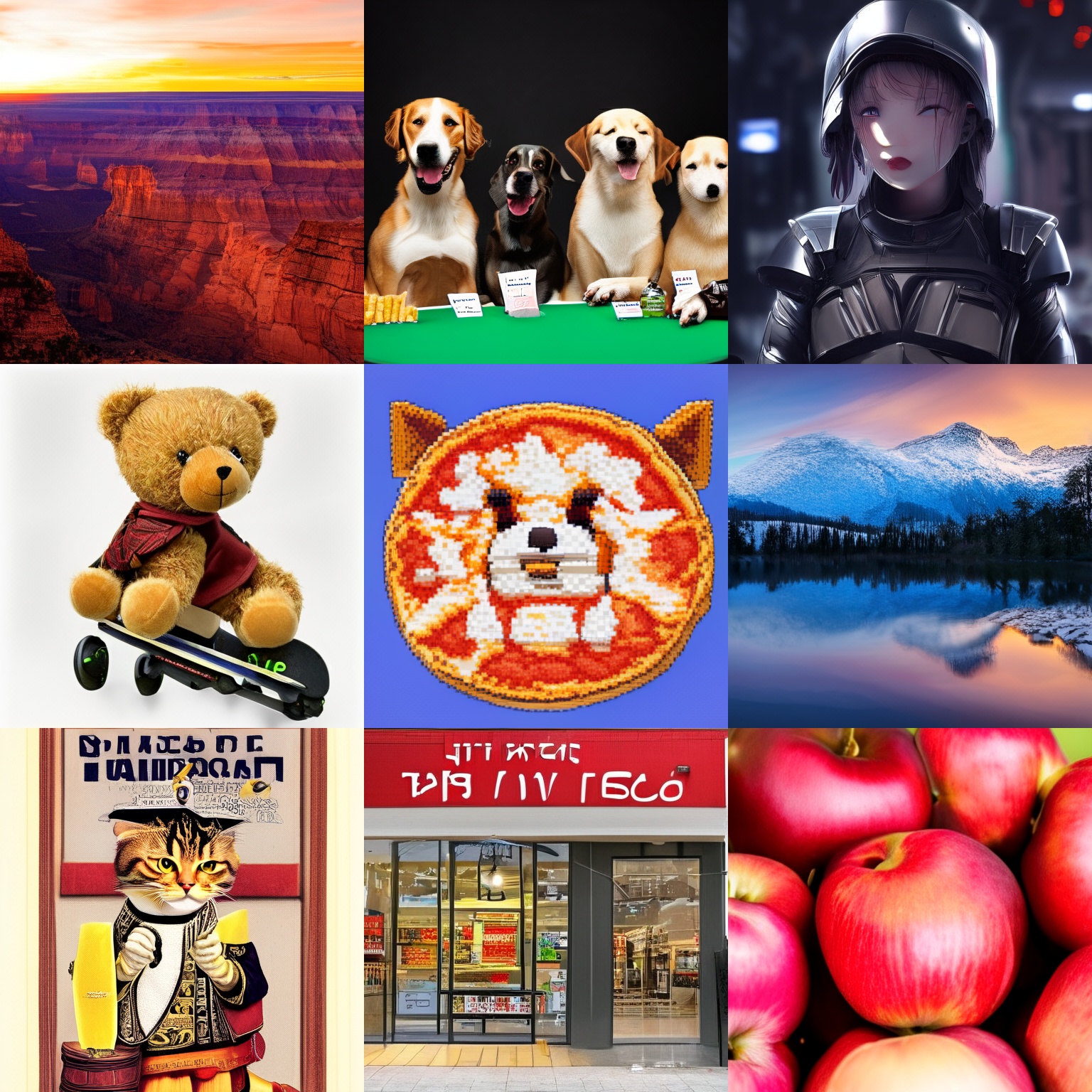}
    \caption{\texttt{223M} model}
    \end{subfigure}
    \begin{subfigure}[b]{0.32\textwidth}
    \centering
    \includegraphics[width=\textwidth]{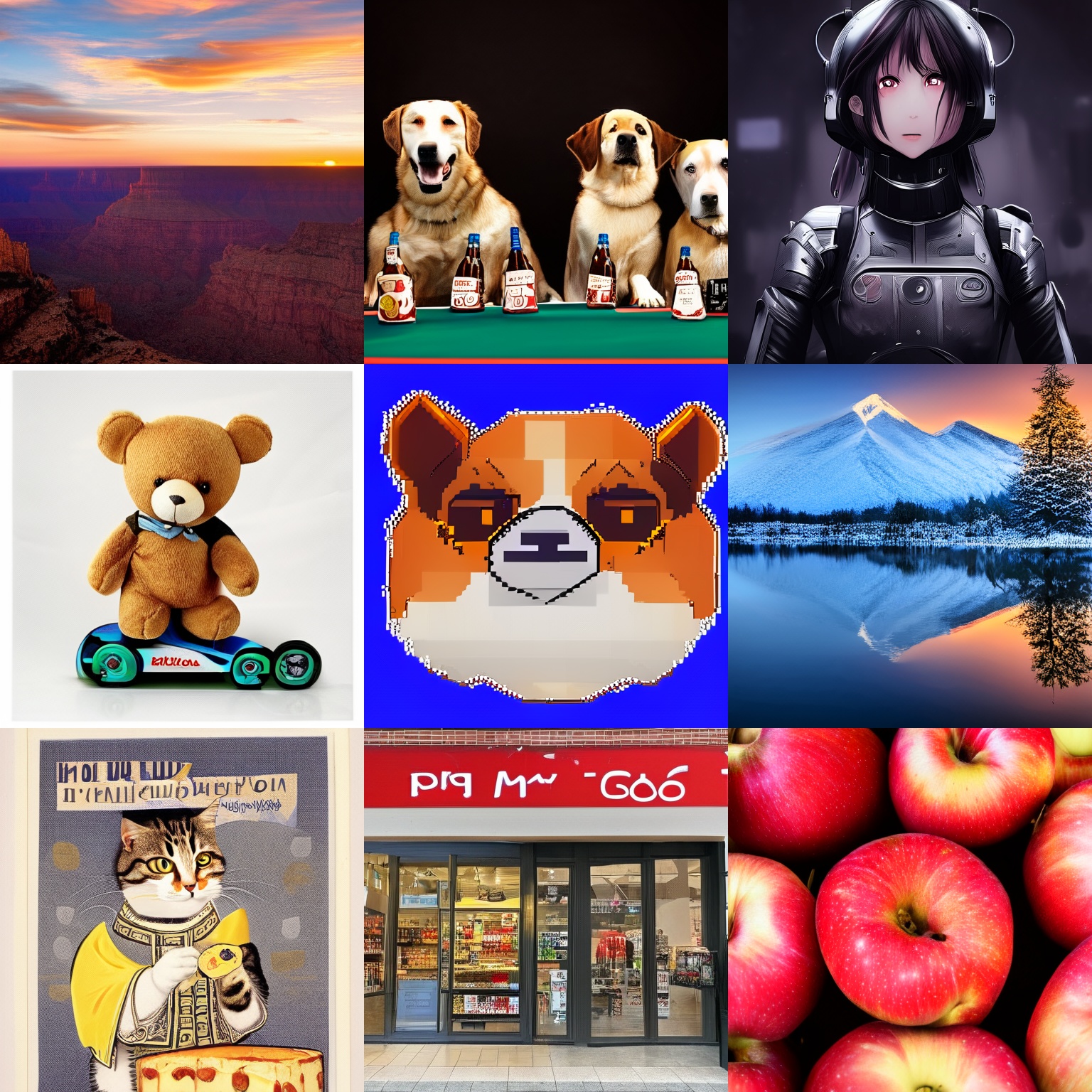}
    \caption{\texttt{318M} model}
    \end{subfigure}
    \begin{subfigure}[b]{0.32\textwidth}
    \centering
    \includegraphics[width=\textwidth]{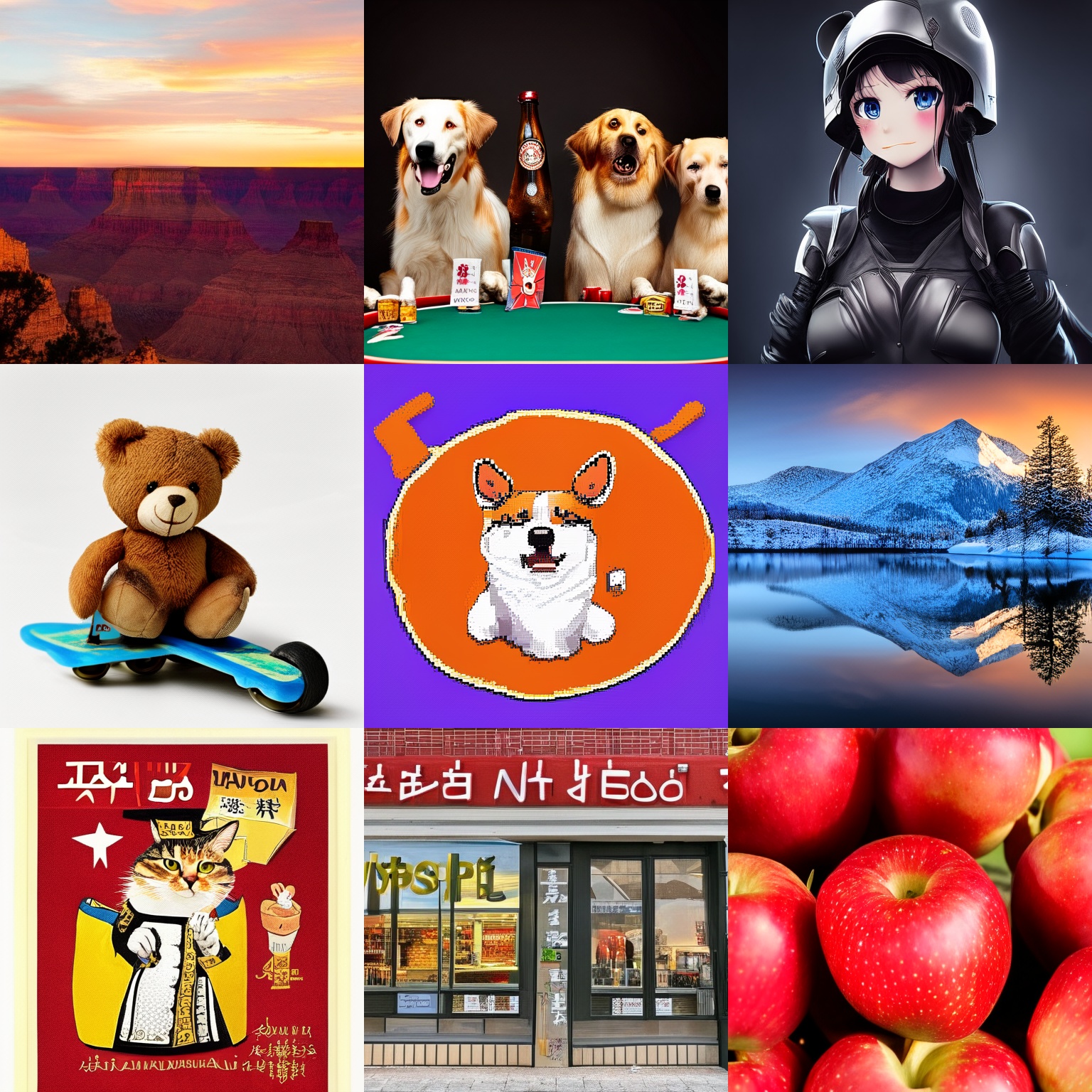}
    \caption{\texttt{430M} model}
    \end{subfigure}
    \hfill
    
    \begin{subfigure}[b]{0.32\textwidth}
    \centering
    \includegraphics[width=\textwidth]{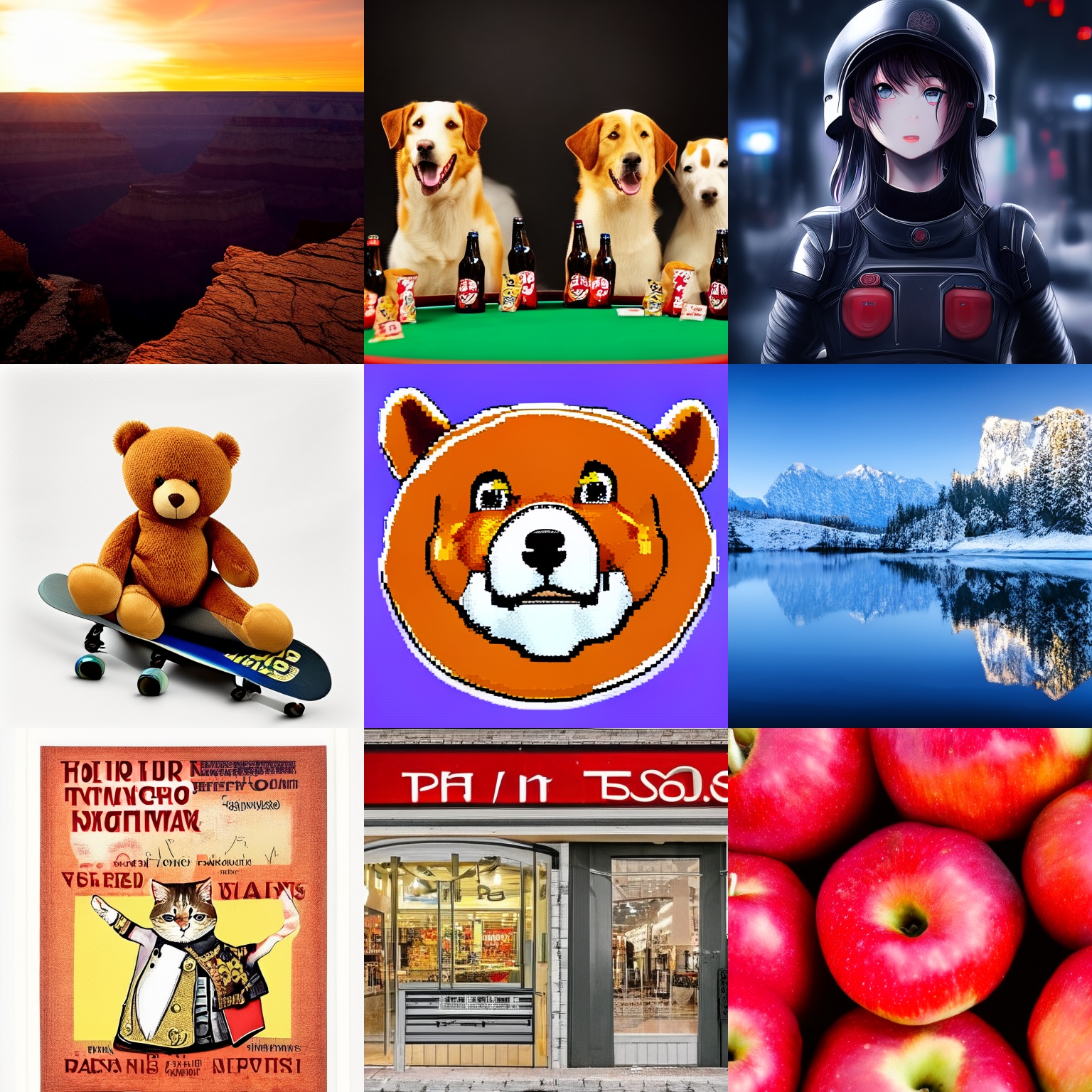}
    \caption{\texttt{558M} model}
    \end{subfigure}
    \begin{subfigure}[b]{0.32\textwidth}
    \centering
    \includegraphics[width=\textwidth]{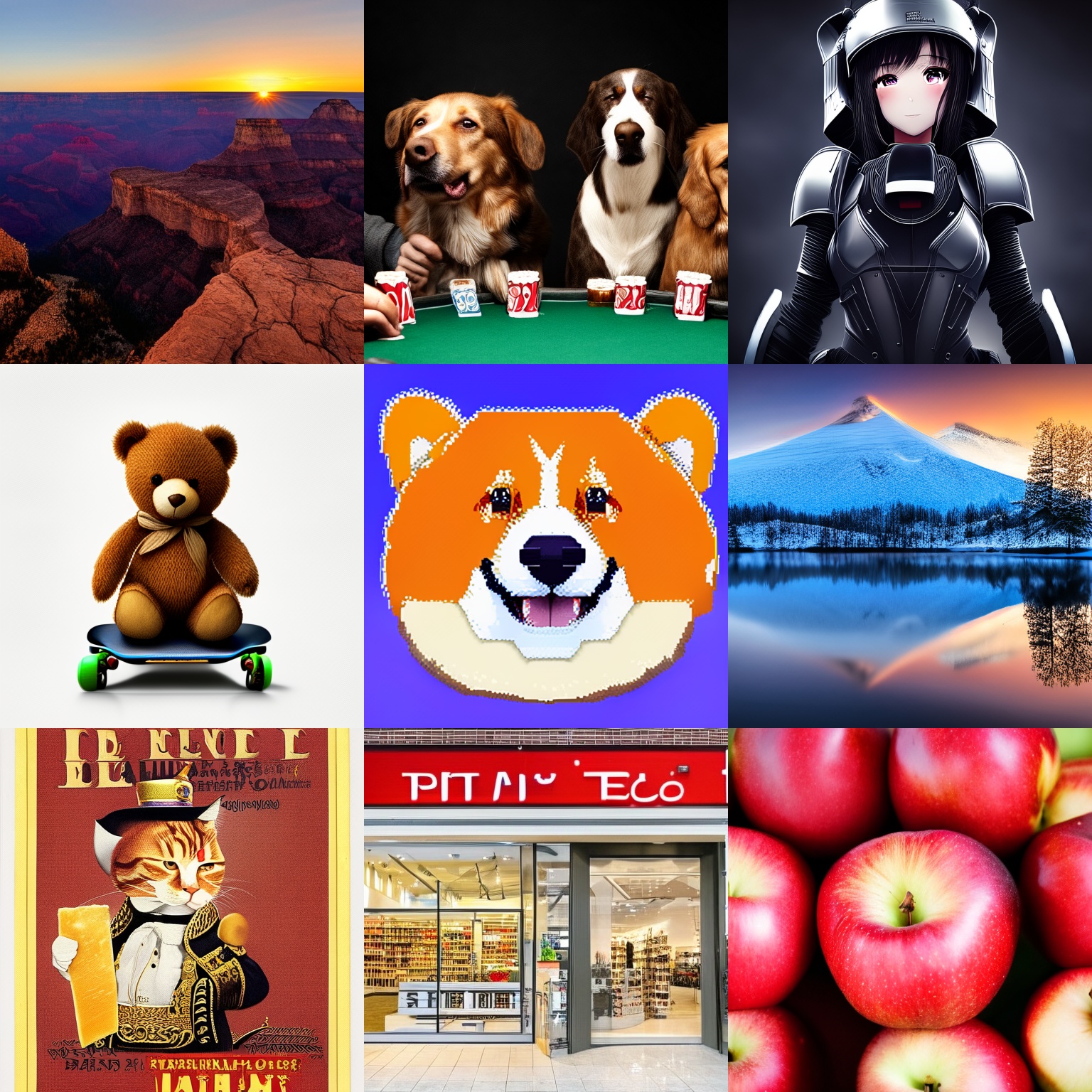}
    \caption{\texttt{704M} model}
    \end{subfigure}
    \begin{subfigure}[b]{0.32\textwidth}
    \centering
    \includegraphics[width=\textwidth]{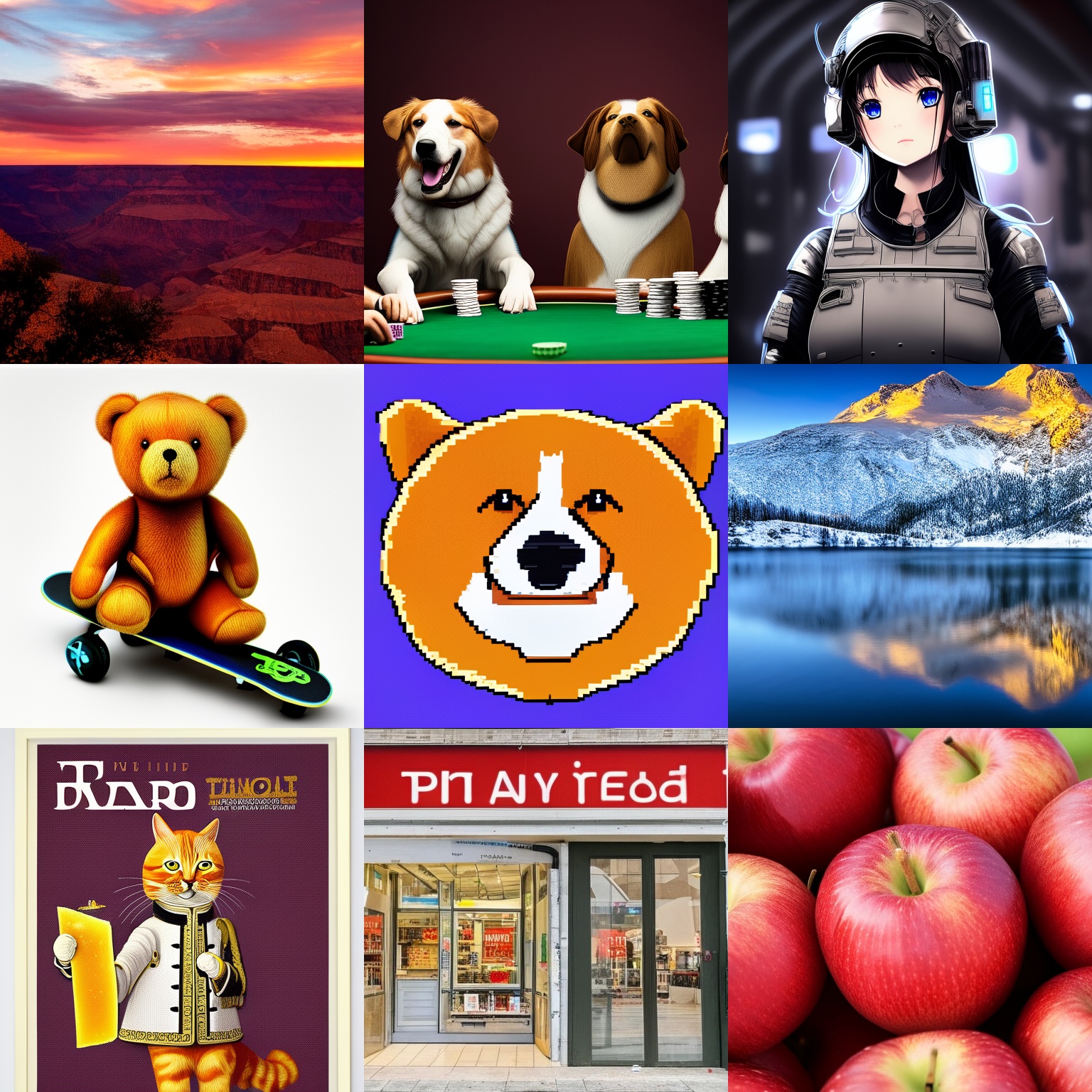}
    \caption{\texttt{2B} model}
    \end{subfigure}

\caption{
Text-to-image results from our scaled LDMs (\texttt{39M} - \texttt{2B}), highlighting the improvement in visual quality with increased model size (note: 39M model is the exception). All images generated using 50-step DDIM sampling and CFG rate of 7.5.
We use representative prompts from PartiPrompts~\cite{yu2022scaling}, including \emph{``a professional photo of a sunset behind the grand canyon.''}, \emph{``Dogs sitting around a poker table with beer bottles and chips. Their hands are holding cards.''}, \emph{`Portrait of anime girl in mechanic armor in night Tokyo.''}, \emph{``a teddy bear on a skateboard.''}, \emph{``a pixel art corgi pizza.''}, \emph{``Snow mountain and tree reflection in the lake.''}, \emph{``a propaganda poster depicting a cat dressed as french emperor napoleon holding a piece of cheese.''}, \emph{``a store front that has the word ‘LDMs’ written on it.''}, and \emph{``ten red apples.''}.
\emph{Check our supplement for additional visual comparisons.}}
\label{fig:t2i_results}
\end{figure*}

\begin{table}[t]
    \centering
    \footnotesize
    \setlength{\tabcolsep}{4pt}
    \resizebox{\linewidth}{!}{
    \begin{tabular}{cccccccccccc}
    \toprule
    \textbf{Params} & \texttt{39M} & \texttt{83M} & \texttt{145M} & \texttt{223M} & \texttt{318M} & \texttt{430M} & \texttt{558M} & \texttt{704M} & \red{\texttt{866M}} & \texttt{2B} & \texttt{5B} \\
    \midrule
    Filters $(c)$ & 64 & 96 & 128 & 160 & 192 & 224 & 256 & 288 & \red{320} & 512 & 768 \\
    GFLOPS & 25.3 & 102.7 & 161.5 & 233.5& 318.5 & 416.6 & 527.8 & 652.0 & \red{789.3} & 1887.5& 4082.6 \\
    Norm. Cost & 0.07 & 0.13 & 0.20 & 0.30 & 0.40 & 0.53 & 0.67 & 0.83 & \red{1.00} & 2.39 & 5.17 \\
    \midrule
    FID~$\downarrow$ & 25.30 & 24.30 & 24.18 & 23.76 & 22.83 & 22.35 & 22.15 &  21.82 & \red{21.55} & 20.98 & 20.14  \\
    CLIP~$\uparrow$ & 0.305 &  0.308 & 0.310 & 0.310 & 0.311 & 0.312 & 0.312 &  0.312 & \red{0.312} &  0.312 & 0.314 \\
    \bottomrule
    \end{tabular}
    }
    \vspace{.5em}
\caption{We scale the baseline LDM (\ie, \texttt{866M} Stable Diffusion v1.5) by changing the base number of channels $c$ that controls the rest of the U-Net architecture as $[c, 2c, 4c, 4c]$ (See Fig.~\ref{fig:scaling_arch}).
GFLOPS are measured for an input latent of shape $64\times 64 \times 4$ with FP32.
We also show a normalized running cost with respect to the baseline model.
The text-to-image performance (FID and CLIP scores) for all scaled LDMs is evaluated on the COCO-2014 validation set with 30k samples, using 50-step DDIM sampling and Classifier-free Guidance (CFG) with a rate of 7.5.
It is worth noting that all the model sizes, and the training and the inference costs reported in this work only refer to the denoising UNet in the latent space, and do not include the \texttt{1.4B} text encoder and the \texttt{250M} latent encoder and decoder.
}
\label{tab:scaling_config}
\end{table}

\paragraph{Efficient non-diffusion generative models.}
Compared to diffusion models, other generative models such as, Variational Autoencoders (VAEs)~\citep{kingma2013auto, rezende2015variational, makhzani2015adversarial, vahdat2020nvae}, Generative Adversarial Networks (GANs)~\citep{goodfellow2020generative, mao2017least, karras2019style, reed2016generative, miyato2018spectral}, and Masked Models~\citep{devlin2018bert, raffel2020exploring, he2022masked, chang2022maskgit, chang2023muse}, are more efficient, as they rely less on an iterative refinement process.
Sauer et al.~\citep{sauer2023stylegan} recently scaled up StyleGAN~\citep{karras2019style} into 1 billion parameters and demonstrated the single-step GANs' effectiveness in modeling text-to-image generation.
Chang et al.~\citep{chang2023muse} scaled up masked transformer models for text-to-image generation.
These non-diffusion generative models can generate high-quality images with less inference cost, which require fewer sampling steps than diffusion models and autoregressive models, but they need more parameters, \ie, 4 billion parameters.

\begin{figure}[t]
    \centering
    \includegraphics[width=.9\linewidth]{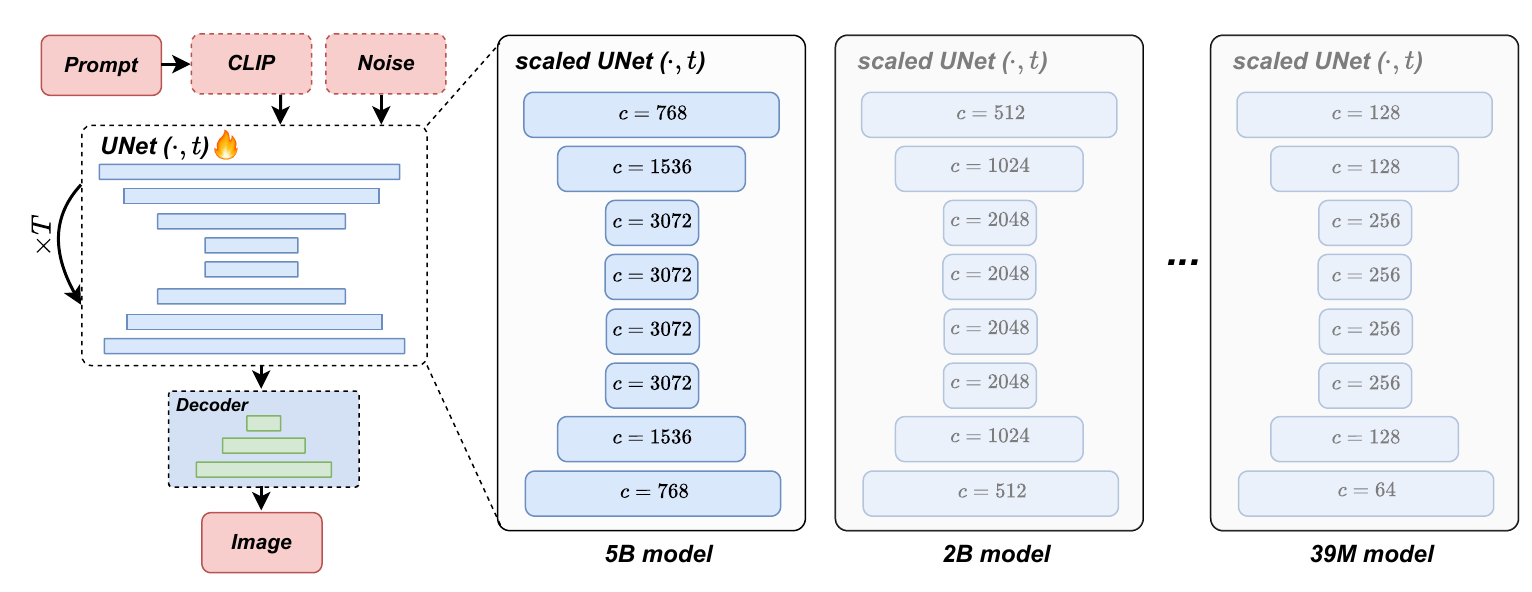}
    \caption{Our scaled latent diffusion models vary in the number of filters within the denoising U-Net.  Other modules remain consistent.  Smooth channel scaling (64 to 768) within residual blocks yields models ranging from \texttt{39M} to \texttt{5B} parameters. For downstream tasks requiring image input, we use an encoder to generate a latent code; this code is then concatenated with the noise vector in the denoising U-Net.}
    \label{fig:scaling_arch}
    \vspace{-1\baselineskip}
\end{figure}

\begin{figure}[!h]
    \centering
    \includegraphics[height=.315\linewidth]{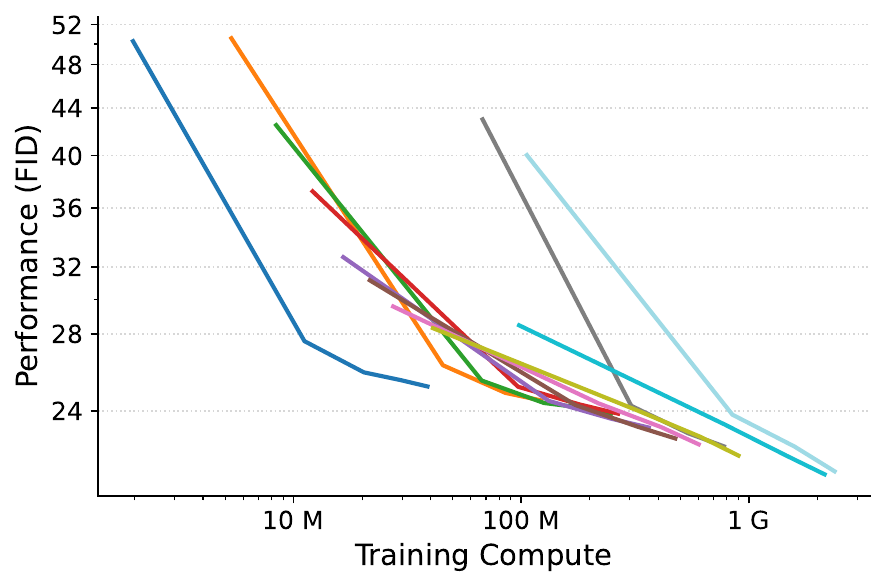}
    \includegraphics[height=.315\linewidth]{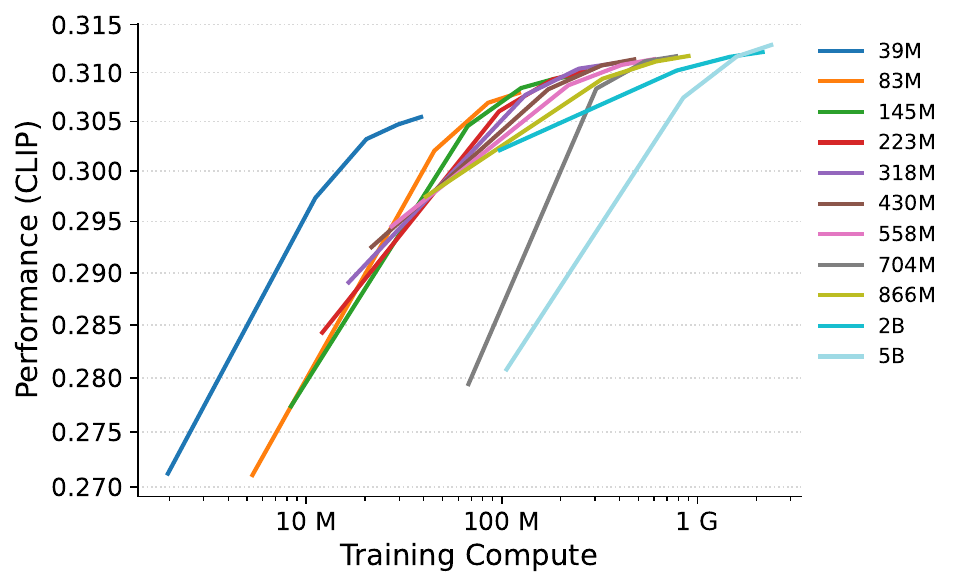}

    \vspace{-.5\baselineskip}
    \caption{In text-to-image generation using 50-step DDIM sampling and CFG rate of 7.5, we observe consistent trends across various model sizes in how quality metrics (FID and CLIP scores) relate to training compute (\ie, the total GFLOPS spend on training). 
    Under moderate training resources, training compute is the most relevant factor dominating quality.
    }
    \label{fig:t2i_compute}
    \vspace{-1\baselineskip}
\end{figure}

\section{Scaling LDMs}

We developed a family of powerful Latent Diffusion Models (LDMs) built upon the widely-used \texttt{866M} Stable Diffusion v1.5 standard~\citep{rombach2022high}\footnote{We adopted SD v1.5 since it is among the most popular diffusion models \url{https://huggingface.co/models?sort=likes}.}. The denoising UNet of our models offers a flexible range of sizes, with parameters spanning from \texttt{39M} to \texttt{5B}.
We incrementally increase the number of filters in the residual blocks while maintaining other architecture elements the same, enabling a predictably controlled scaling.
Table~\ref{tab:scaling_config} shows the architectural differences among our scaled models.
We also provide the relative cost of each model against the baseline model.
Fig.~\ref{fig:scaling_arch} shows the architectural differences during scaling.
Models were trained using the web-scale aesthetically filtered text-to-image dataset, \ie, WebLI~\citep{chen2022pali}.
All the models are trained for 500K steps, batch size 2048, and learning rate 1e-4.
This allows for all the models to have reached a point where we observe diminishing returns.
Fig.~\ref{fig:t2i_results} demonstrates the consistent generation capabilities across our scaled models. We used the common practice of 50 sampling steps with the DDIM sampler, 7.5 classifier-free guidance rate, for text-to-image generation. The visual quality of the results exhibits a clear improvement as model size increases.

In order to evaluate the performance of the scaled models, we test the text-to-image performance of scaled models on the validation set of COCO 2014~\citep{lin2014microsoft} with 30k samples.
For downstream performance, specifically real-world super-resolution, we test the performance of scaled models on the validation of DIV2K with 3k randomly cropped patches, which are degraded with the RealESRGAN degradation~\citep{wang2021real}.

\subsection{Training compute scales text-to-image performance}
\label{sec:scalingt2i}

We find that our scaled LDMs, across various model sizes, exhibit similar trends in generative performance relative to training compute cost, especially after training stabilizes, which typically occurs after 200K iterations.
These trends demonstrate a smooth scaling in learning capability between different model sizes.
To elaborate, Fig.~\ref{fig:t2i_compute} illustrates a series of training runs with models varying in size from 39 million to 5 billion parameters, where the training compute cost is quantified as the product of relative cost shown in Table~\ref{tab:scaling_config} and training iterations.
Model performance is evaluated by using the same sampling steps and sampling parameters.
In scenarios with moderate training compute (i.e., $<1G$, see Fig.~\ref{fig:t2i_compute}), the generative performance of T2I models scales well with additional compute resources.

\begin{figure}[t]
\centering
\includegraphics[height=.315\linewidth]{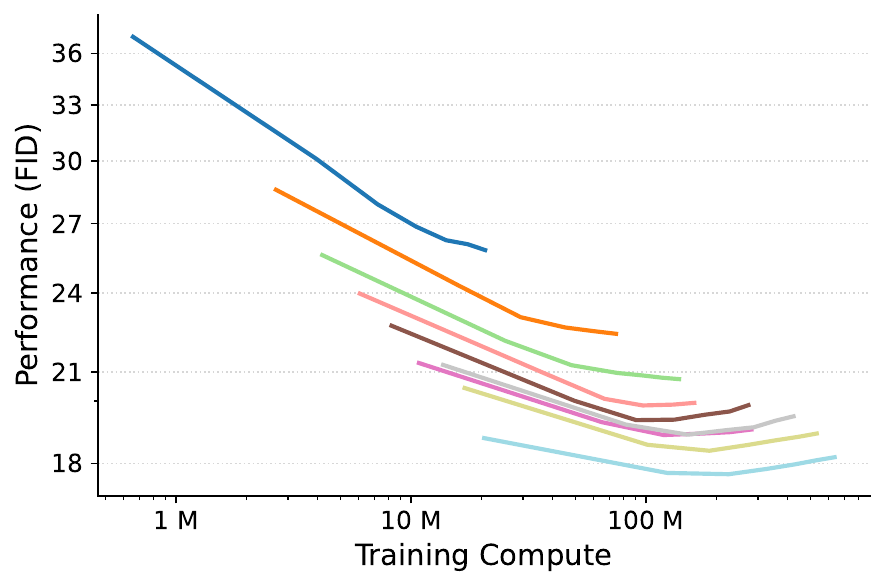}
\includegraphics[height=.315\linewidth]{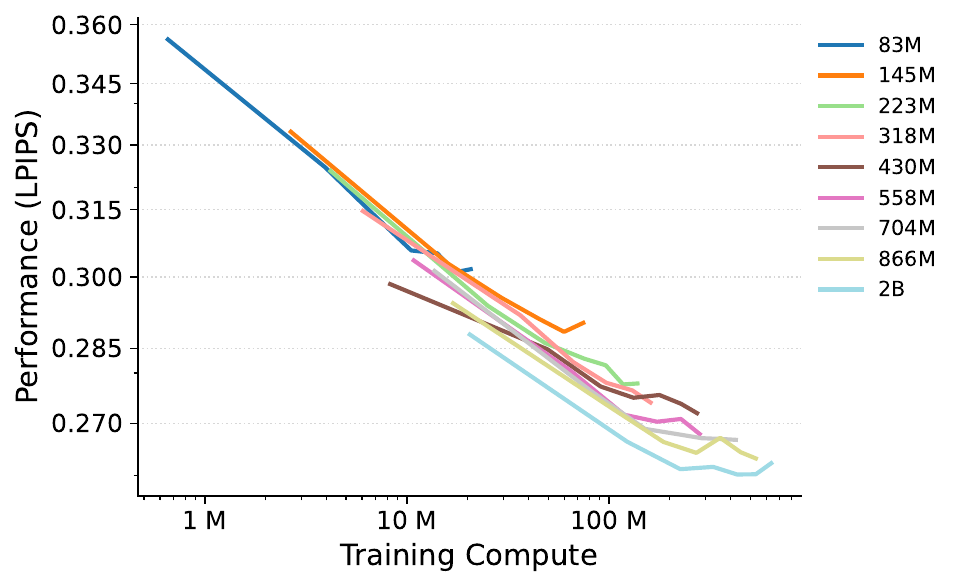}
\vspace{-1\baselineskip}
\caption{In $4\times$ real image super-resolution using 50-step DDIM sampling, FID and LPIPS scores reveal an interesting divergence. Model size drives FID score improvement, while training compute most impacts LPIPS score. Despite this, visual assessment (Fig.~\ref{fig:sr}) confirms the importance of model size for superior detail recovery (similarly as observed in the text-to-image pretraining).
}
\label{fig:sr_compute}
\vspace{-1\baselineskip}
\end{figure}

\begin{figure}[!ht]
    \scriptsize
    \centering
    \def\xwidth{0.12\linewidth}
    \setlength{\tabcolsep}{1pt}
    \begin{tabular}[t]{c c c c c c}
    &
    \includegraphics[width=\xwidth]{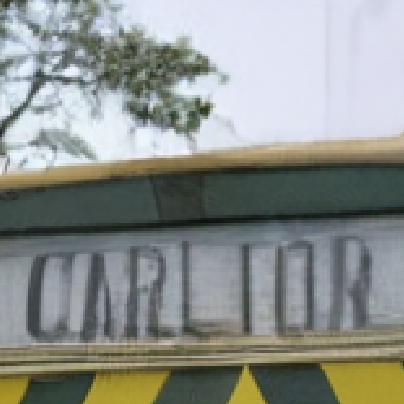} &
    \includegraphics[width=\xwidth]{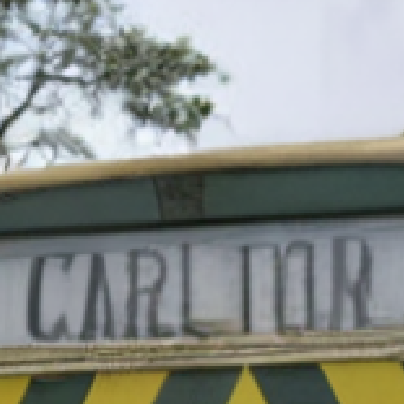} &
    \includegraphics[width=\xwidth]{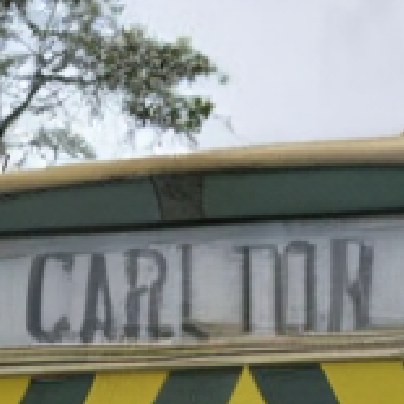} &
    \includegraphics[width=\xwidth]{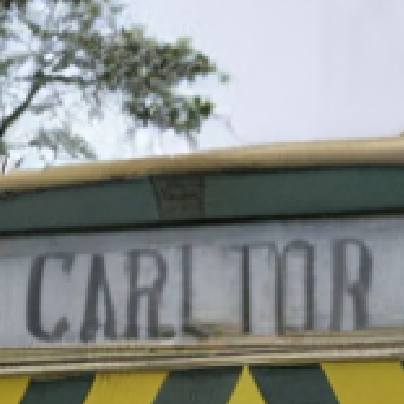} &
    \includegraphics[width=\xwidth]{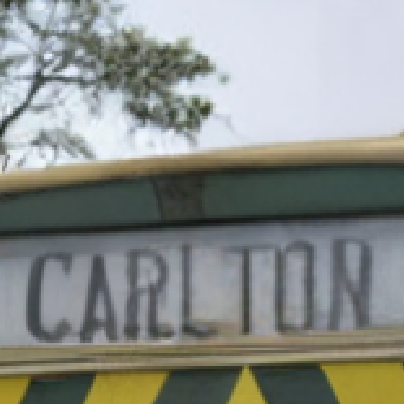}
    \\[-0.2em]
    & \scriptsize \texttt{83M} & \scriptsize \texttt{145M} & \scriptsize \texttt{223M} & \scriptsize \texttt{318M} & \scriptsize \texttt{430M}
    \\
    \multirow[t]{3}{*}{
    \includegraphics[width=0.2645\linewidth, height=0.2645\linewidth]{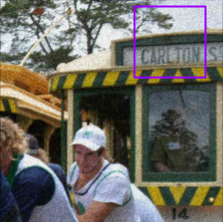}
    } &
    \includegraphics[width=\xwidth]{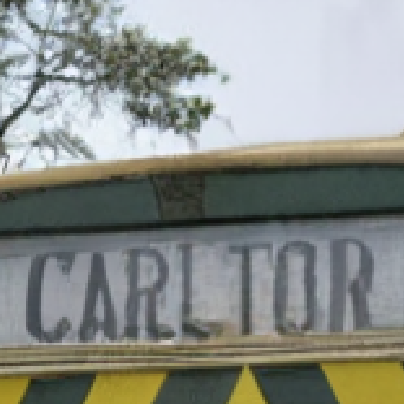} &
    \includegraphics[width=\xwidth]{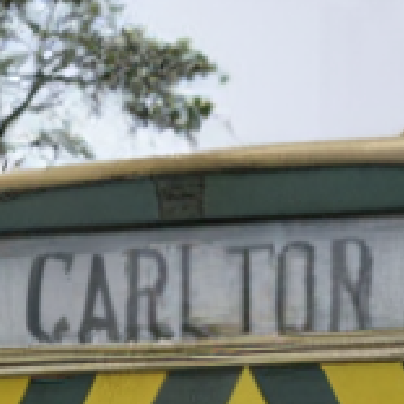} &
    \includegraphics[width=\xwidth]{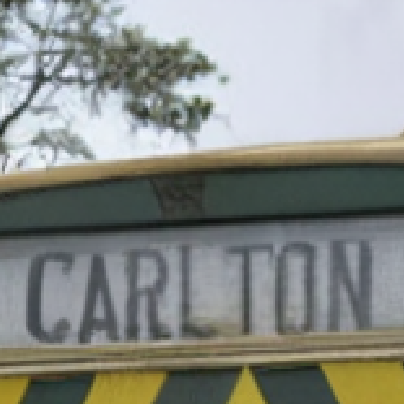} &
    \includegraphics[width=\xwidth]{figures/sr/scaling_sr_c512_2.png} &
    \includegraphics[width=\xwidth]{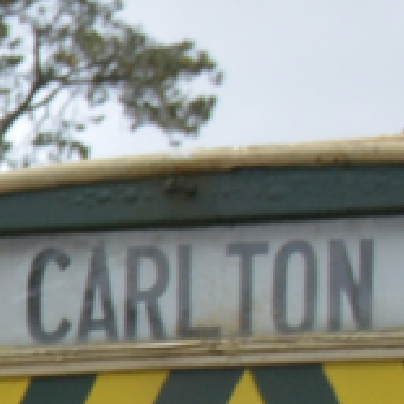}
    \\[-0.2em]
    \scriptsize LR &  \scriptsize \texttt{558M} & \scriptsize \texttt{704M} & \scriptsize \texttt{866M} & \scriptsize \texttt{2B} & \scriptsize HR
    \\
    &
    \includegraphics[width=\xwidth]{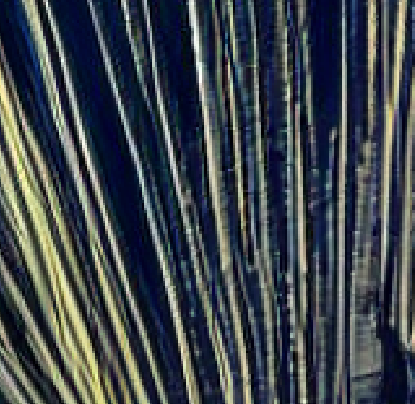} &
    \includegraphics[width=\xwidth]{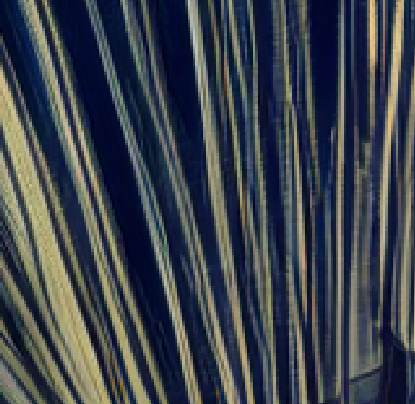} &
    \includegraphics[width=\xwidth]{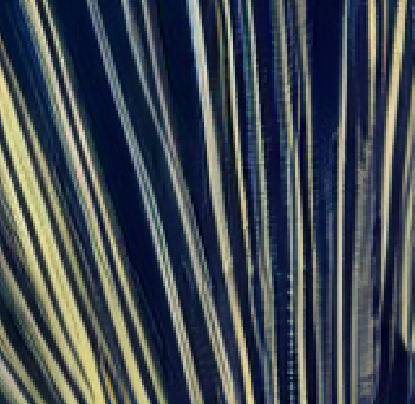} &
    \includegraphics[width=\xwidth]{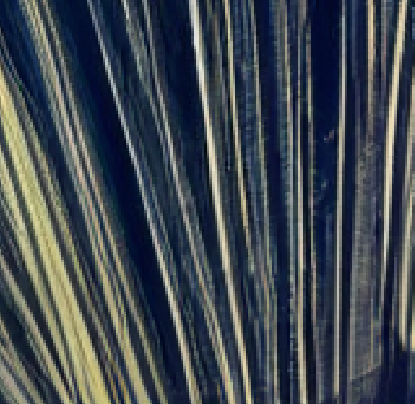} &
    \includegraphics[width=\xwidth]{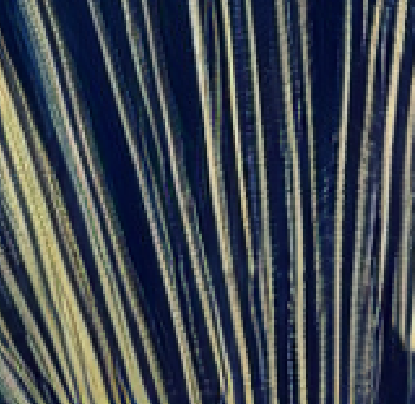}
    \\[-0.2em]
    & \scriptsize \texttt{83M} & \scriptsize \texttt{145M} & \scriptsize \texttt{223M} & \scriptsize \texttt{318M} & \scriptsize \texttt{430M}
    \\
    \multirow[t]{3}{*}{
    \includegraphics[width=0.26\linewidth,height=0.26\linewidth]{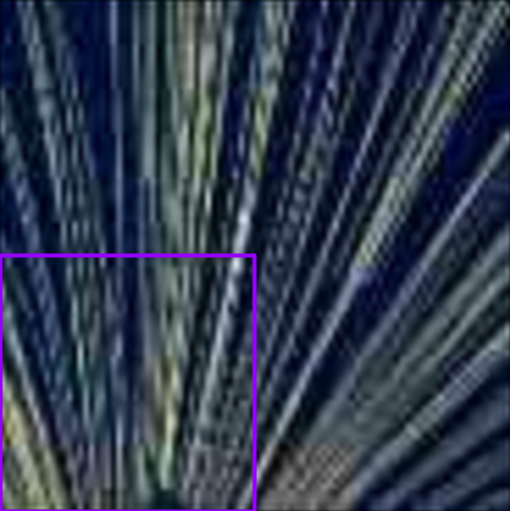}
    } &
    \includegraphics[width=\xwidth]{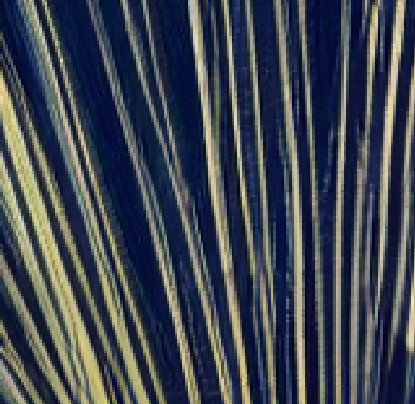} &
    \includegraphics[width=\xwidth]{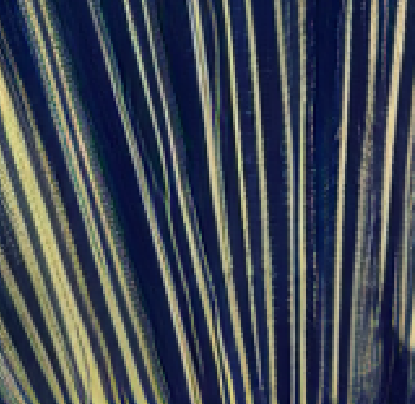} &
    \includegraphics[width=\xwidth]{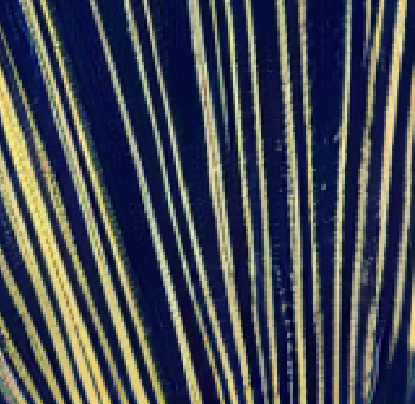} &
    \includegraphics[width=\xwidth]{figures/sr/scaling_sr_c512_1.png} &
    \includegraphics[width=\xwidth]{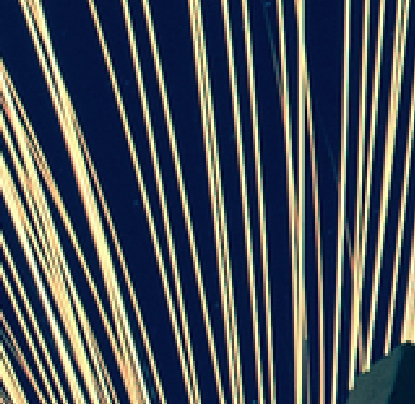}
    \\[-0.2em]
    \scriptsize LR &  \scriptsize \texttt{558M} & \scriptsize \texttt{704M} & \scriptsize \texttt{866M} & \scriptsize \texttt{2B} & \scriptsize HR
    \end{tabular}
    \vspace{-.5\baselineskip}
    \caption{In 4$\times$ super-resolution using 50-step DDIM sampling, visual quality directly improves with increased model size. As these scaled models vary in pretraining performance, the results clearly demonstrate that pretraining boosts super-resolution capabilities in both quantitative (Fig~\ref{fig:sr_compute}) and qualitative ways. \emph{Additional results are given in supplementary material.}}
    \label{fig:sr}
    \vspace{-1\baselineskip}
\end{figure}

\begin{figure*}[!ht]
\scriptsize
    \centering
    \def\xwidth{0.135\linewidth}
    \setlength{\tabcolsep}{0.5pt}
    \begin{tabular}[t]{c c c c c}
    &
    \includegraphics[width=\xwidth]{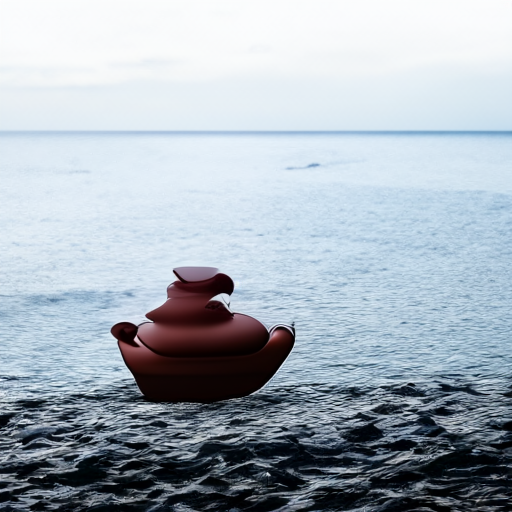} &
    \includegraphics[width=\xwidth]{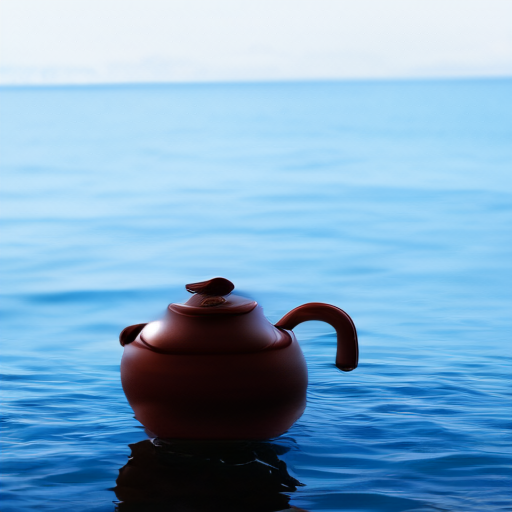} &
    \includegraphics[width=\xwidth]{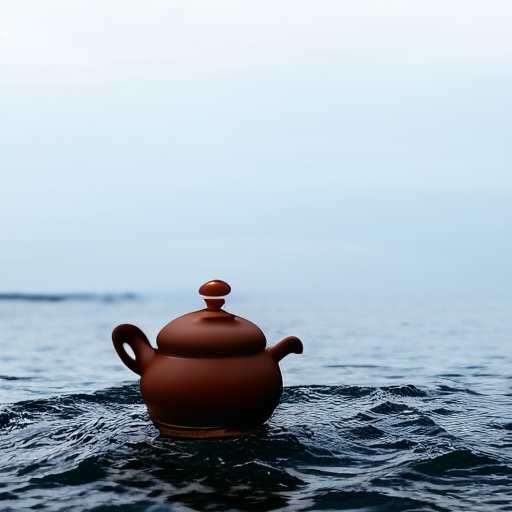} &
    \includegraphics[width=\xwidth]{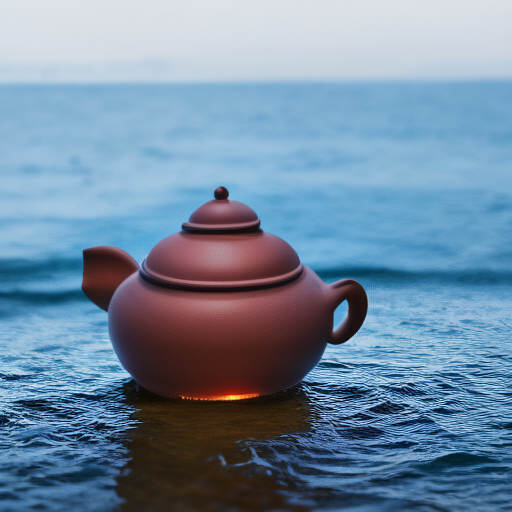} \\
    & \texttt{83M} & \texttt{145M} & \texttt{223M} & \texttt{318M} \\
    \multirow[t]{3}{*}{
    \includegraphics[width=0.3\linewidth]{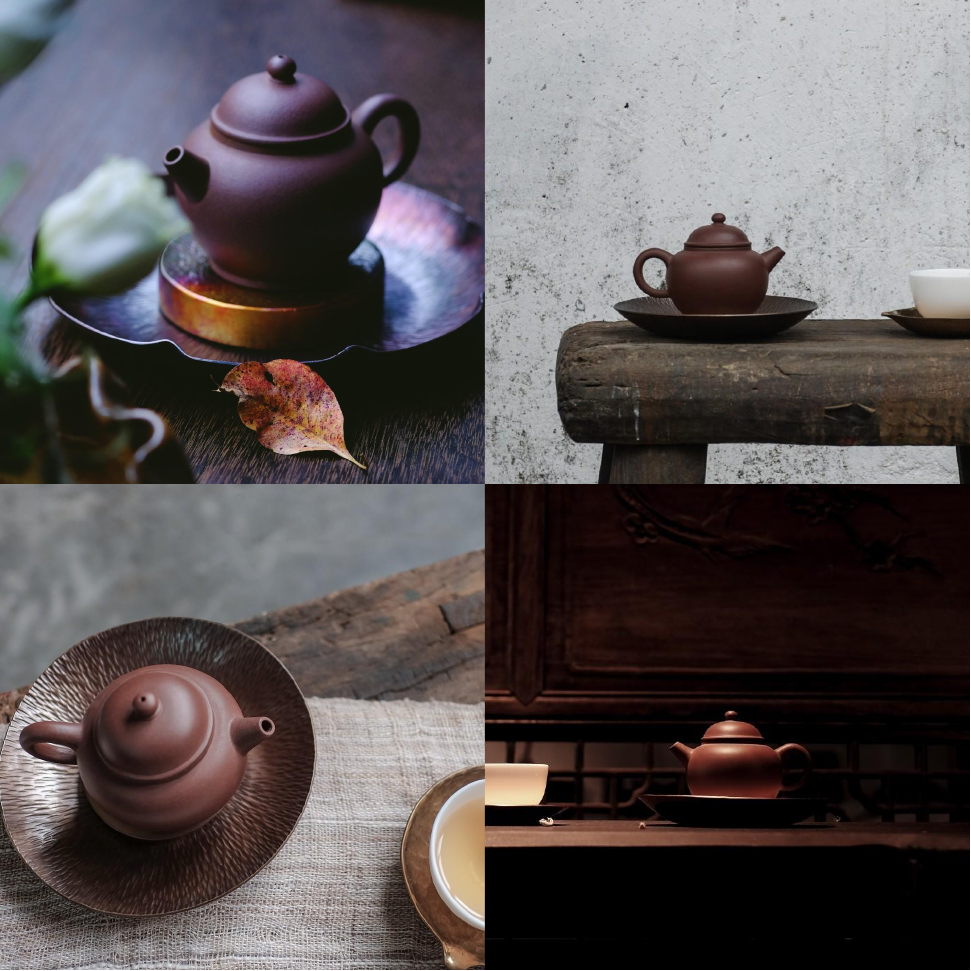}
    } &
    \includegraphics[width=\xwidth]{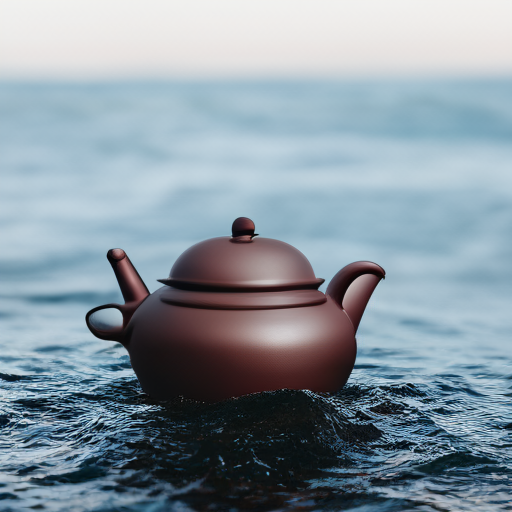} &
    \includegraphics[width=\xwidth]{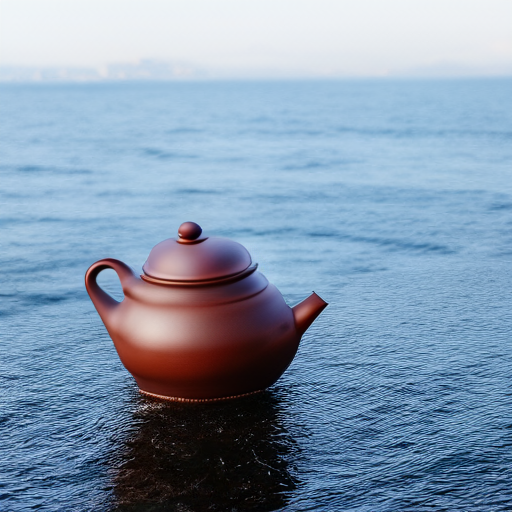} &
    \includegraphics[width=\xwidth]{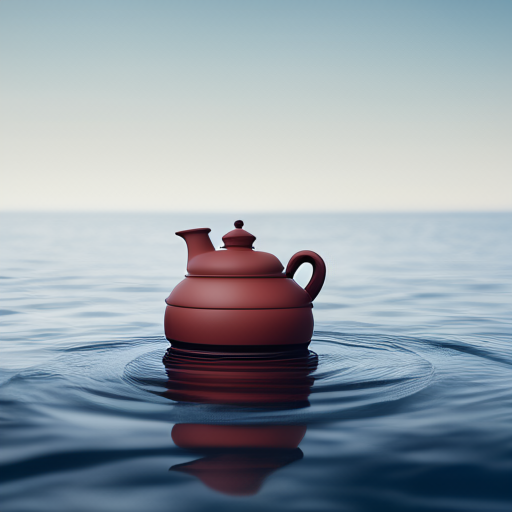} &
    \includegraphics[width=\xwidth]{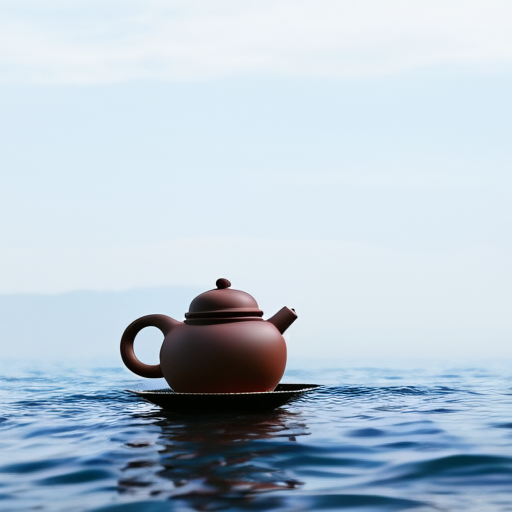} \\
    Inputs & \texttt{430M} &  \texttt{558M} &  \texttt{866M} &  \texttt{2B} \\
    \\
    &
    \includegraphics[width=\xwidth]{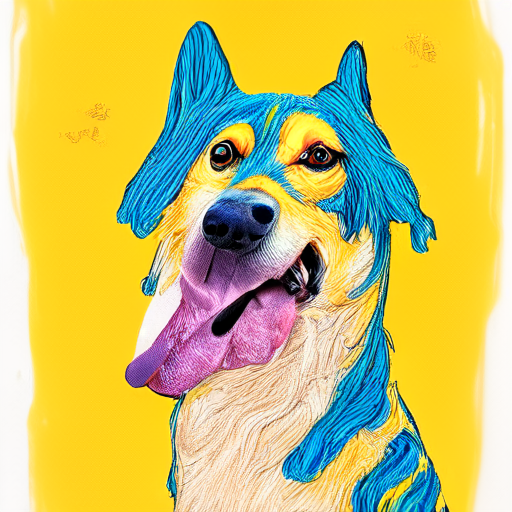} &
    \includegraphics[width=\xwidth]{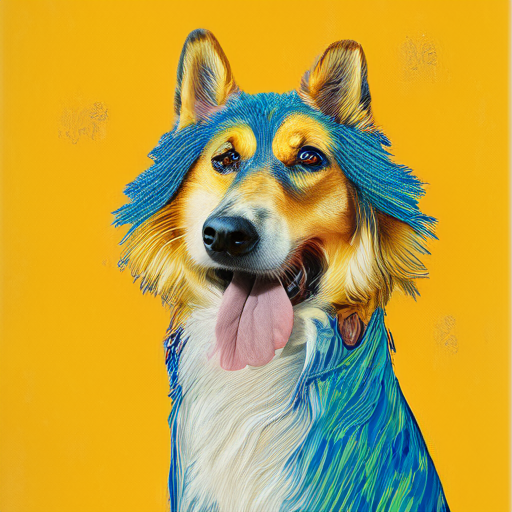} &
    \includegraphics[width=\xwidth]{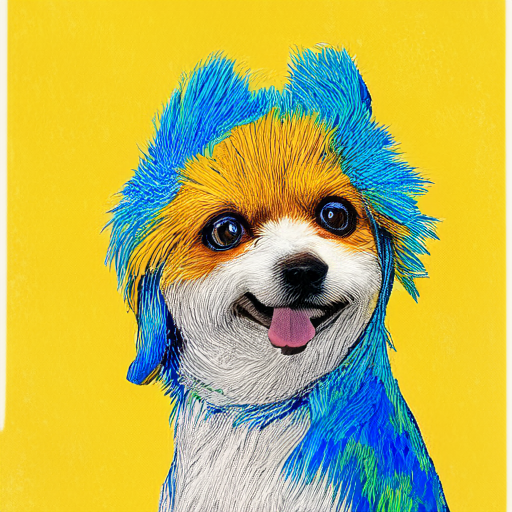} &
    \includegraphics[width=\xwidth]{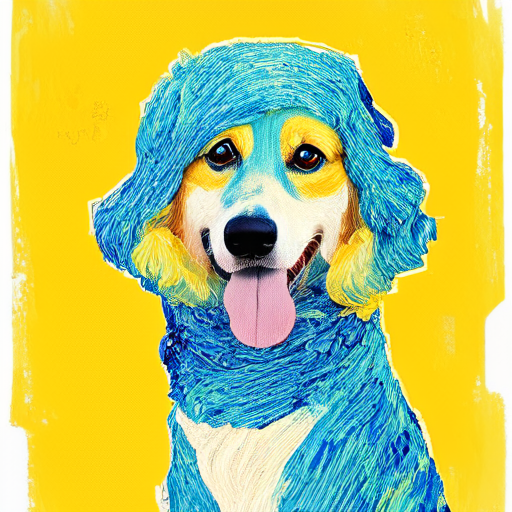} \\
    &  \texttt{83M} &  \texttt{145M} &  \texttt{223M} &  \texttt{318M} \\
    \multirow[t]{3}{*}{
    \includegraphics[width=0.3\linewidth]{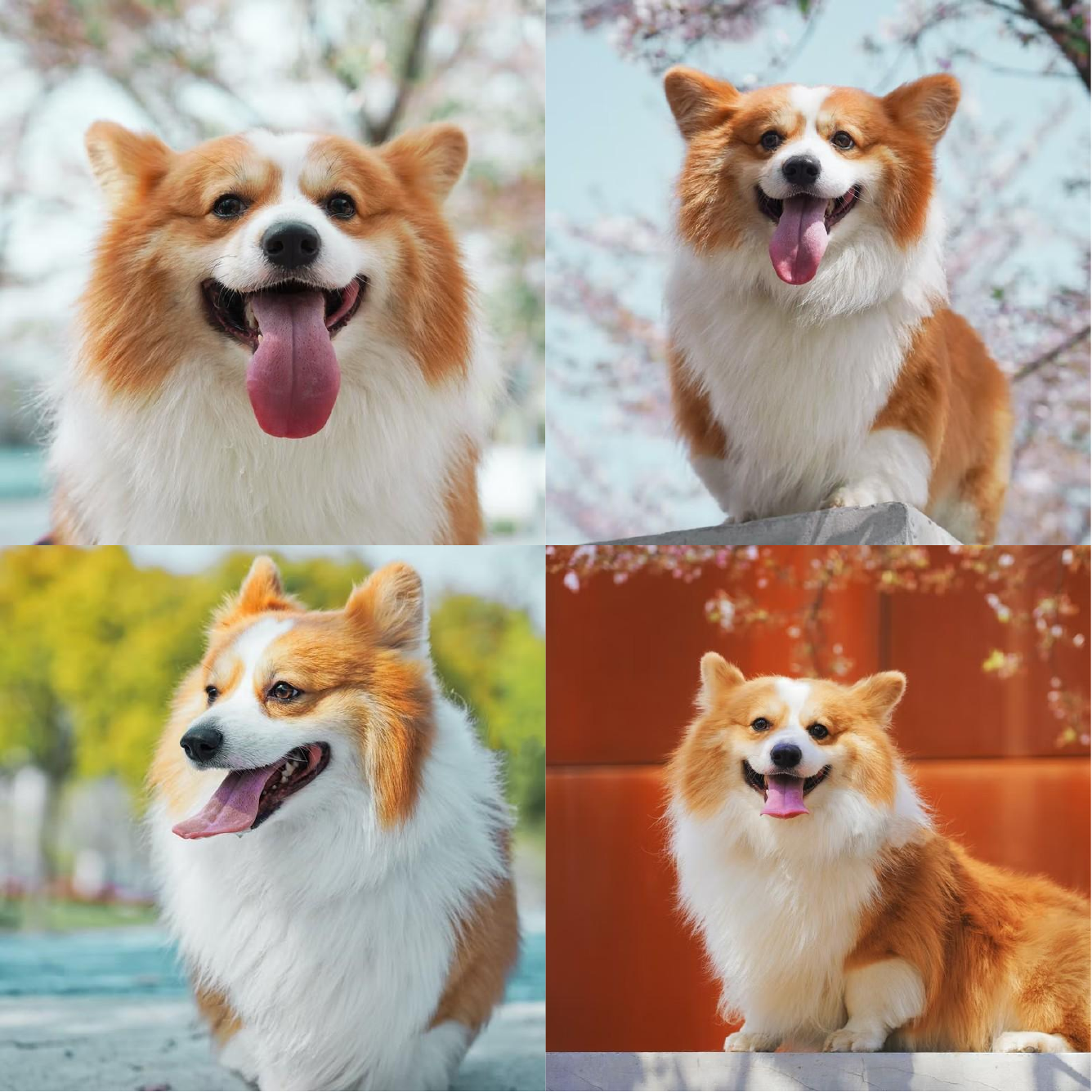}
    } &
    \includegraphics[width=\xwidth]{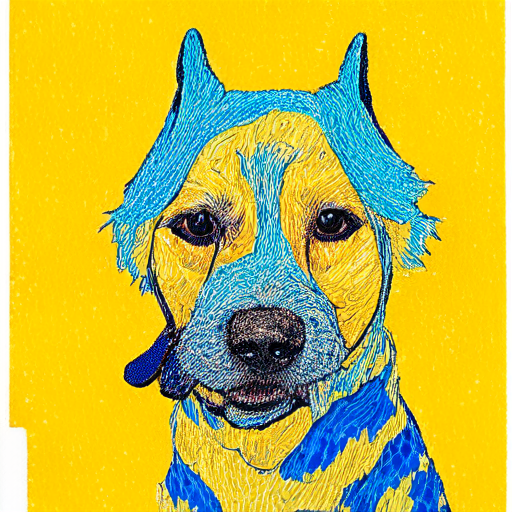} &
    \includegraphics[width=\xwidth]{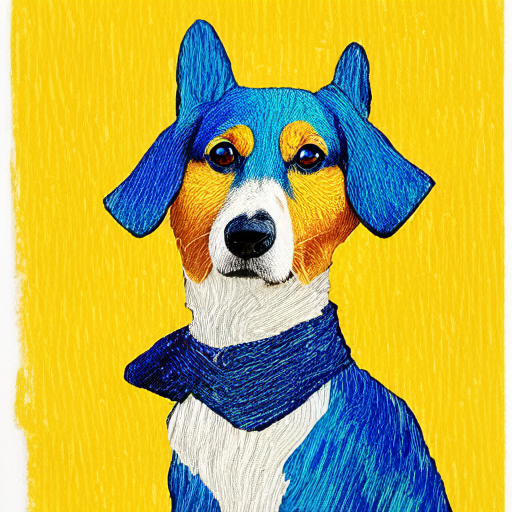} &
    \includegraphics[width=\xwidth]{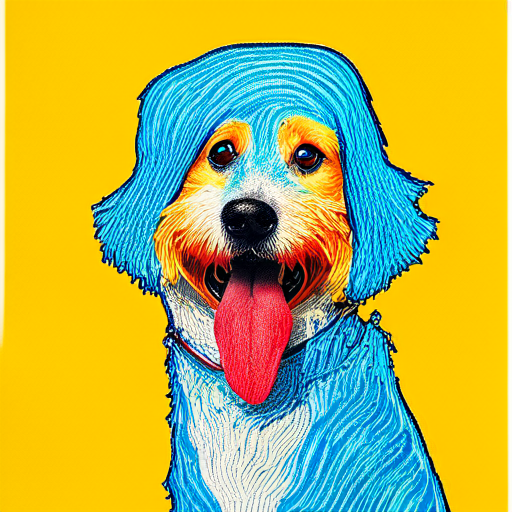} &
    \includegraphics[width=\xwidth]{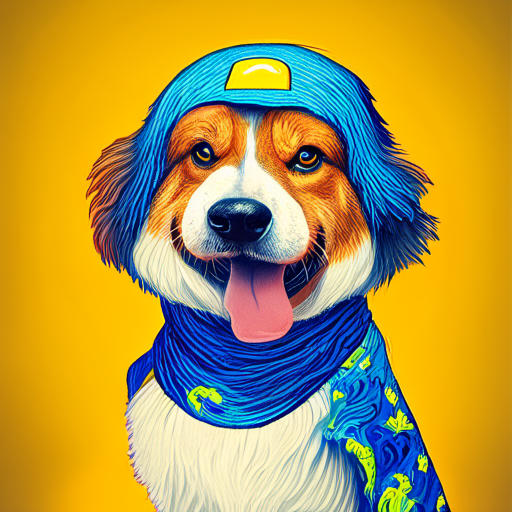} \\
    Inputs &  \texttt{430M} &  \texttt{558M} &  \texttt{866M} &  \texttt{2B} \\
    
    \end{tabular}
    \vspace{-.5\baselineskip}
    \caption{
    Visualization of the Dreambooth results (using 50-step DDIM sampling and CFG rate of 7.5) shows two distinct tiers based on model size.  Smaller models (\texttt{83M}-\texttt{223M}) perform similarly, as do larger ones (\texttt{318M}-\texttt{2B}),  with a clear quality advantage for the larger group. \emph{Additional results are given in supplementary material.}}
    \label{fig:dreambooth}
    \vspace{-1\baselineskip}
\end{figure*}

\subsection{Pretraining scales downstream performance}
\label{sec:scalingsr}
Using scaled models based on their pretraining on text-to-image data, we finetune these models on the downstream tasks of real-world super-resolution~\citep{saharia2022image, sahak2023denoising} and DreamBooth~\citep{ruiz2023dreambooth}.
The performance of these pretrained models is shown in Table.~\ref{tab:scaling_config}.
In the left panel of Fig.~\ref{fig:sr_compute}, we present the generative performance FID versus training compute on the super-resolution (SR) task.
It can be seen that the performance of SR models is more dependent on the model size than training compute.
Our results demonstrate a clear limitation of smaller models: they cannot reach the same performance levels as larger models, regardless of training compute.

While the distortion metric LPIPS shows some inconsistencies compared to the generative metric FID (Fig.~\ref{fig:sr_compute}), Fig.~\ref{fig:sr} clearly demonstrates that larger models excel in recovering fine-grained details compared to smaller models.

The key takeaway from Fig.~\ref{fig:sr_compute} is that large super-resolution models achieve superior results even after short finetuning periods compared to smaller models. This suggests that pretraining performance (dominated by the pretraining model sizes) has a greater influence on the super-resolution FID scores than the duration of finetuning (\ie, training compute for finetuning).

Furthermore, we compare the visual results of the DreamBooth finetuning on the different models in Fig.~\ref{fig:dreambooth}. We observe a similar trend between  visual quality and  model size.
\emph{Please see our supplement for more discussions on the other quality metrics.}

\subsection{Scaling sampling-efficiency}

\begin{figure}[!t]
    \centering
    \begin{subfigure}[b]{\linewidth}
    \includegraphics[width=\linewidth]{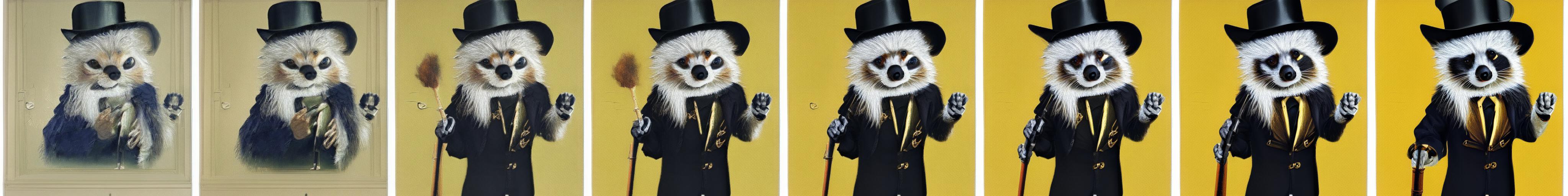}
    \caption{50-step sampling results of the 145M model}
    \end{subfigure}
    \begin{subfigure}[b]{\linewidth}
    \includegraphics[width=\linewidth]{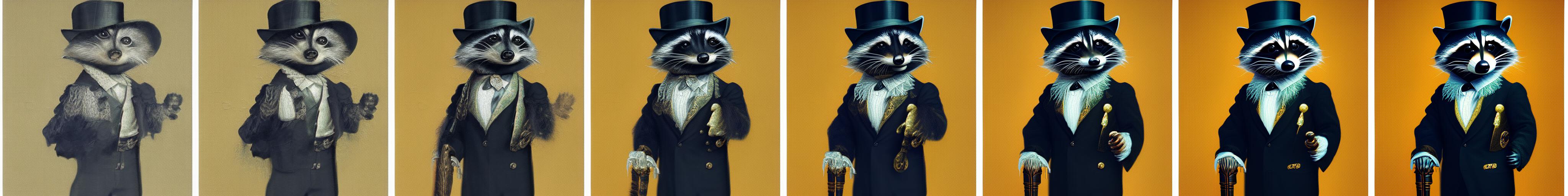}
    \caption{50-step sampling results of the 866M model}
    \end{subfigure}
    \caption{Visualization of text-to-image results with 50-step DDIM sampling and different CFG rates (from left to right in each row: $(1.5, 2.0, 3.0, 4.0, 5.0, 6.0, 7.0, 8.0)$). The prompt used is ``\emph{A raccoon wearing formal clothes, wearing a top hat and holding a cane. Oil painting in the style of Rembrandt.}''.
    We observe that changes in CFG rates impact visual quality more significantly than the prompt semantic accuracy.
    We use the FID score for quantitative determination of optimal sampling performance (Fig.~\ref{fig:cfgrate}) because it directly measures visual quality, unlike the CLIP score, which focuses on semantic similarity. }
    \label{fig:cfgratevisual}
    \vspace{-1\baselineskip}
\end{figure}

\begin{figure}[t]
    \centering
    \includegraphics[width=\linewidth]{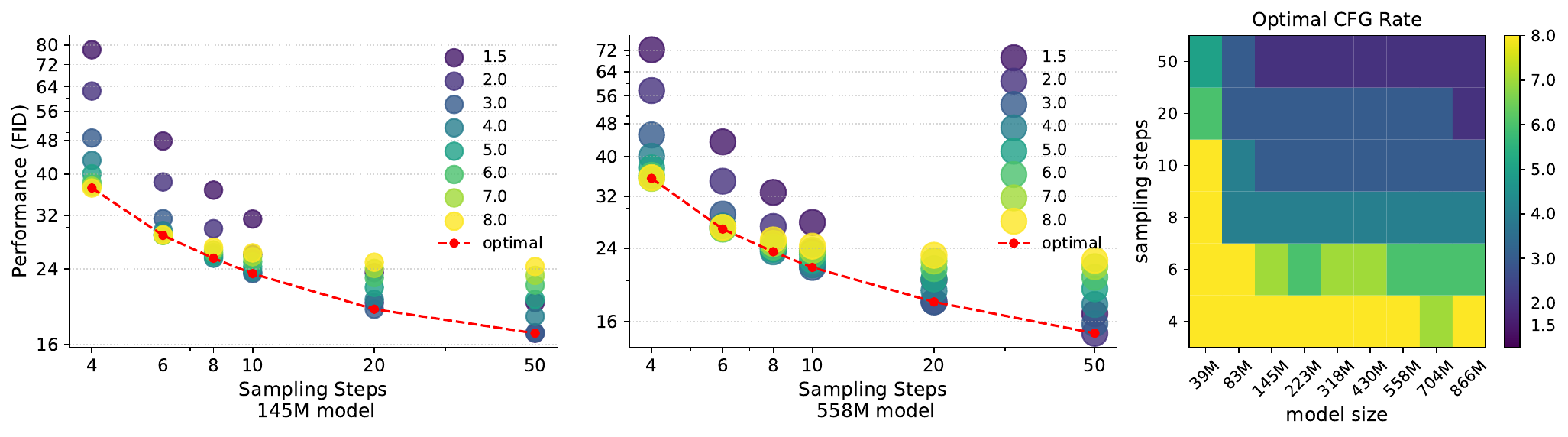}
    \vspace{-1.5\baselineskip}
    \caption{
    The impact of the CFG rate on text-to-image generation depends on the model size and sampling steps. As demonstrated in the left and center panels, the optimal CFG rate changes as the sampling steps increased.
    To determine the optimal performance (according to the FID score) of each model and each sampling steps, we systematically sample the model at various CFG rates and identify the best one. As a reference of the optimal performance, the right panel shows the CFG rate corresponding to the optimal performance of each model for a given number of sampling steps.
    }
    \label{fig:cfgrate}
\end{figure}

\subsubsection{Analyzing the effect of CFG rate.}
\label{sec:optimalparams}

Text-to-image generative models require nuanced evaluation beyond single metrics. Sampling parameters are vital for customization, with the Classifier-Free Guidance (CFG) rate~\citep{ho2022classifier} directly influencing the balance between visual fidelity and semantic alignment with text prompt. Rombach et al.~\citep{rombach2022high} experimentally demonstrate that different CFG rates result in different CLIP and FID scores.

\begin{figure}[!t]
    \centering 
    \def\xwidth{.3\linewidth}
    \includegraphics[height=\xwidth]{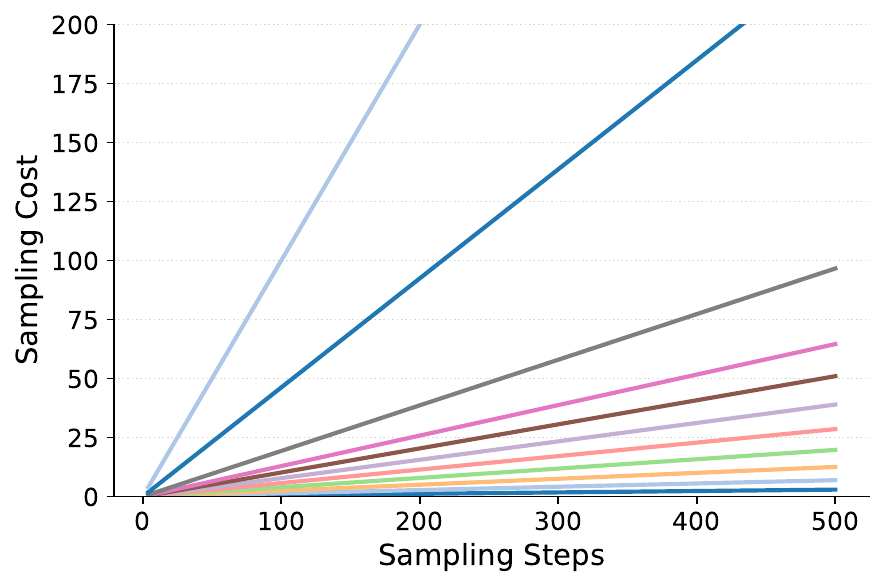}
    \includegraphics[height=\xwidth]{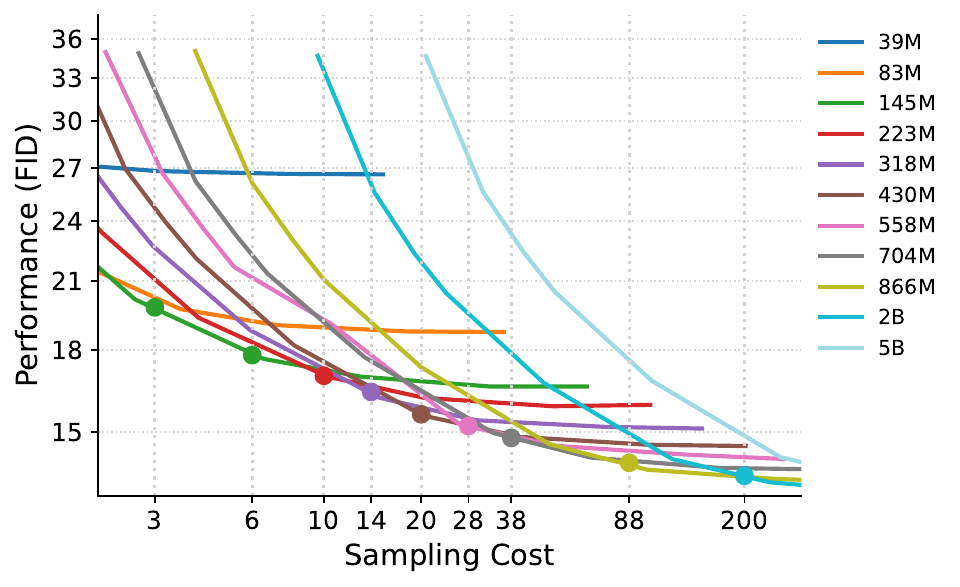}
    \vspace{-.5\baselineskip}
    \caption{%
    Comparison of text-to-image performance of models with varying sizes. The left figure shows the relationship between sampling cost (normalized cost $\times$ sampling steps) and sampling steps for different model sizes. The right figure plots the optimal text-to-image FID score among CFG rates of $(1.5, 2.0, 3.0, 4.0, 5.0, 6.0, 7.0, 8.0)$ as a function of the sampling cost for the same models.
    Key Observation: Smaller models achieve better FID scores than larger models for a fixed sampling cost. For instance, at a cost of 3, the 83M model achieves the best FID compared to the larger models. This suggests that smaller models can be more efficient in achieving good results with lower costs.
    }
    \label{fig:optiamlrules}
\end{figure}

\begin{figure}[!ht]
    \centering
    \def\xwidth{.24\linewidth}
    
     \begin{subfigure}[b]{\linewidth}
    \imageWithNote{\xwidth}{\scriptsize \raggedleft \texttt{83M\,}}{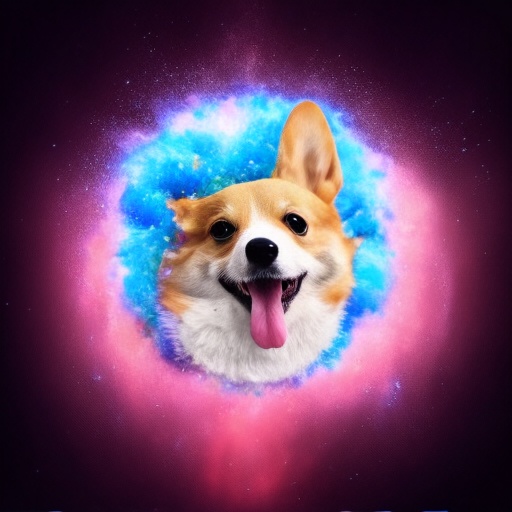}
    \imageWithNote{\xwidth}{\scriptsize \raggedleft \texttt{145M}}{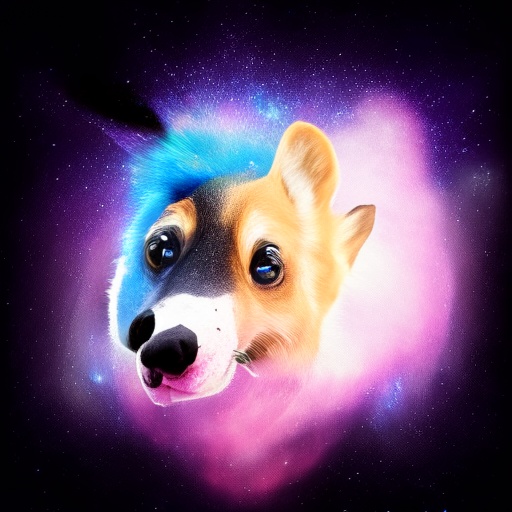}
    \imageWithNote{\xwidth}{\scriptsize \raggedleft \texttt{223M}}{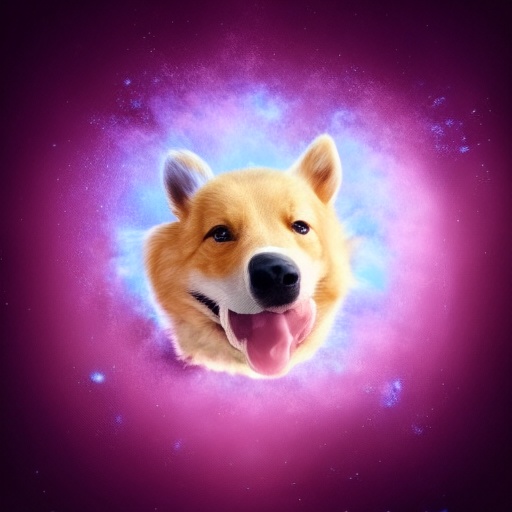}
    \imageWithNote{\xwidth}{\scriptsize \raggedleft \texttt{318M}}{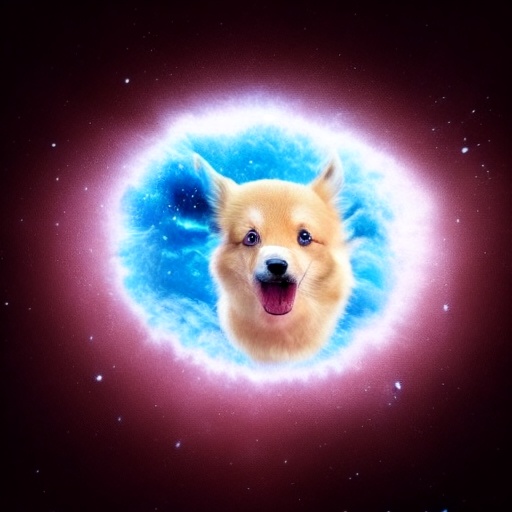}
    \hfill
    
    \imageWithNote{\xwidth}{\scriptsize \raggedleft \texttt{430M}}{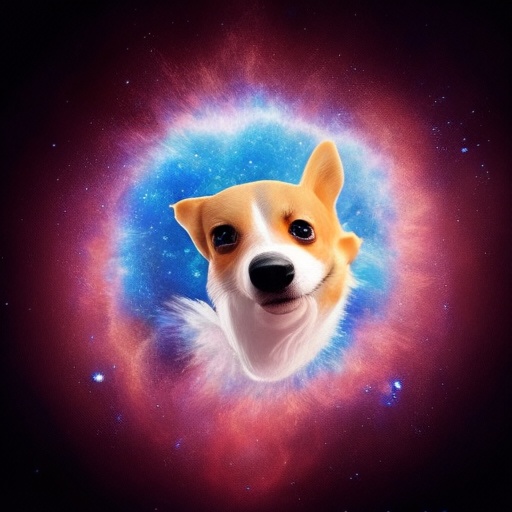}
    \imageWithNote{\xwidth}{\scriptsize \raggedleft \texttt{558M}}{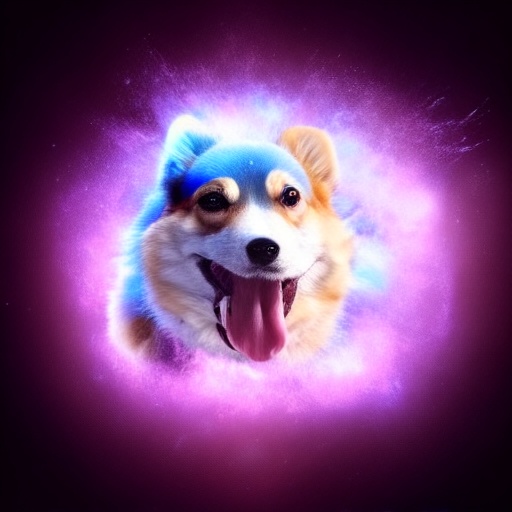}
    \imageWithNote{\xwidth}{\scriptsize \raggedleft \texttt{704M}}{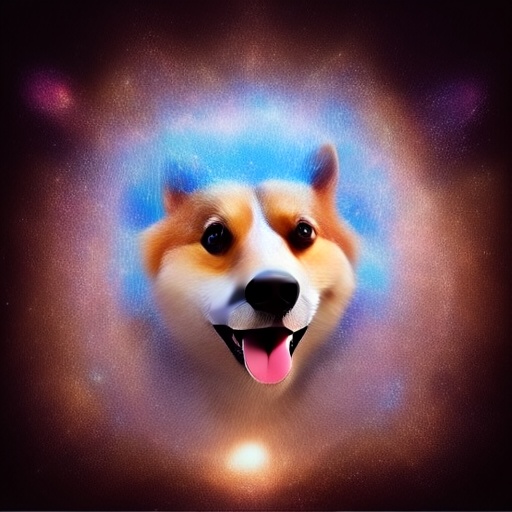}
    \imageWithNote{\xwidth}{\scriptsize \raggedleft \texttt{866M}}{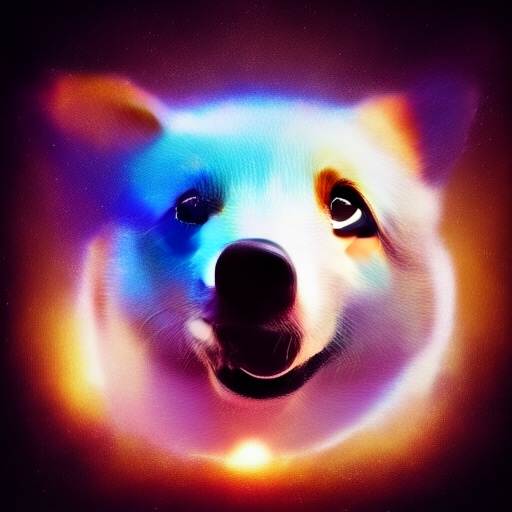}
    \vspace{.5em}
    \caption{Prompt: \emph{``A corgi's head depicted as a nebula.''}. Sampling Cost $\approx$ 6.}
    \end{subfigure}
    \vspace{1em}
    \begin{subfigure}[b]{\linewidth}
    
    \imageWithNote{\xwidth}{\scriptsize \raggedleft \texttt{83M\,}}{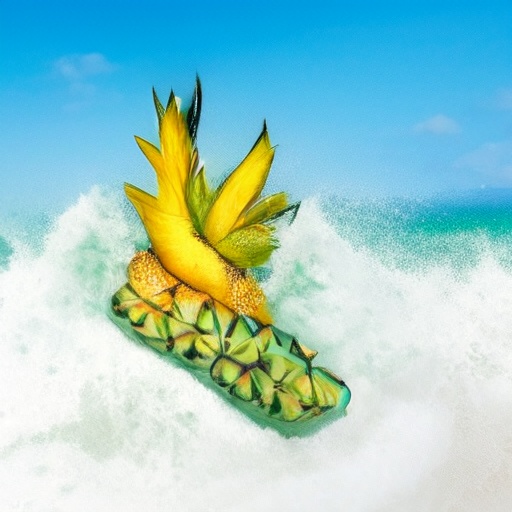}
    \imageWithNote{\xwidth}{\scriptsize \raggedleft \texttt{145M}}{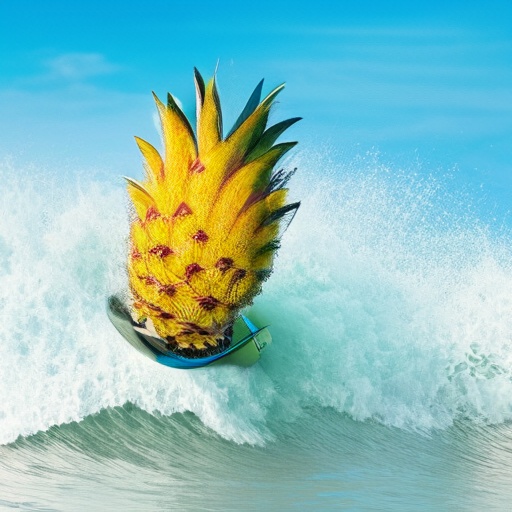}
    \imageWithNote{\xwidth}{\scriptsize \raggedleft \texttt{223M}}{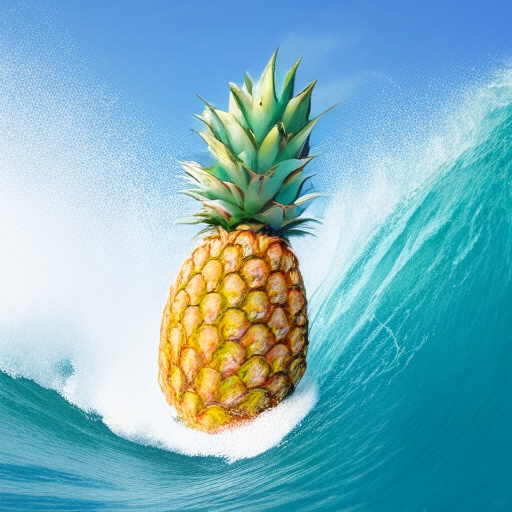}
    \imageWithNote{\xwidth}{\scriptsize \raggedleft \texttt{318M}}{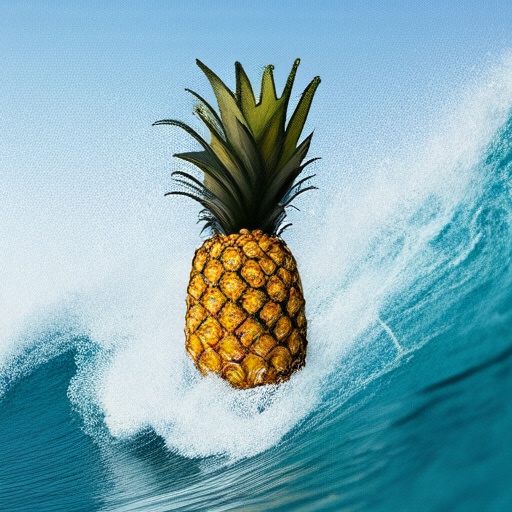}
    \hfill
    
    \imageWithNote{\xwidth}{\scriptsize \raggedleft \texttt{430M}}{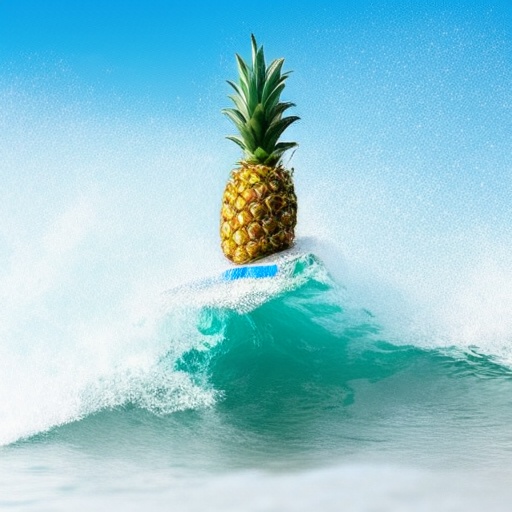}
    \imageWithNote{\xwidth}{\scriptsize \raggedleft \texttt{558M}}{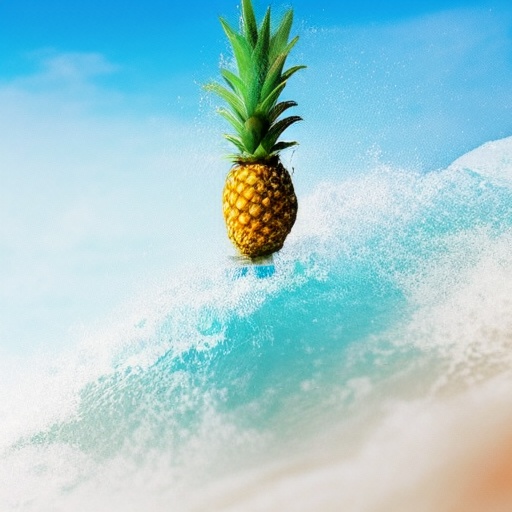}
    \imageWithNote{\xwidth}{\scriptsize \raggedleft \texttt{704M}}{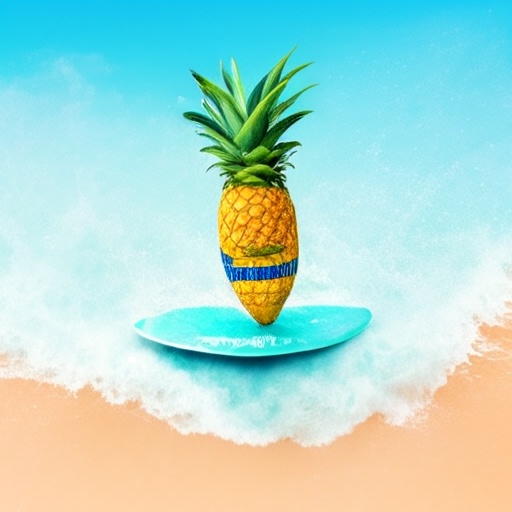}
    \imageWithNote{\xwidth}{\scriptsize \raggedleft \texttt{866M}}{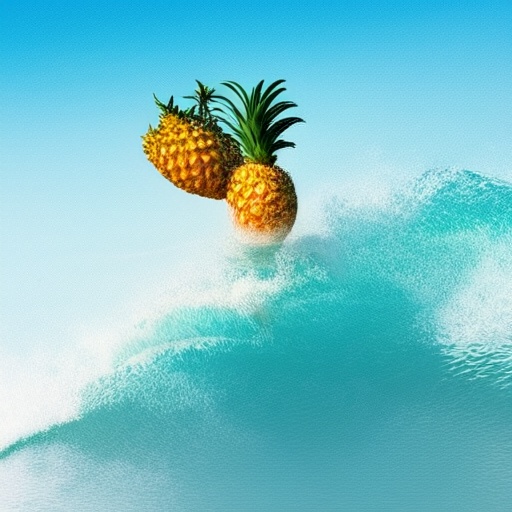}
    \vspace{.5em}
    \caption{Prompt: \emph{``A pineapple surfing on a wave.''}. Sampling Cost $\approx$ 12.}
    \end{subfigure}
    \vspace{-2\baselineskip}
    \caption{Text-to-image results of the scaled LDMs under approximately the same inference cost (normalized cost $\times$ sampling steps). Smaller models can produce comparable or even better visual results than larger models under similar sampling cost.
    }
    \label{fig:optimalvisual}
    \vspace{-1\baselineskip}
\end{figure}

\begin{figure}[ht]
    \centering
    \def\xwidth{.3\linewidth}
     \includegraphics[height=\xwidth]{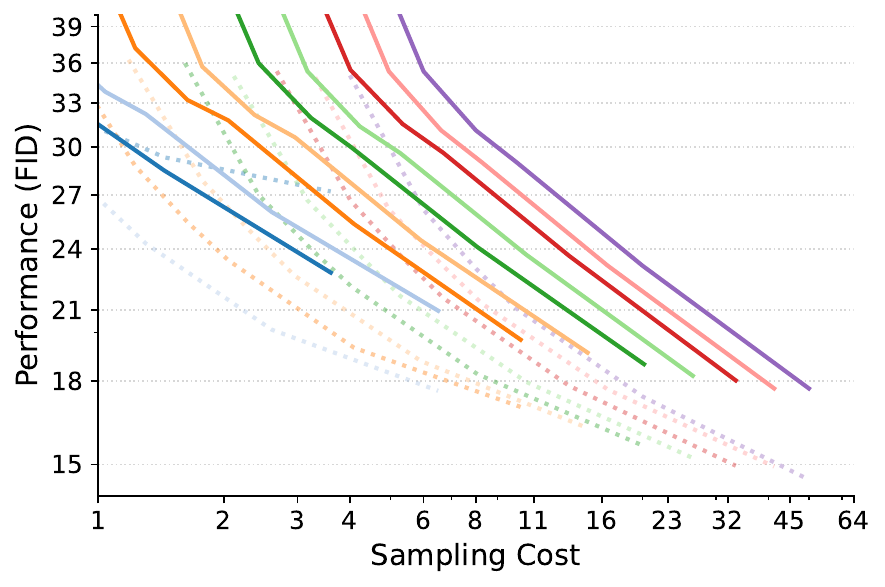}
    \includegraphics[height=\xwidth]{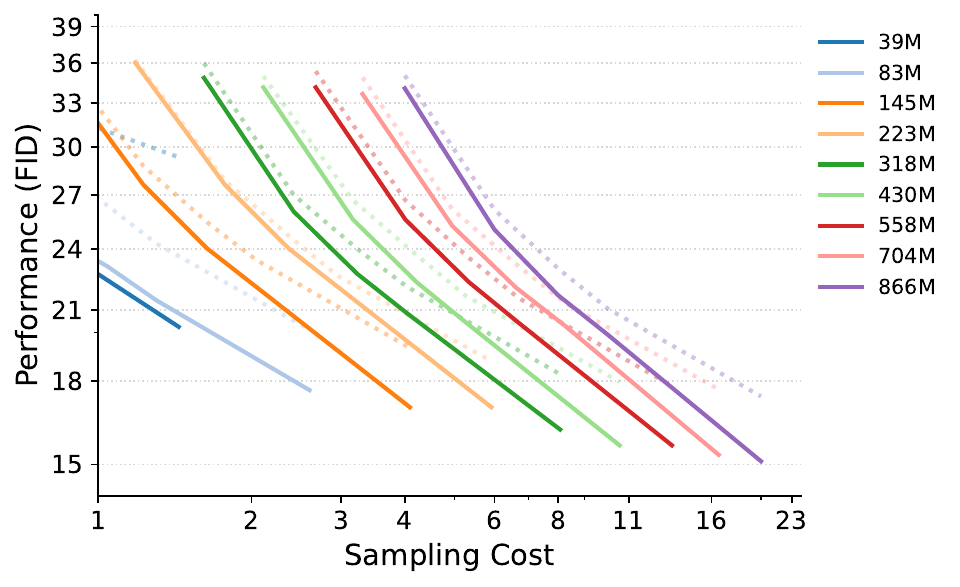}
    \vspace{-.5\baselineskip}
    \caption{\emph{Left}: Text-to-image performance FID as a function of the sampling cost (normalized cost $\times$ sampling steps) for the DDPM sampler (solid curves) and the DDIM sampler (dashed curves). \emph{Right}: Text-to-image performance FID as a function of the sampling cost for the second-order DPM-Solver++ sampler (solid curves) and the DDIM sampler (dashed curves). Suggested by the trends shown in Fig.~\ref{fig:optiamlrules}, we only show the sampling steps $\leq 50$ as using more steps does not improve the performance.}
    \label{fig:scalingsampler}
    \vspace{-1\baselineskip}
\end{figure}

In this study, we find that CFG rate as a sampling parameter yields inconsistent results across different model sizes.
Hence, it is interesting to quantitatively determine the \textit{optimal} CFG rate for each model size and sampling steps using either FID or CLIP score.
We demonstrate this by sampling the scaled models using different CFG rates, \ie, $(1.5, 2.0, 3.0, 4.0, 5.0, 6.0, 7.0, 8.0)$ and comparing their quantitative and qualitative results. 
In Fig.~\ref{fig:cfgratevisual}, we present visual results of two models under varying CFG rates, highlighting the impact on the visual quality. We observed that changes in CFG rates impact visual quality more significantly than prompt semantic accuracy and therefore opted to use the FID score for quantitative determination of the optimal CFG rate. performance. Fig.~\ref{fig:cfgrate} shows how different classifier-free guidance rates affect the FID scores in text-to-image generation (see figure caption for more details).

\subsubsection{Scaling efficiency trends.}
\label{sec:optimalmodelsizes}

Using the optimal CFG rates established for each model at various number of sampling steps, we analyze the optimal performance to understand the sampling efficiency of different LDM sizes.
Specifically, in Fig.~\ref{fig:optiamlrules}, we present a comparison between different models and their optimal performance given the sampling cost (normalized cost $\times$ sampling steps).
By tracing the points of optimal performance across various sampling cost---represented by the dashed vertical line---we observe a consistent trend: smaller models frequently outperform larger models across a range of sampling cost in terms of FID scores.
Furthermore, to visually substantiate better-quality results generated by smaller models against larger ones, Fig.~\ref{fig:optimalvisual} compares the results of different scaled models, which highlights that the performance of smaller models can indeed match their larger counterparts under similar sampling cost conditions.
\emph{
Please see our supplement for more visual comparisons.
}

\subsection{Scaling sampling-efficiency in different samplers}
\label{sec:samplerscaling}

To assess the generalizability of observed scaling trends in sampling efficiency, we compared scaled LDM performance using different diffusion samplers. In addition to the default DDIM sampler, we employed two representative alternatives: the stochastic DDPM sampler~\citep{ho2020denoising} and the high-order DPM-Solver++~\citep{lu2022dpm2}.

Experiments illustrated in Fig.~\ref{fig:scalingsampler} reveal that the DDPM sampler typically produces lower-quality results than DDIM with fewer sampling steps, while the DPM-Solver++ sampler generally outperforms DDIM in image quality (see the figure caption for details).
Importantly, we observe consistent sampling-efficiency trends with the DDPM and DPM-Solver++ sampler as seen with the default DDIM: smaller models tend to achieve better performance than larger models under the same sampling cost. Since the DPM-Solver++ sampler is not designed for use beyond 20 steps, we focused our testing within this range.
This finding demonstrates that the scaling properties of LDMs remain consistent regardless of the diffusion sampler used.

\subsection{Scaling downstream sampling-efficiency}
\label{sec:scalingsamplingsr}

\begin{figure}[t]
    \centering
    \def\xwidth{.315\linewidth}
    \includegraphics[height=\xwidth]{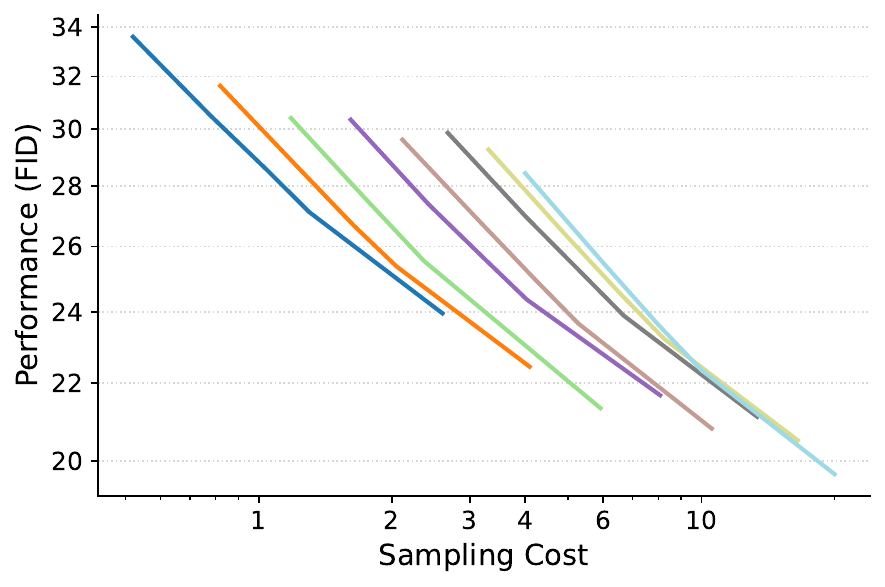}
    \includegraphics[height=\xwidth]{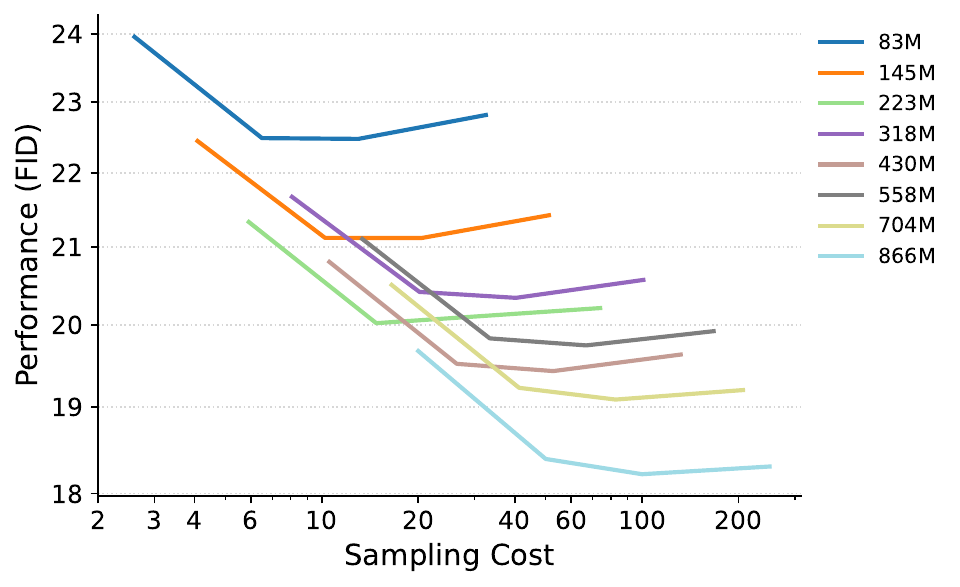}
    \vspace{-1.5\baselineskip}
    \caption{Super-resolution performance vs. sampling cost for different model sizes. \emph{Left:} FID scores of super-resolution models under limited sampling steps (less than or equal to 20). Smaller models tend to achieve lower (better) FID scores within this range.
    \emph{Right:} FID scores of super-resolution models under a larger number of sampling steps (greater than 20).
    Performance differences between models become less pronounced as sampling steps increase.
    }
    \label{fig:sroptiamlrules}
    \vspace{-1\baselineskip}
\end{figure}

Here, we investigate the scaling sampling-efficiency of LDMs on downstream tasks, specifically focusing on the super-resolution task.
Unlike our earlier discussions on optimal sampling performance, there is limited literature demonstrating the positive impacts of SR performance without using classifier-free guidance.
Thus, our approach directly uses the SR sampling result without applying classifier-free guidance.
Inspired from Fig.~\ref{fig:sr_compute}, where the scaled downstream LDMs have significant performance difference in 50-step sampling, we investigate sampling efficiency from two different aspects, \ie, fewer sampling steps $[4, 20]$ and more sampling steps $(20, 250]$. 
As shown in the left part of Fig.~\ref{fig:sroptiamlrules}, the scaling sampling-efficiency still holds in the SR tasks when the number of sampling steps is less than or equal to 20 steps.
Beyond this threshold, however, larger models demonstrate greater sampling-efficiency than smaller models, as illustrated in the right part of Fig.~\ref{fig:sroptiamlrules}.
This observation suggests the consistent sampling efficiency of scaled models on fewer sampling steps from text-to-image generation to super-resolution tasks.

\begin{figure}[!t]
    \centering
    \def\xwidth{.31\linewidth}
    \includegraphics[height=\xwidth]{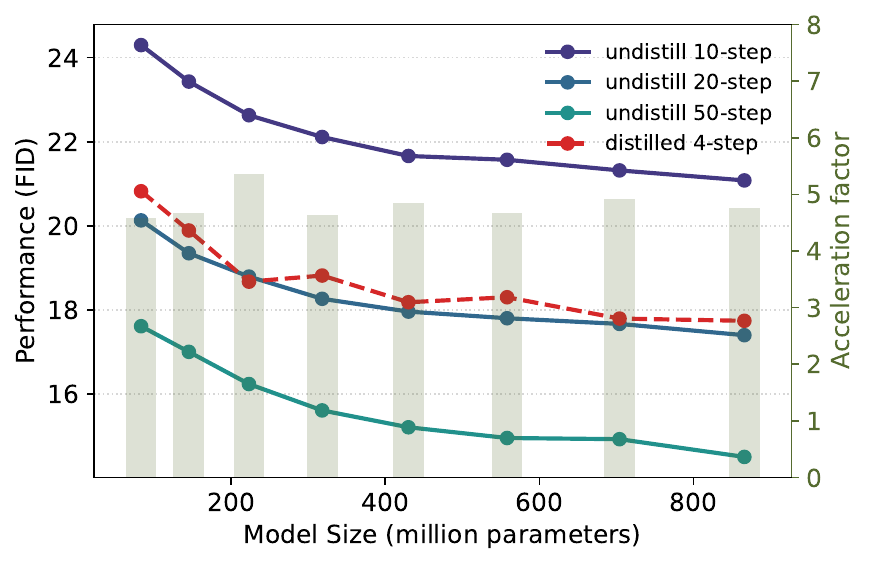}
    \includegraphics[height=\xwidth]{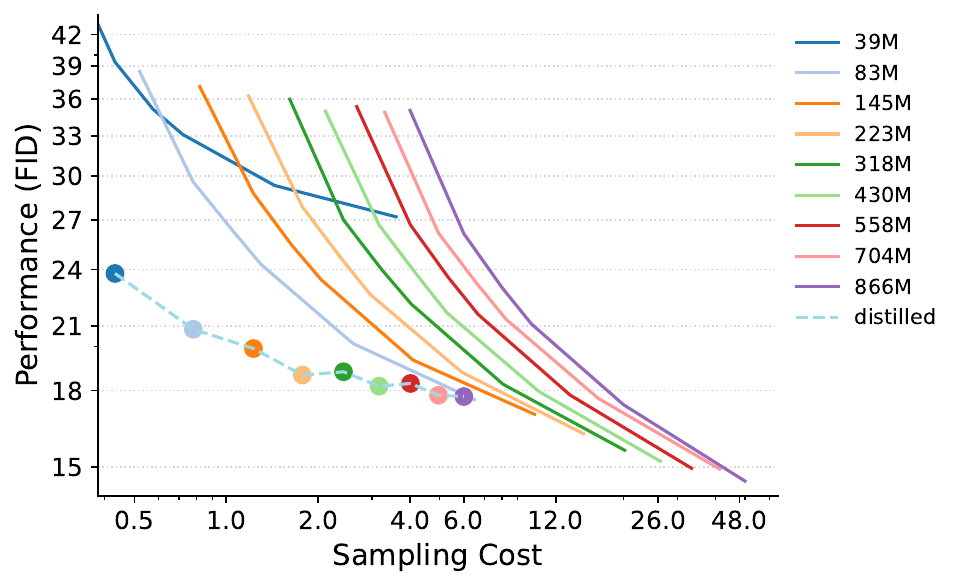}
    \vspace{-.5\baselineskip}
    \caption{Distillation improves text-to-image performance and scalability. \emph{Left:} Distilled Latent Diffusion Models (LDMs) consistently exhibit lower (better) FID scores compared to their undistilled counterparts across varying model sizes. The consistent acceleration factor (approx. $5\times$) indicates that the benefits of distillation scale well with model size.
    \emph{Right:} Distilled models using only 4 sampling steps achieve FID scores comparable to undistilled models using significantly more steps. Interestingly, at a sampling cost of 7, the distilled \texttt{866M} model performs similarly to the smaller, undistilled \texttt{83M} model, suggesting improved efficiency. 
    }
    \label{fig:distll}
    \vspace{-1\baselineskip}
\end{figure}

\subsection{Scaling sampling-efficiency in distilled LDMs.}
\label{sec:scalingdistill}

We have featured the scaling sampling-efficiency of latent diffusion models, which demonstrates that smaller model sizes exhibit higher sampling efficiency.
A notable caveat, however, is that smaller models typically imply reduced modeling capability. This poses a challenge for recent diffusion distillation methods~\citep{luhman2021knowledge, salimans2022progressive, song2023consistency, sauer2023adversarial, gu2023boot, mei2023conditional,luo2023latent, lin2024sdxl} that heavily depend on modeling capability.
One might expect a contradictory conclusion and believe the distilled large models sample faster than distilled small models.
In order to demonstrate the sampling efficiency of scaled models after distillation, we distill our previously scaled models with conditional consistency distillation~\citep{song2023consistency, mei2023conditional} on text-to-image data and compare those distilled models on their optimal performance.

To elaborate, we test all distilled models with the same 4-step sampling, which is shown to be able to achieve the best sampling performance; we then compare each distilled model with the undistilled one on the normalized sampling cost.
We follow the same practice discussed in Section~\ref{sec:optimalparams} for selecting the optimal CFG rate and compare them under the same relative inference cost.
The results shown in the left part of Fig.~\ref{fig:distll} demonstrate that distillation significantly improves the generative performance for all models in 4-step sampling, with FID improvements across the board.
By comparing these distilled models with the undistilled models in the right part of Fig.~\ref{fig:distll}, we demonstrate that distilled models outperform undistilled models at the same sampling cost.
However, at the specific sampling cost, \ie, sampling cost $\approx \texttt{8}$, the smaller undistilled 83M model still achieves similar performance to the larger distilled 866M model.
The observation further supports our proposed scaling sampling-efficiency after diffusion distillation.

\section{Conclusion}
In this paper, we investigated scaling properties of Latent Diffusion Models (LDMs), specifically through scaling model size from 39 million to 5 billion parameters.
We trained these scaled models from scratch on a web-scale text-to-image dataset and then finetuned the pretrained models for downstream tasks.
Our findings unveil that, under identical sampling costs, smaller models frequently outperform larger models, suggesting a promising direction for accelerating LDMs in terms of model size.
We further show that the sampling efficiency is consistent in multiple axes. For example, it is invariant to various diffusion samplers (stochastic and deterministic), and also holds true for distilled models.
We believe this analysis of scaling sampling efficiency would  be instrumental in guiding future developments of LDMs, specifically for balancing model size against performance and efficiency in a broad spectrum of practical applications.

\paragraph{Limitations and future work.}
This work utilizes visual quality inspection alongside established metrics like FID and CLIP scores. We opted to avoid human evaluations due to the immense number of different combinations needed for the more than 1000 variants considered in this study.
However, it is important to acknowledge the potential discrepancy between visual quality and quantitative metrics, which is actively discussed in recent works~\citep{zhang2021cross,jayasumana2023rethinking,cho2023davidsonian}.

Claims regarding the scalability of latent diffusion models are made specifically for the particular model family studied in this work~\citep{rombach2022high}. Extending this analysis to other model families, particularly those incorporating transformer-based backbones such as DiT~\citep{peebles2023scalable, mei2023t1}, SiT~\citep{ma2024sit}, MM-DiT~\citep{esser2024scaling}, and DiS~\citep{fei2024scalable}, and cascaded diffusion models such as Imagen3~\citep{baldridge2024imagen} and Stable Cascade~\citep{pernias2023wurstchen}, would be a valuable direction for future research.

\section{Acknowledgments}
Vishal M. Patel was supported by NSF CAREER award 2045489.
We are grateful to Keren Ye, Jason Baldridge, Kelvin Chan for their valuable feedback. We also extend our gratitude to Shlomi Fruchter, Kevin Murphy, Mohammad Babaeizadeh, and Han Zhang for their instrumental
contributions in facilitating the initial implementation of the latent diffusion models.

\bibliography{main}
\bibliographystyle{tmlr}

\newpage
\appendix

\input{supp}

\end{document}

%% file: math_commands.tex
\usepackage{amsmath,amsfonts,bm}

\def\eqref#1{equation~\ref{#1}}

\def\1{\bm{1}}

\DeclareMathAlphabet{\mathsfit}{\encodingdefault}{\sfdefault}{m}{sl}
\SetMathAlphabet{\mathsfit}{bold}{\encodingdefault}{\sfdefault}{bx}{n}

%% file: supp.tex
\section{Scaling the text-to-image performance}
\label{supp:t2i}

\begin{figure}[ht]
    \centering
    \vspace{-1\baselineskip}
    \begin{subfigure}[b]{\textwidth}
    \centering
    \imageWithNote{0.19\textwidth}{\scriptsize \raggedleft \texttt{83M}}{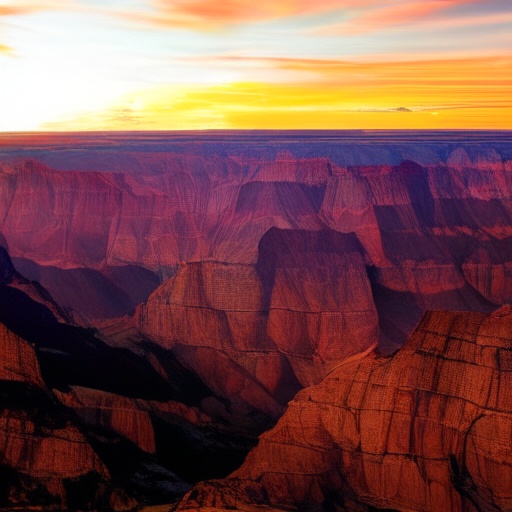}
    \imageWithNote{0.19\textwidth}{\scriptsize \raggedleft \texttt{145M}}{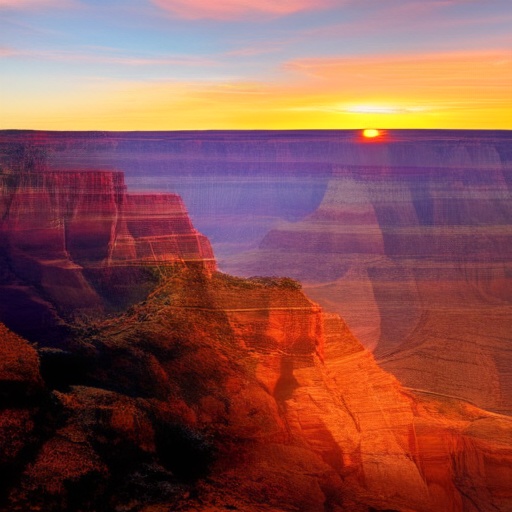}
    \imageWithNote{0.19\textwidth}{\scriptsize \raggedleft \texttt{223M}}{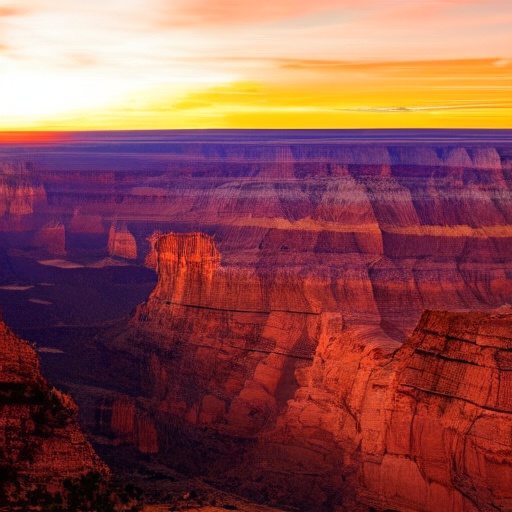}
    \imageWithNote{0.19\textwidth}{\scriptsize \raggedleft \texttt{318M}}{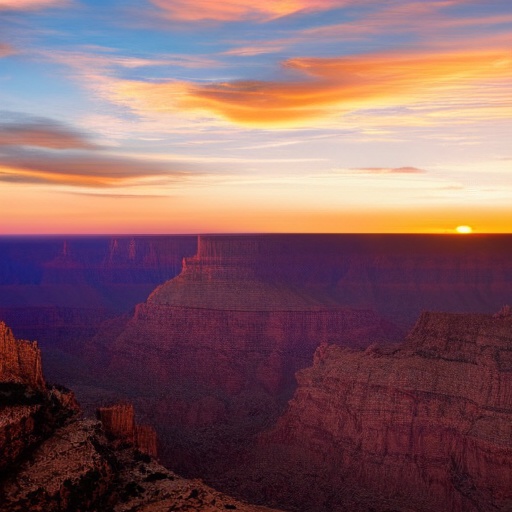}
    \imageWithNote{0.19\textwidth}{\scriptsize \raggedleft \texttt{430M}}{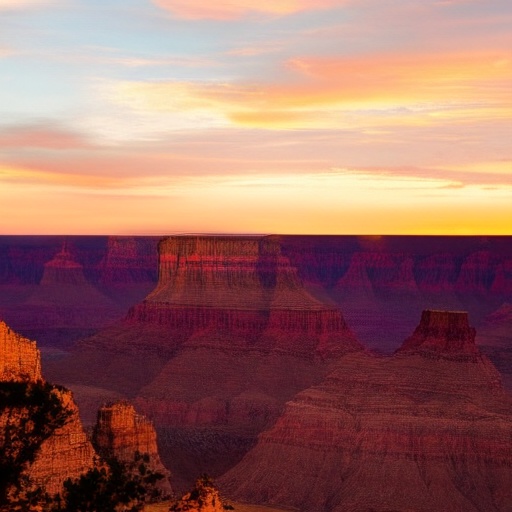}
    \imageWithNote{0.19\textwidth}{\scriptsize \raggedleft \texttt{558M}}{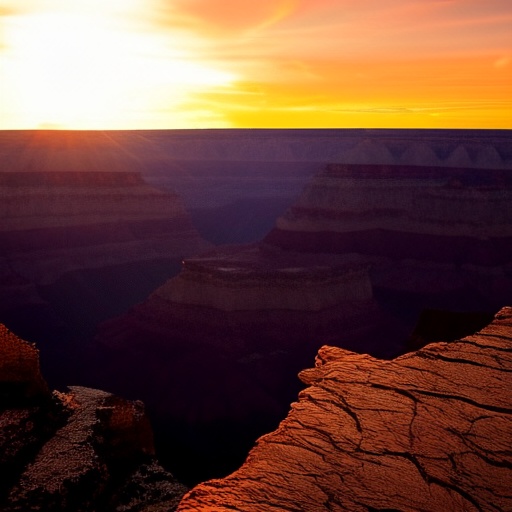}
    \imageWithNote{0.19\textwidth}{\scriptsize \raggedleft \texttt{704M}}{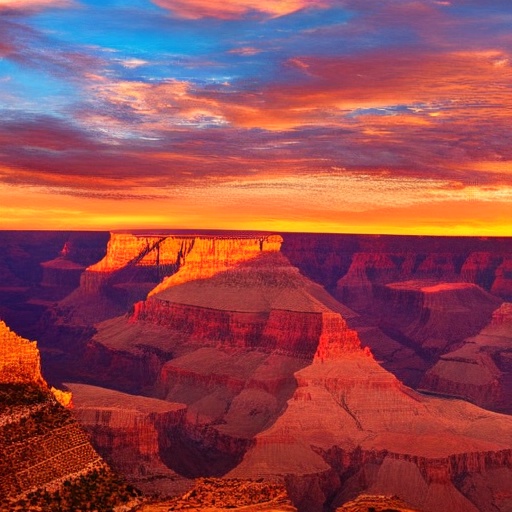}
    \imageWithNote{0.19\textwidth}{\scriptsize \raggedleft \texttt{866M}}{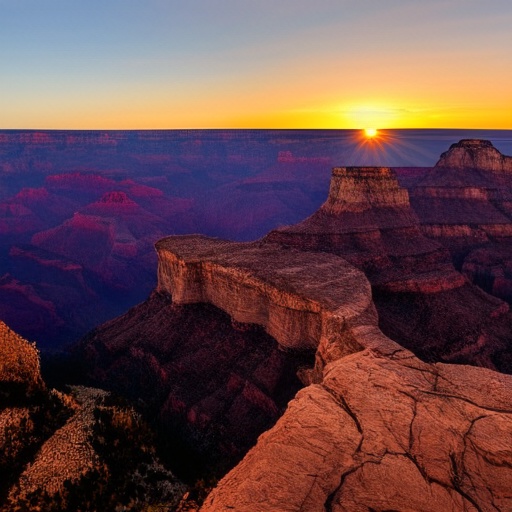}
    \imageWithNote{0.19\textwidth}{\scriptsize \raggedleft \texttt{2B}}{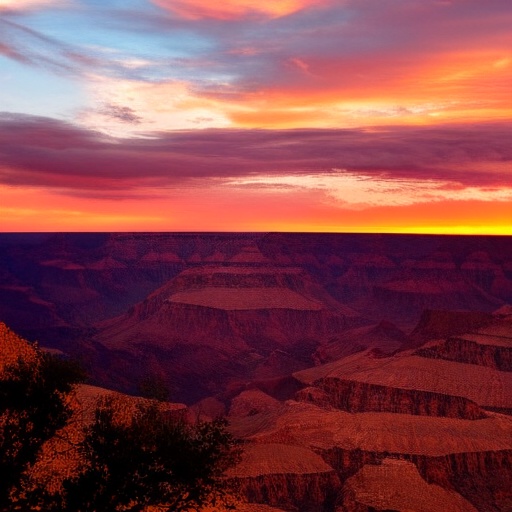}
    \imageWithNote{0.19\textwidth}{\scriptsize \raggedleft \texttt{5B}}{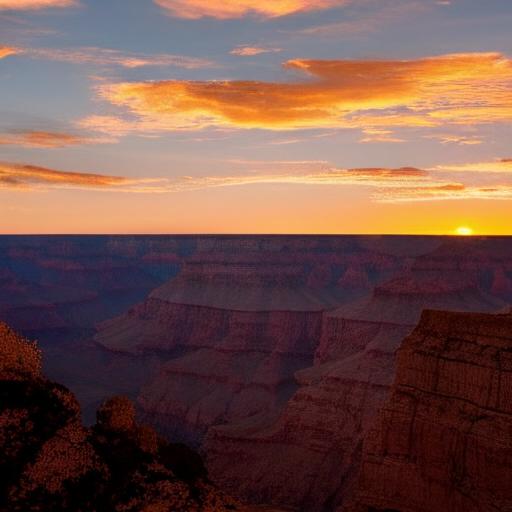}
    \caption{Prompt: \emph{``a professional photo of a sunset behind the grand canyon.''}}
    \end{subfigure}
    \begin{subfigure}[b]{\textwidth}
    \centering
    \imageWithNote{0.19\textwidth}{\scriptsize \raggedleft \texttt{83M}}{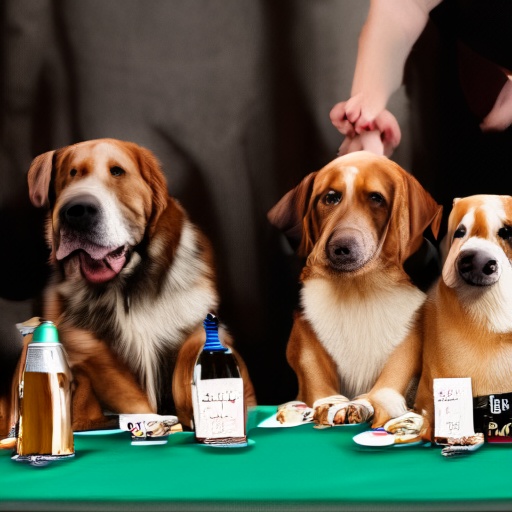}
    \imageWithNote{0.19\textwidth}{\scriptsize \raggedleft \texttt{145M}}{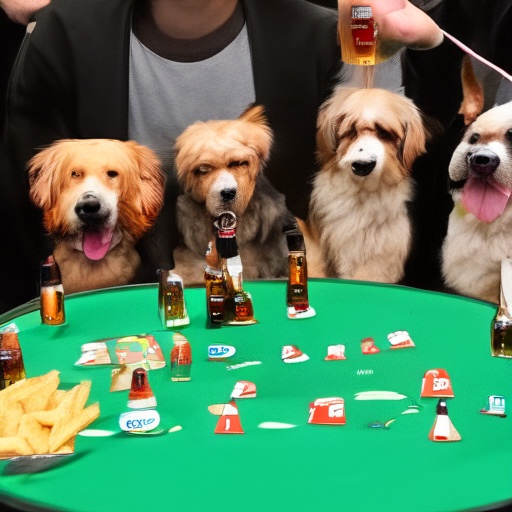}
    \imageWithNote{0.19\textwidth}{\scriptsize \raggedleft \texttt{223M}}{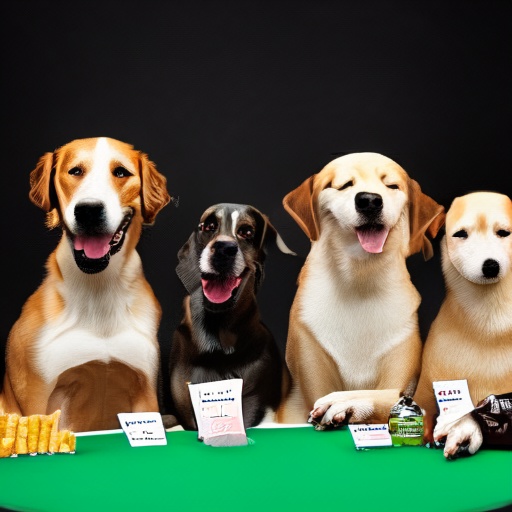}
    \imageWithNote{0.19\textwidth}{\scriptsize \raggedleft \texttt{318M}}{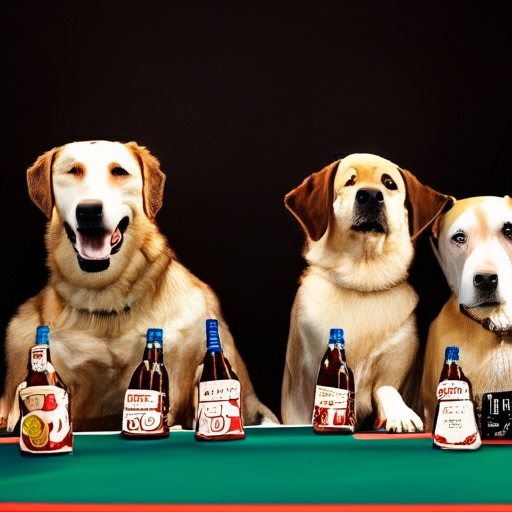}
    \imageWithNote{0.19\textwidth}{\scriptsize \raggedleft \texttt{430M}}{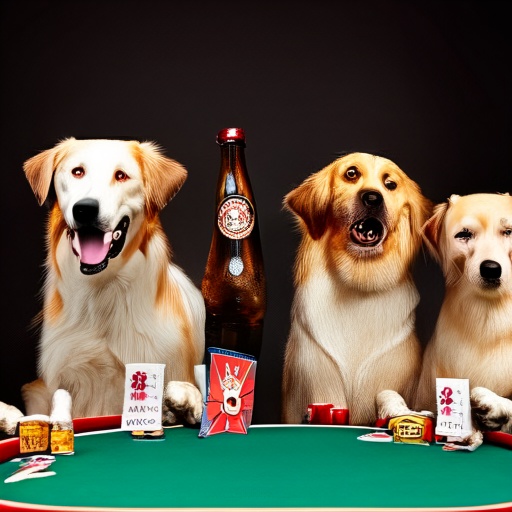}
    \imageWithNote{0.19\textwidth}{\scriptsize \raggedleft \texttt{558M}}{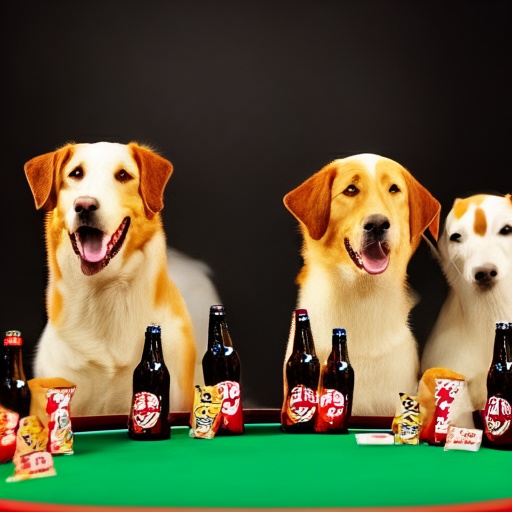}
    \imageWithNote{0.19\textwidth}{\scriptsize \raggedleft \texttt{704M}}{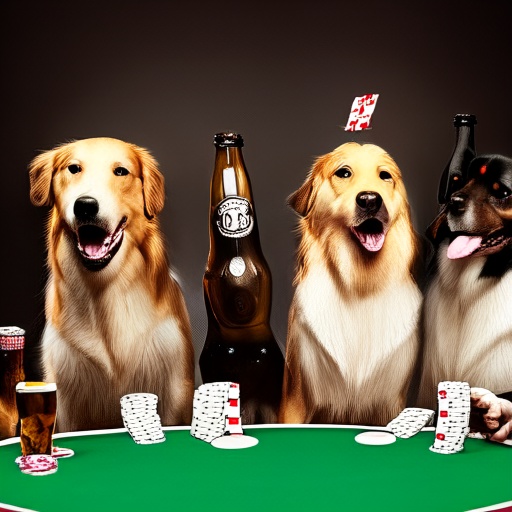}
    \imageWithNote{0.19\textwidth}{\scriptsize \raggedleft \texttt{866M}}{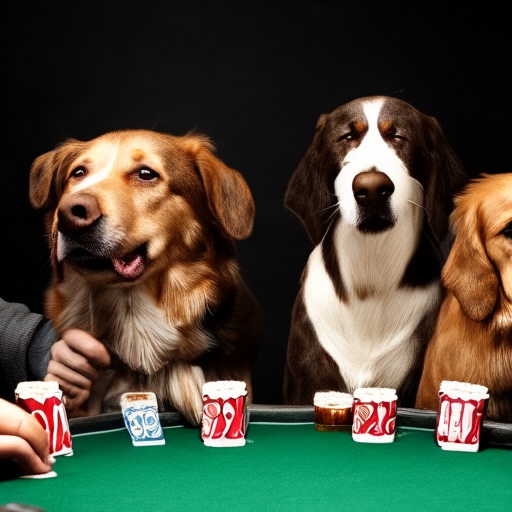}
    \imageWithNote{0.19\textwidth}{\scriptsize \raggedleft \texttt{2B}}{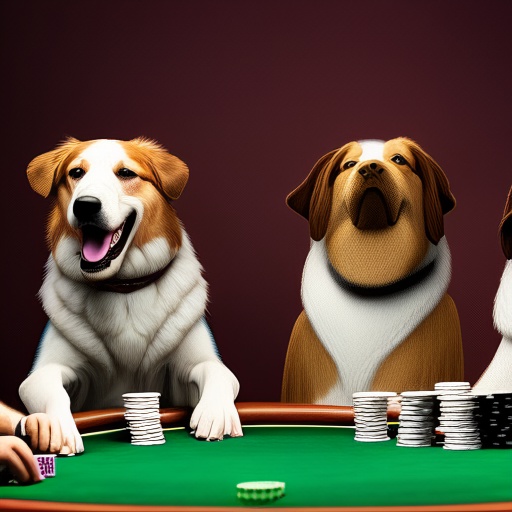}
    \imageWithNote{0.19\textwidth}{\scriptsize \raggedleft \texttt{5B}}{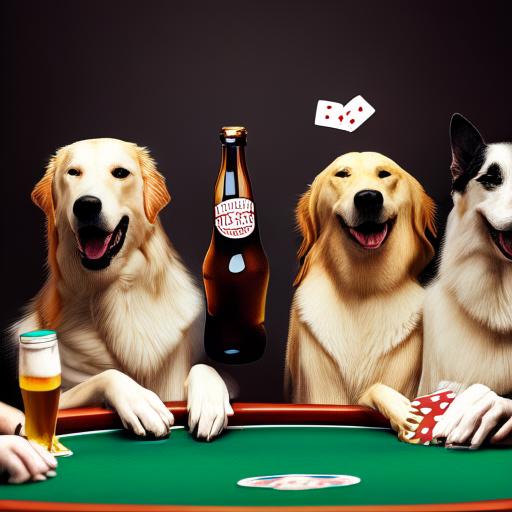}
    \caption{Prompt: \emph{``Dogs sitting around a poker table with beer bottles and chips. Their hands are holding cards.''}}
    \end{subfigure}
    \begin{subfigure}[b]{\textwidth}
    \centering
    \imageWithNote{0.19\textwidth}{\scriptsize \raggedleft \texttt{83M}}{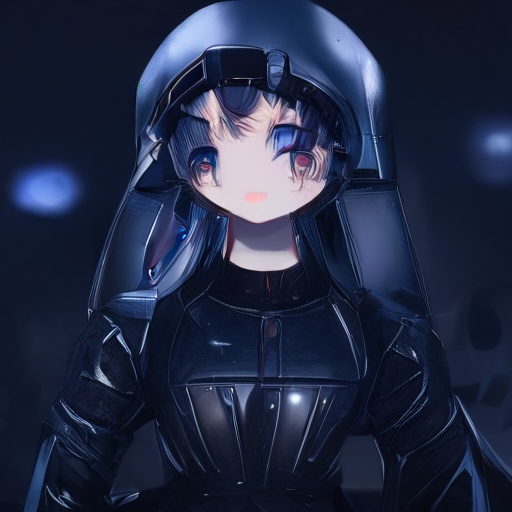}
    \imageWithNote{0.19\textwidth}{\scriptsize \raggedleft \texttt{145M}}{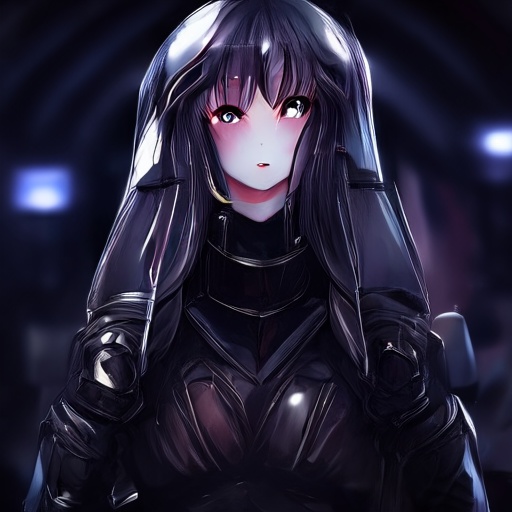}
    \imageWithNote{0.19\textwidth}{\scriptsize \raggedleft \texttt{223M}}{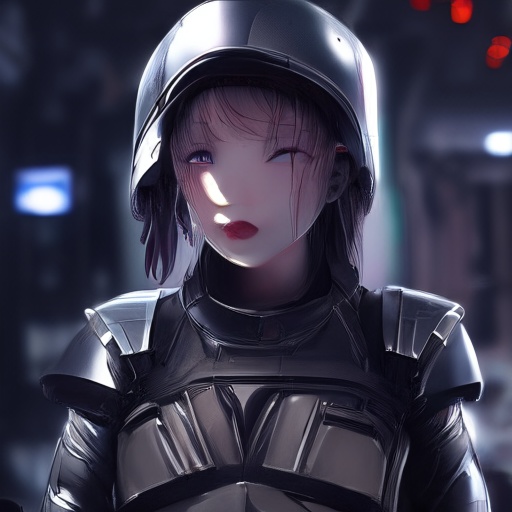}
    \imageWithNote{0.19\textwidth}{\scriptsize \raggedleft \texttt{318M}}{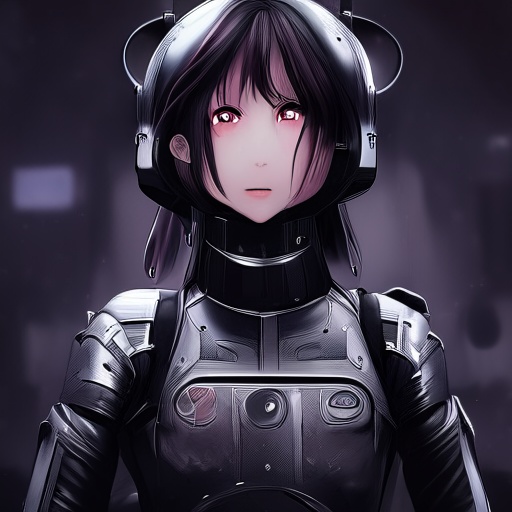}
    \imageWithNote{0.19\textwidth}{\scriptsize \raggedleft \texttt{430M}}{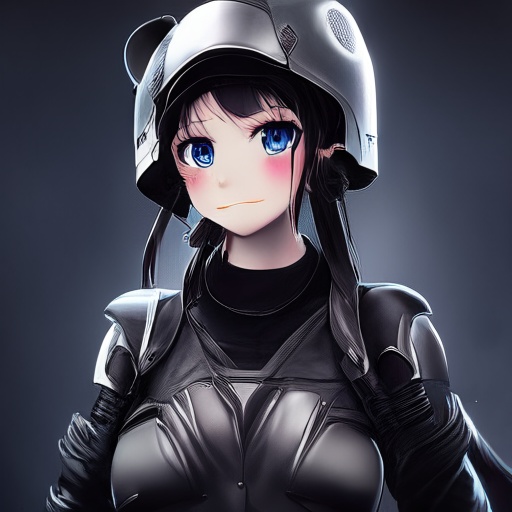}
    \imageWithNote{0.19\textwidth}{\scriptsize \raggedleft \texttt{558M}}{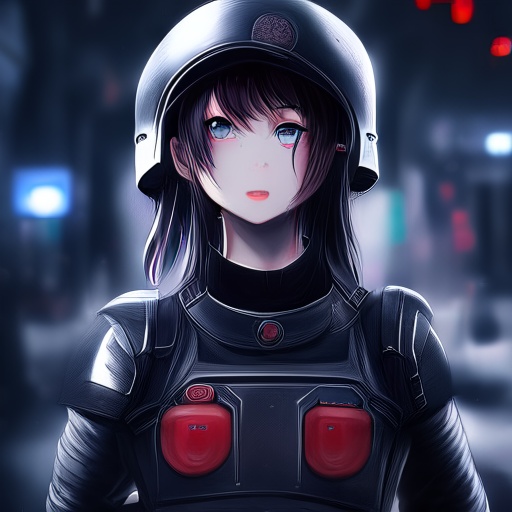}
    \imageWithNote{0.19\textwidth}{\scriptsize \raggedleft \texttt{704M}}{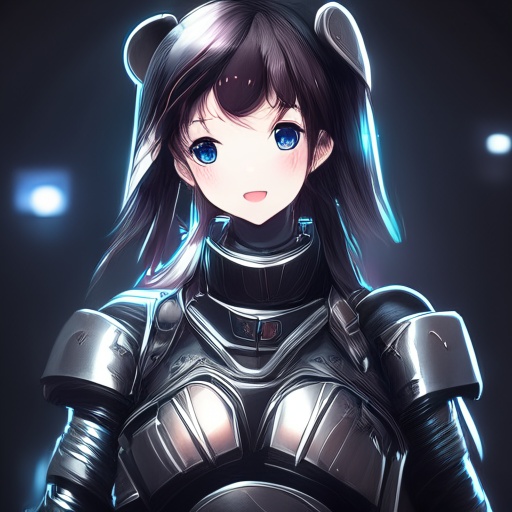}
    \imageWithNote{0.19\textwidth}{\scriptsize \raggedleft \texttt{866M}}{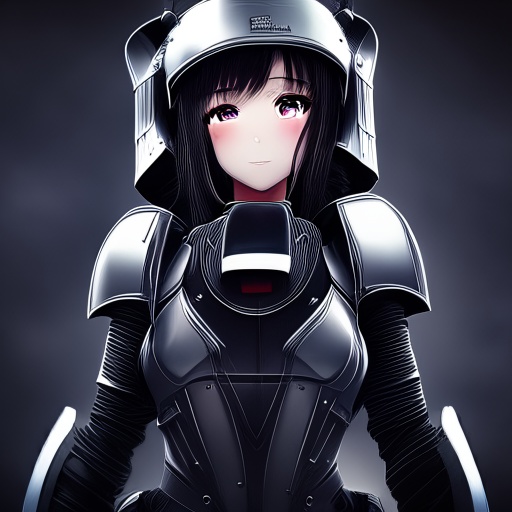}
    \imageWithNote{0.19\textwidth}{\scriptsize \raggedleft \texttt{2B}}{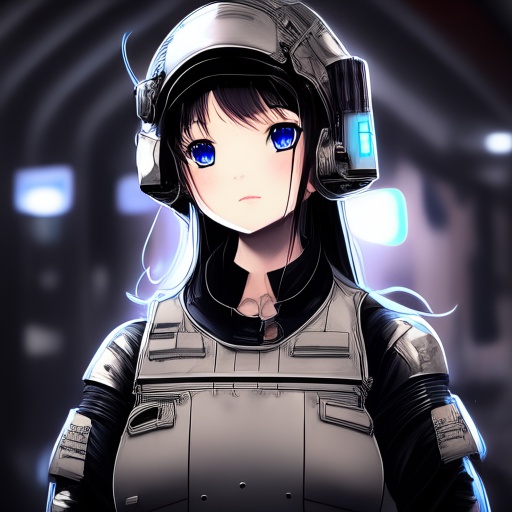}
    \imageWithNote{0.19\textwidth}{\scriptsize \raggedleft \texttt{5B}}{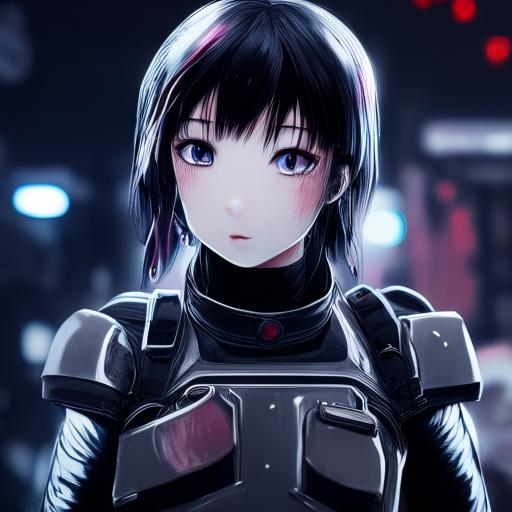}
    \caption{Prompt: \emph{`Portrait of anime girl in mechanic armor in night Tokyo.''}}
    \end{subfigure}
    \caption{Text-to-image results from our scaled LDMs (\texttt{83M} - \texttt{5B}), highlighting the improvement in visual quality with increased model size.}
    \label{fig:suppvisualcompare1}
    \vspace{-6\baselineskip}
\end{figure}

\begin{figure}[htbp]
    \centering
    \vspace{-1\baselineskip}
    \begin{subfigure}[b]{\textwidth}
    \centering
    \imageWithNote{0.19\textwidth}{\scriptsize \raggedleft \texttt{83M}}{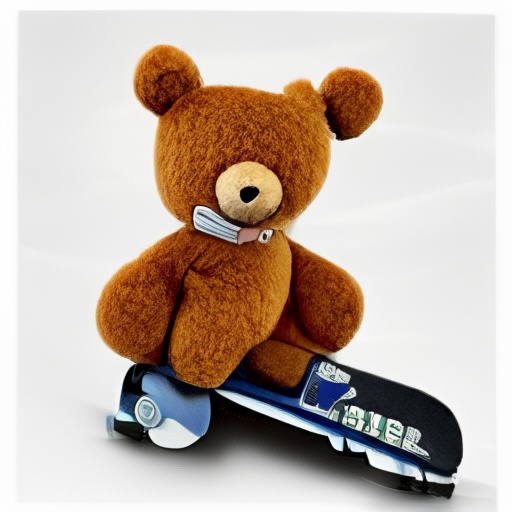}
    \imageWithNote{0.19\textwidth}{\scriptsize \raggedleft \texttt{145M}}{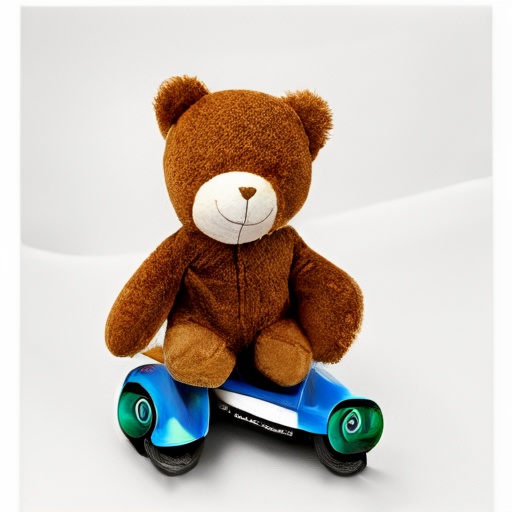}
    \imageWithNote{0.19\textwidth}{\scriptsize \raggedleft \texttt{223M}}{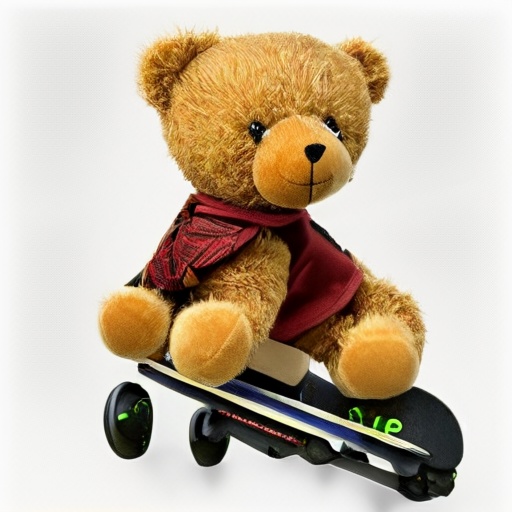}
    \imageWithNote{0.19\textwidth}{\scriptsize \raggedleft \texttt{318M}}{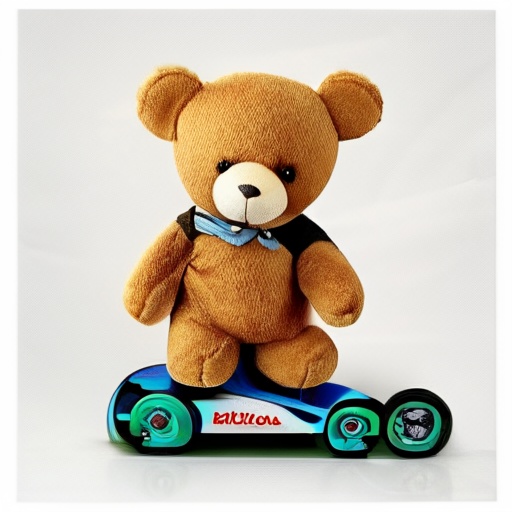}
    \imageWithNote{0.19\textwidth}{\scriptsize \raggedleft \texttt{430M}}{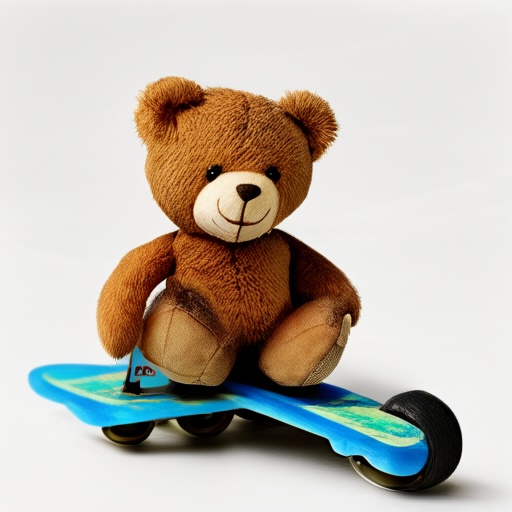}
    \imageWithNote{0.19\textwidth}{\scriptsize \raggedleft \texttt{558M}}{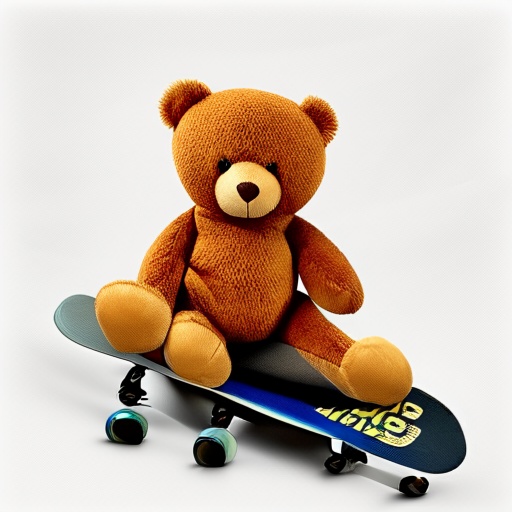}
    \imageWithNote{0.19\textwidth}{\scriptsize \raggedleft \texttt{704M}}{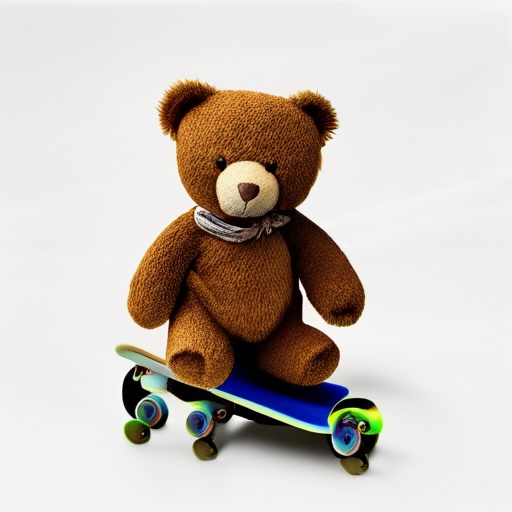}
    \imageWithNote{0.19\textwidth}{\scriptsize \raggedleft \texttt{866M}}{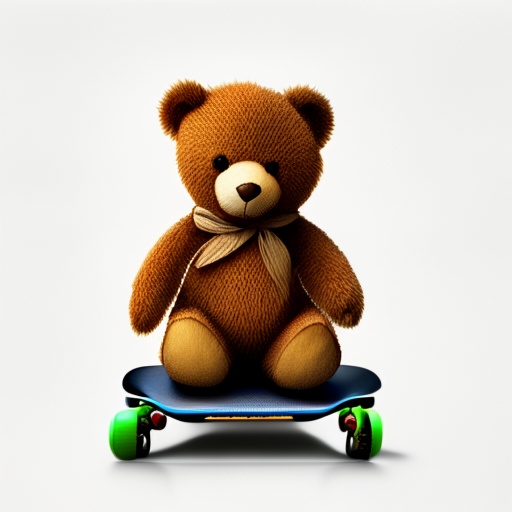}
    \imageWithNote{0.19\textwidth}{\scriptsize \raggedleft \texttt{2B}}{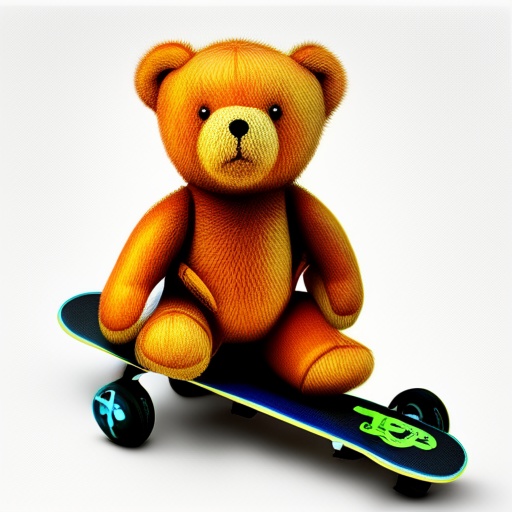}
    \imageWithNote{0.19\textwidth}{\scriptsize \raggedleft \texttt{5B}}{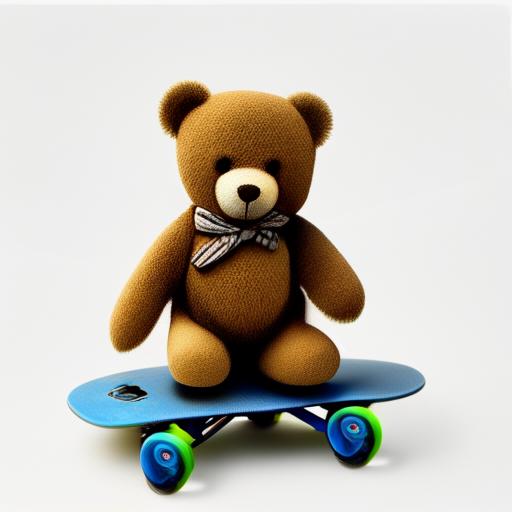}
    \caption{Prompt: \emph{``a teddy bear on a skateboard.''}}
    \end{subfigure}
    
    \begin{subfigure}[b]{\textwidth}
    \centering
    \imageWithNote{0.19\textwidth}{\scriptsize \raggedleft \texttt{83M}}{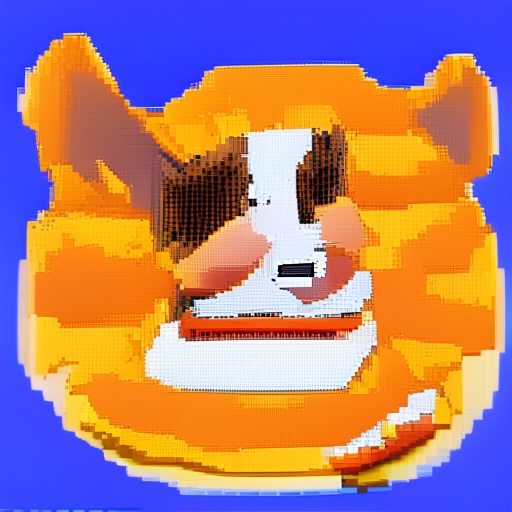}
    \imageWithNote{0.19\textwidth}{\scriptsize \raggedleft \texttt{145M}}{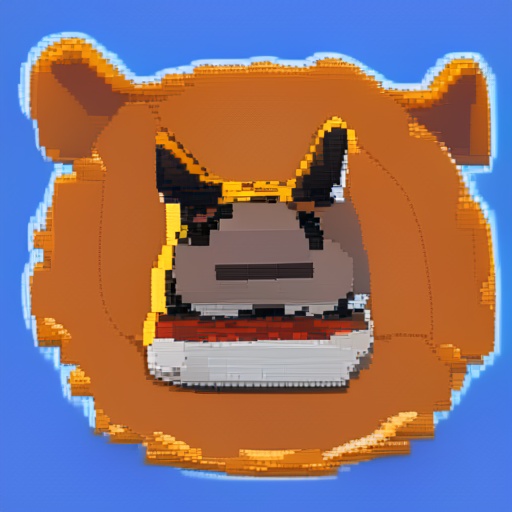}
    \imageWithNote{0.19\textwidth}{\scriptsize \raggedleft \texttt{223M}}{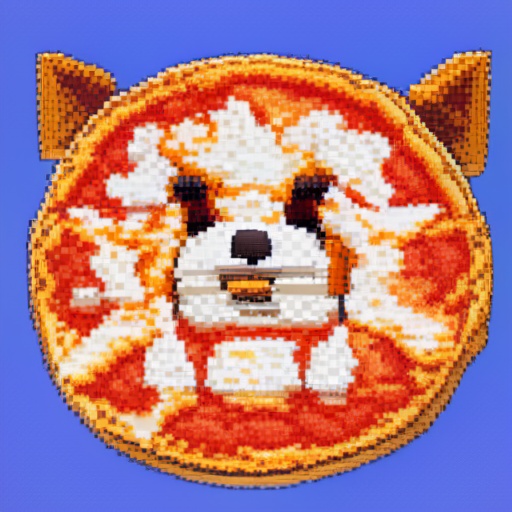}
    \imageWithNote{0.19\textwidth}{\scriptsize \raggedleft \texttt{318M}}{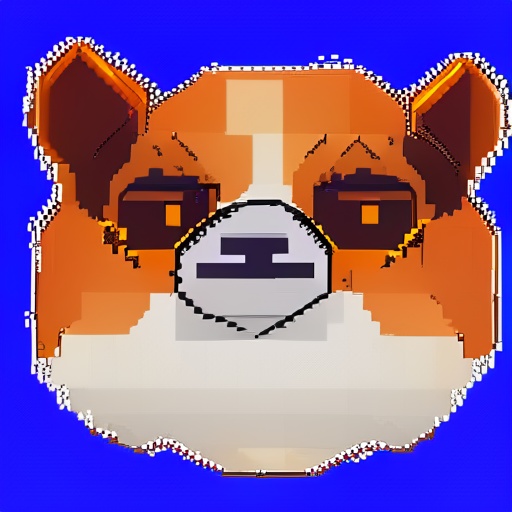}
    \imageWithNote{0.19\textwidth}{\scriptsize \raggedleft \texttt{430M}}{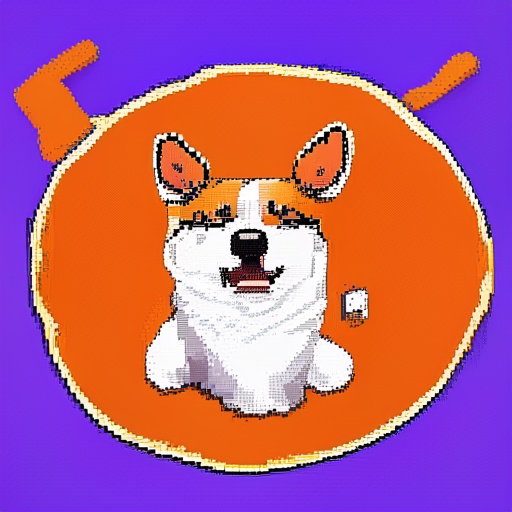}
    \imageWithNote{0.19\textwidth}{\scriptsize \raggedleft \texttt{558M}}{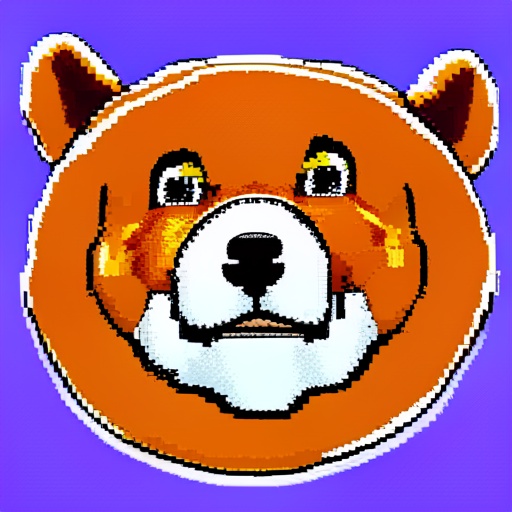}
    \imageWithNote{0.19\textwidth}{\scriptsize \raggedleft \texttt{704M}}{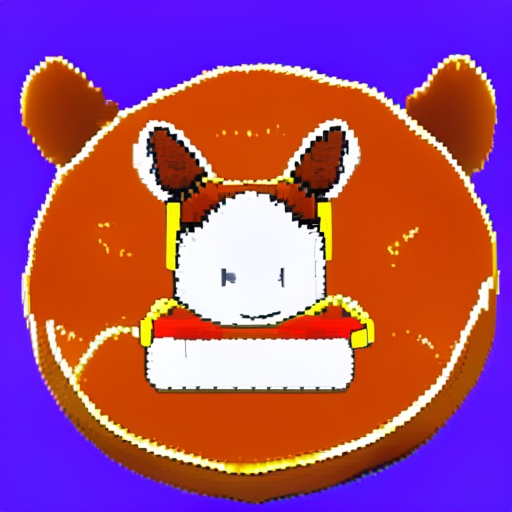}
    \imageWithNote{0.19\textwidth}{\scriptsize \raggedleft \texttt{866M}}{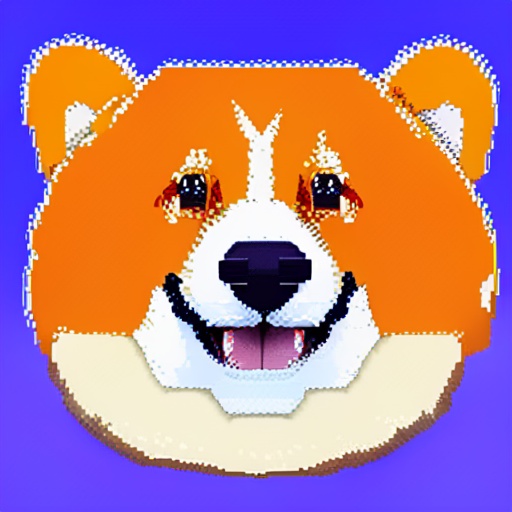}
    \imageWithNote{0.19\textwidth}{\scriptsize \raggedleft \texttt{2B}}{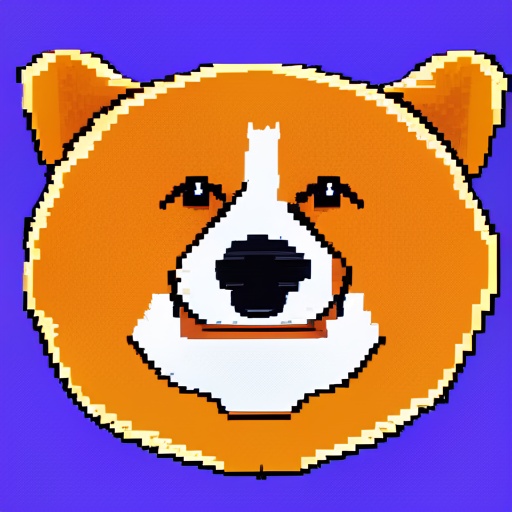}
    \imageWithNote{0.19\textwidth}{\scriptsize \raggedleft \texttt{5B}}{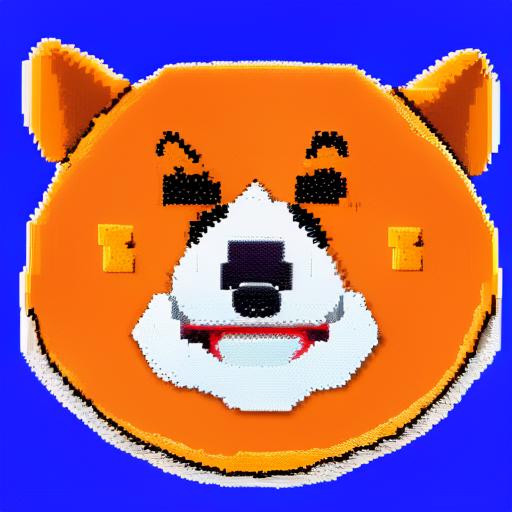}
    \caption{Prompt: \emph{``a pixel art corgi pizza.''}}
    \end{subfigure}

    \begin{subfigure}[b]{\textwidth}
    \centering
    \imageWithNote{0.19\textwidth}{\scriptsize \raggedleft \texttt{83M}}{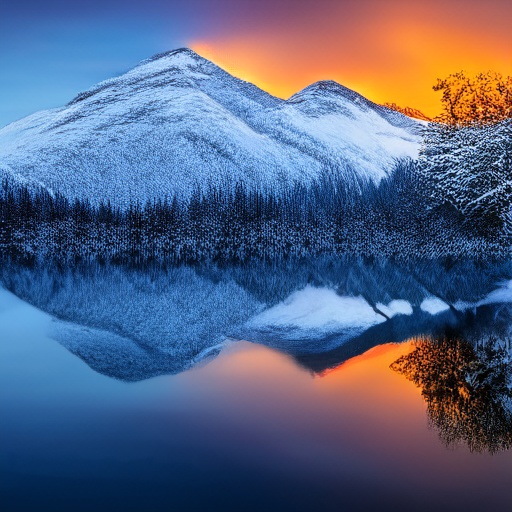}
    \imageWithNote{0.19\textwidth}{\scriptsize \raggedleft \texttt{145M}}{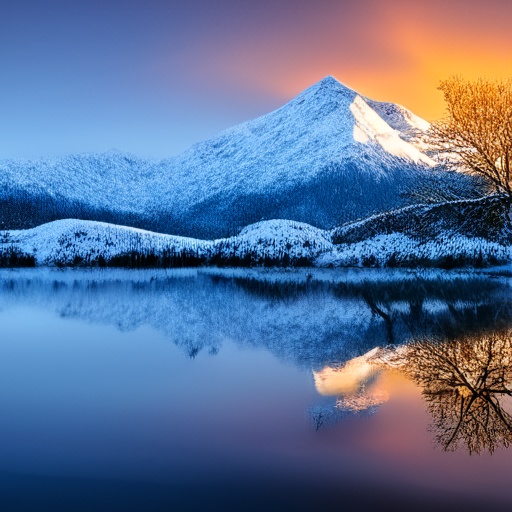}
    \imageWithNote{0.19\textwidth}{\scriptsize \raggedleft \texttt{223M}}{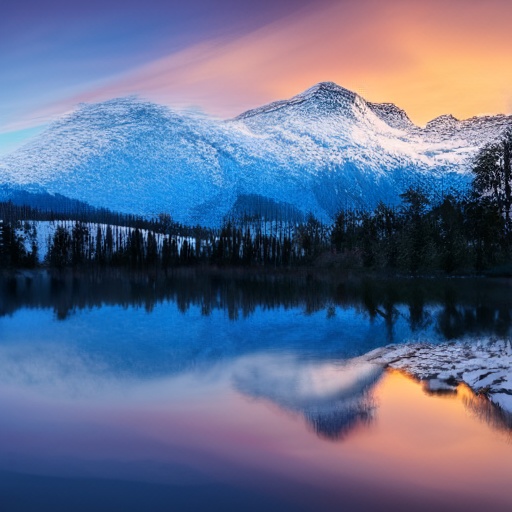}
    \imageWithNote{0.19\textwidth}{\scriptsize \raggedleft \texttt{318M}}{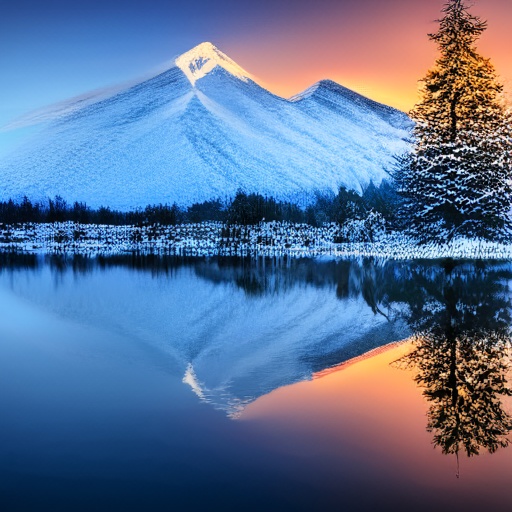}
    \imageWithNote{0.19\textwidth}{\scriptsize \raggedleft \texttt{430M}}{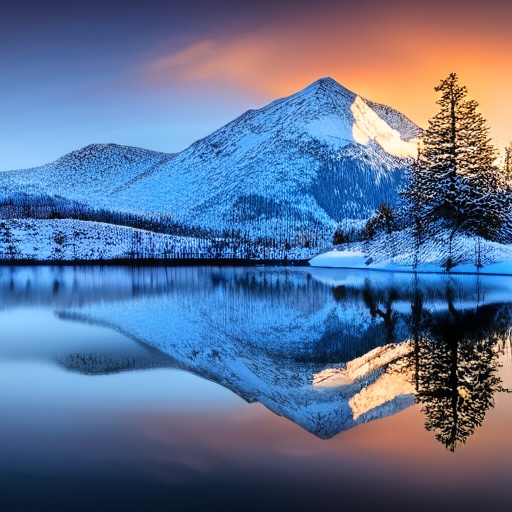}
    \imageWithNote{0.19\textwidth}{\scriptsize \raggedleft \texttt{558M}}{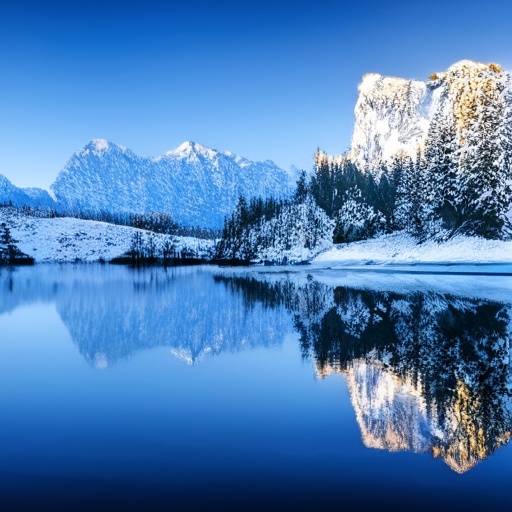}
    \imageWithNote{0.19\textwidth}{\scriptsize \raggedleft \texttt{704M}}{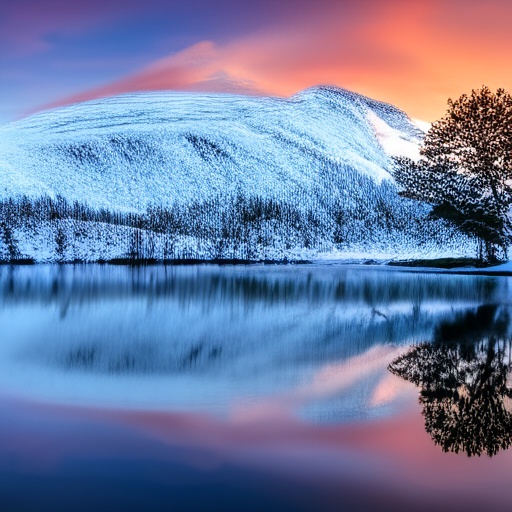}
    \imageWithNote{0.19\textwidth}{\scriptsize \raggedleft \texttt{866M}}{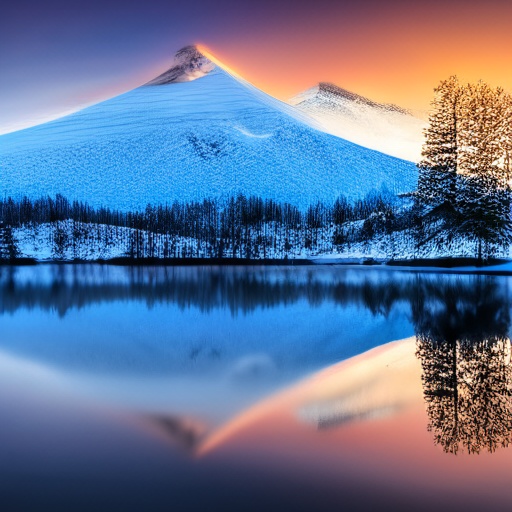}
    \imageWithNote{0.19\textwidth}{\scriptsize \raggedleft \texttt{2B}}{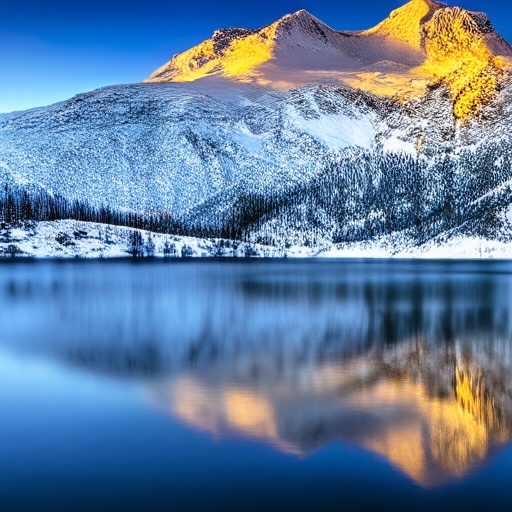}
    \imageWithNote{0.19\textwidth}{\scriptsize \raggedleft \texttt{5B}}{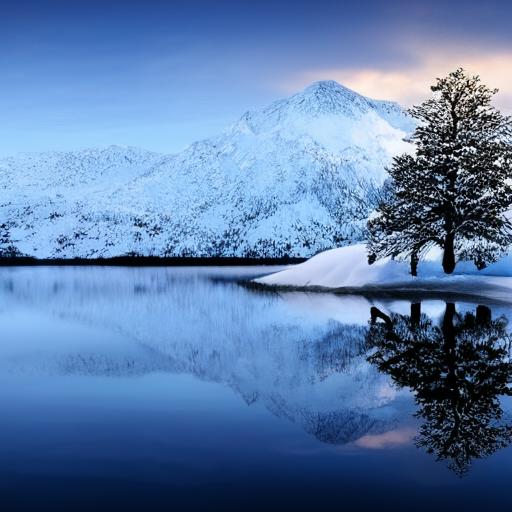}
    \caption{Prompt: \emph{``Snow mountain and tree reflection in the lake.''}}
    \end{subfigure}
    \caption{Text-to-image results from our scaled LDMs (\texttt{83M} - \texttt{5B}), highlighting the improvement in visual quality with increased model size.}
    \label{fig:suppvisualcompare2}
    \vspace{-2\baselineskip}
\end{figure}

\begin{figure}[htbp]
    \centering
    \vspace{-1\baselineskip}
    \begin{subfigure}[b]{\textwidth}
    \centering
    \imageWithNote{0.19\textwidth}{\scriptsize \raggedleft \texttt{83M}}{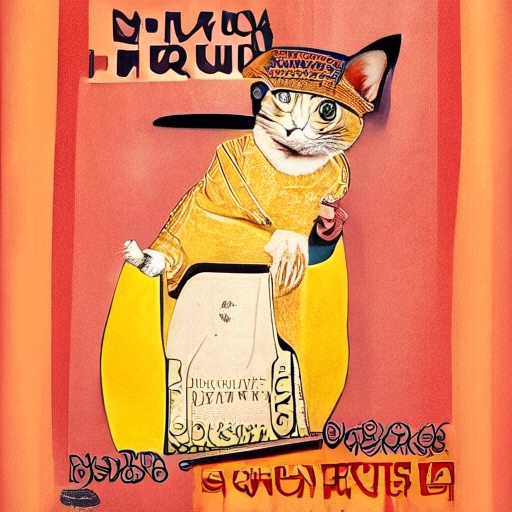}
    \imageWithNote{0.19\textwidth}{\scriptsize \raggedleft \texttt{145M}}{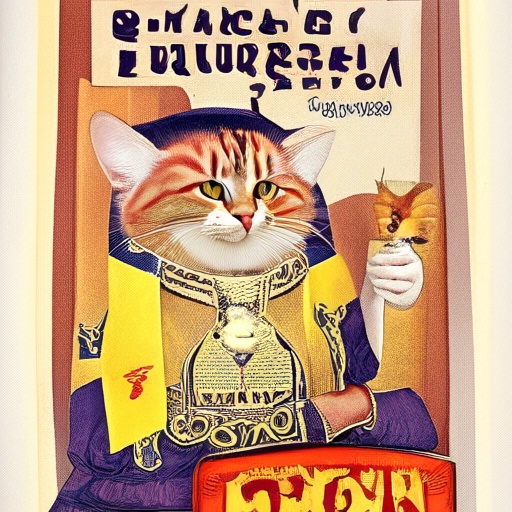}
    \imageWithNote{0.19\textwidth}{\scriptsize \raggedleft \texttt{223M}}{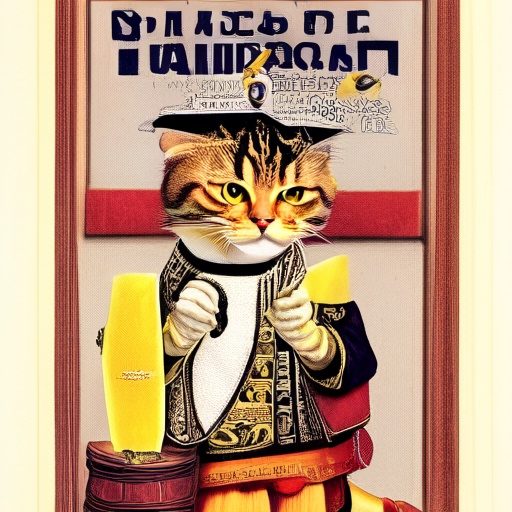}
    \imageWithNote{0.19\textwidth}{\scriptsize \raggedleft \texttt{318M}}{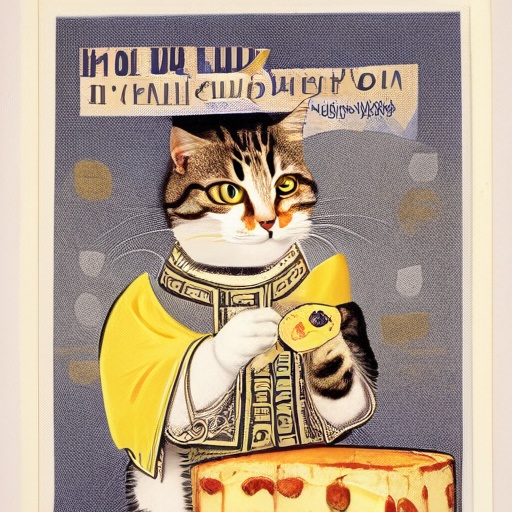}
    \imageWithNote{0.19\textwidth}{\scriptsize \raggedleft \texttt{430M}}{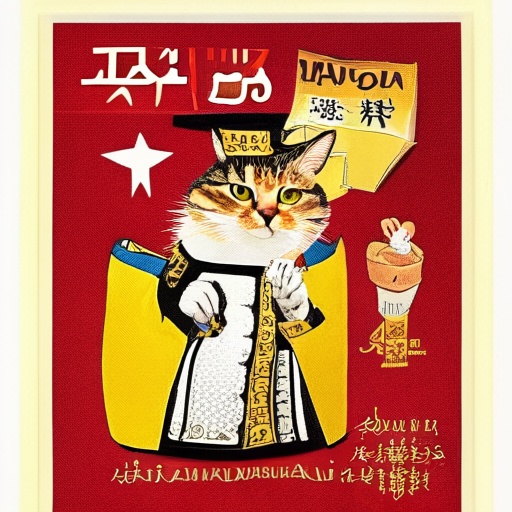}
    \imageWithNote{0.19\textwidth}{\scriptsize \raggedleft \texttt{558M}}{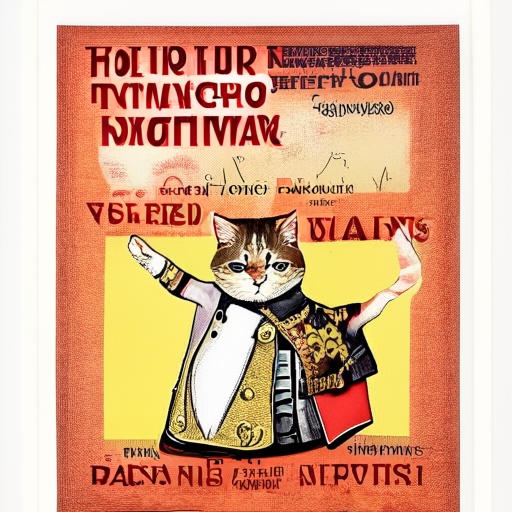}
    \imageWithNote{0.19\textwidth}{\scriptsize \raggedleft \texttt{704M}}{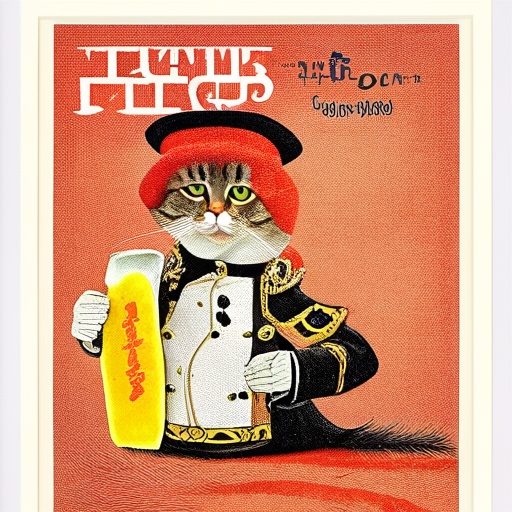}
    \imageWithNote{0.19\textwidth}{\scriptsize \raggedleft \texttt{866M}}{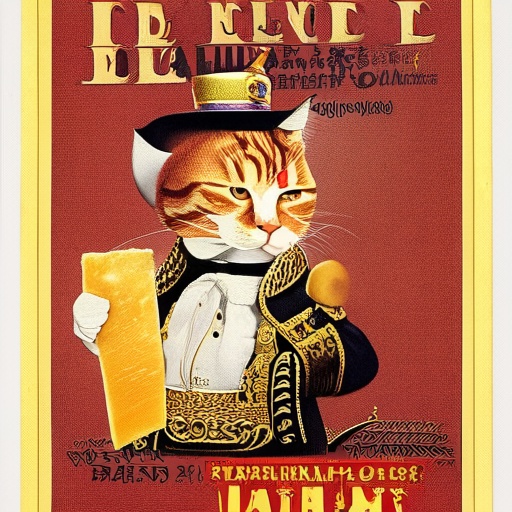}
    \imageWithNote{0.19\textwidth}{\scriptsize \raggedleft \texttt{2B}}{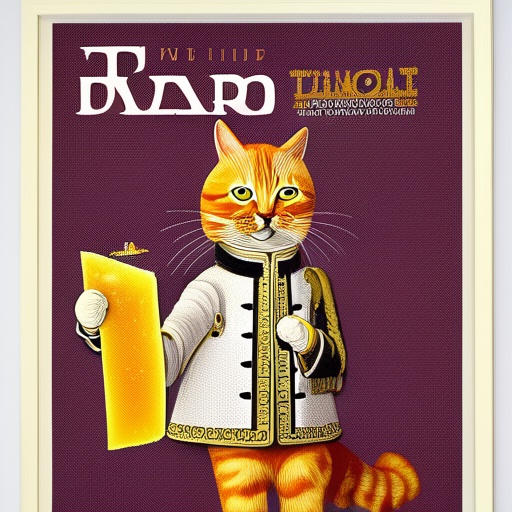}
    \imageWithNote{0.19\textwidth}{\scriptsize \raggedleft \texttt{5B}}{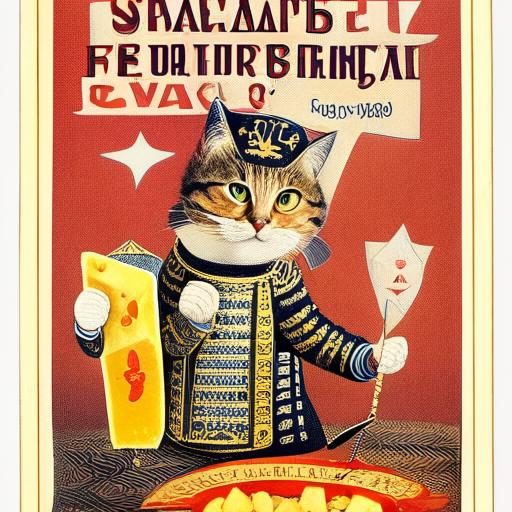}
    \caption{Prompt: \emph{``a propaganda poster depicting a cat dressed as french emperor napoleon holding a piece of cheese.''}}
    \end{subfigure}
    
    \begin{subfigure}[b]{\textwidth}
    \centering
    \imageWithNote{0.19\textwidth}{\scriptsize \raggedleft \texttt{83M}}{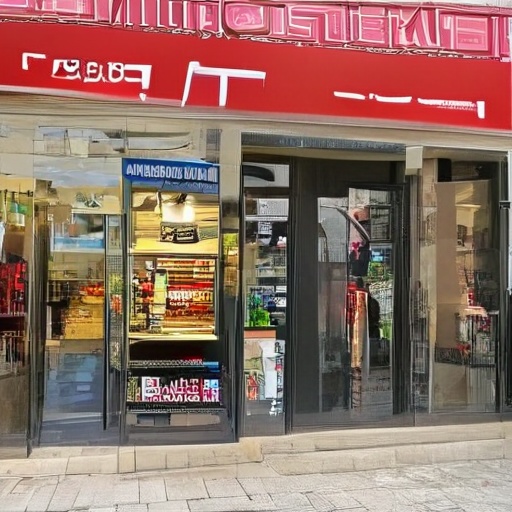}
    \imageWithNote{0.19\textwidth}{\scriptsize \raggedleft \texttt{145M}}{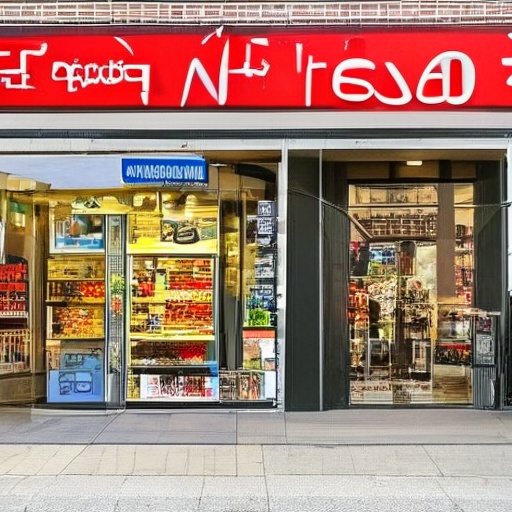}
    \imageWithNote{0.19\textwidth}{\scriptsize \raggedleft \texttt{223M}}{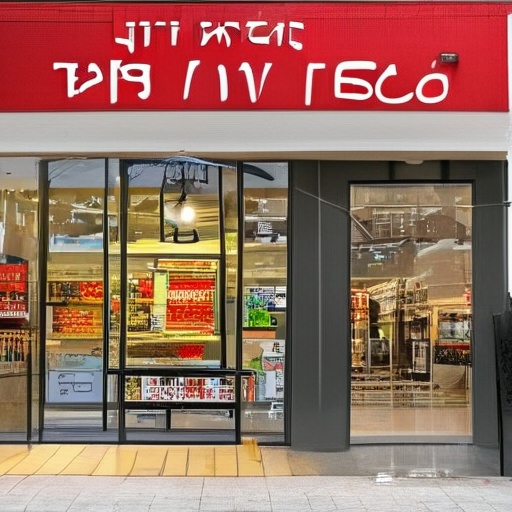}
    \imageWithNote{0.19\textwidth}{\scriptsize \raggedleft \texttt{318M}}{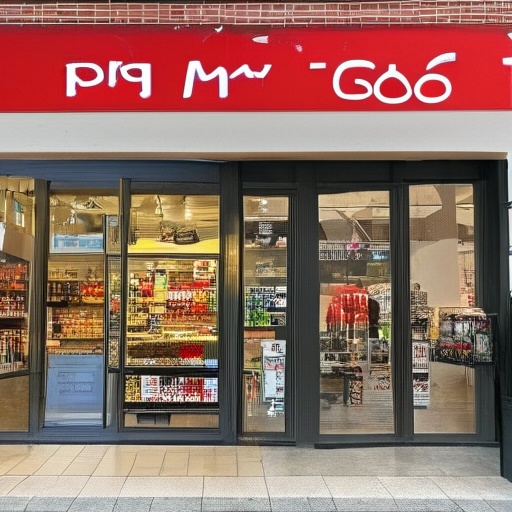}
    \imageWithNote{0.19\textwidth}{\scriptsize \raggedleft \texttt{430M}}{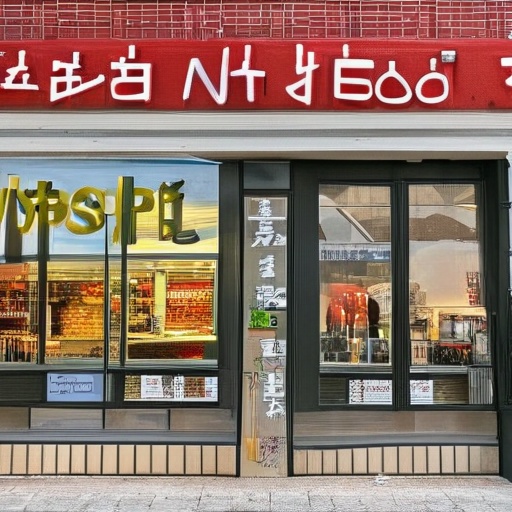}
    \imageWithNote{0.19\textwidth}{\scriptsize \raggedleft \texttt{558M}}{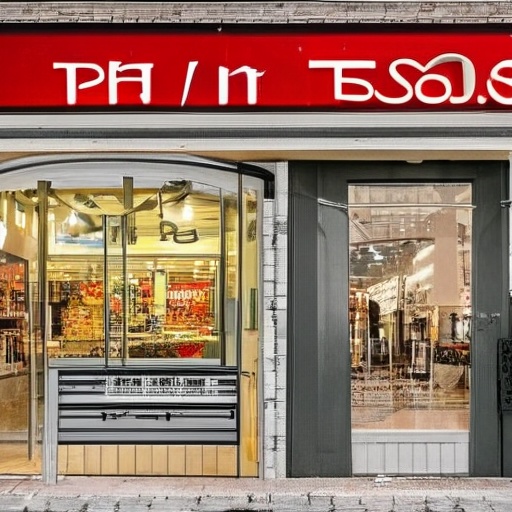}
    \imageWithNote{0.19\textwidth}{\scriptsize \raggedleft \texttt{704M}}{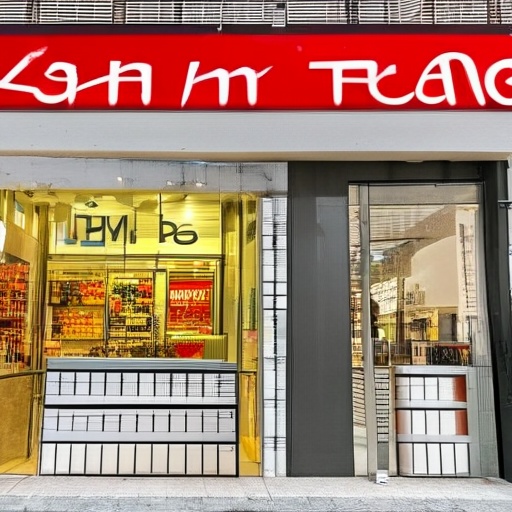}
    \imageWithNote{0.19\textwidth}{\scriptsize \raggedleft \texttt{866M}}{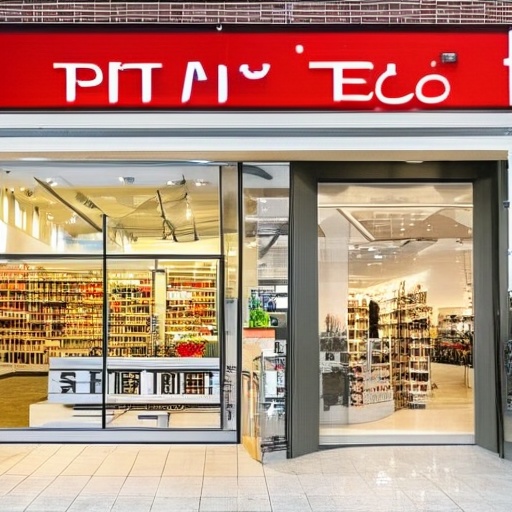}
    \imageWithNote{0.19\textwidth}{\scriptsize \raggedleft \texttt{2B}}{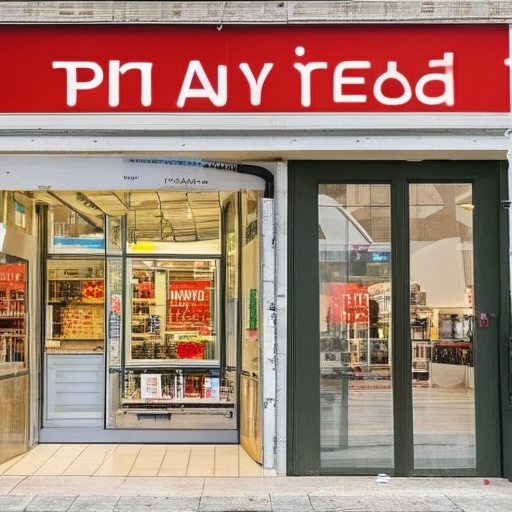}
    \imageWithNote{0.19\textwidth}{\scriptsize \raggedleft \texttt{5B}}{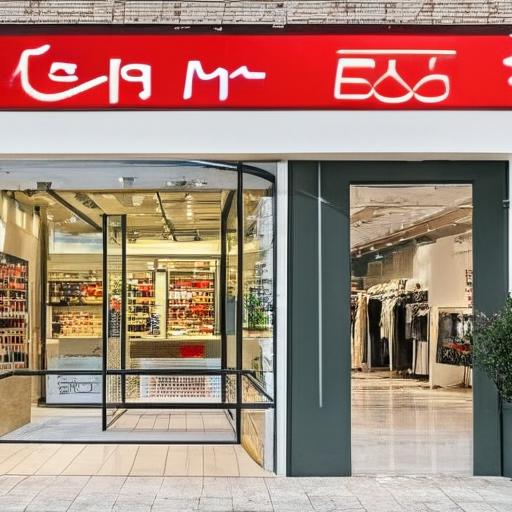}
    \caption{Prompt:  \emph{``a store front that has the word ‘LDMs’ written on it.''}}
    \end{subfigure}

    \begin{subfigure}[b]{\textwidth}
    \centering
    \imageWithNote{0.19\textwidth}{\scriptsize \raggedleft \texttt{83M}}{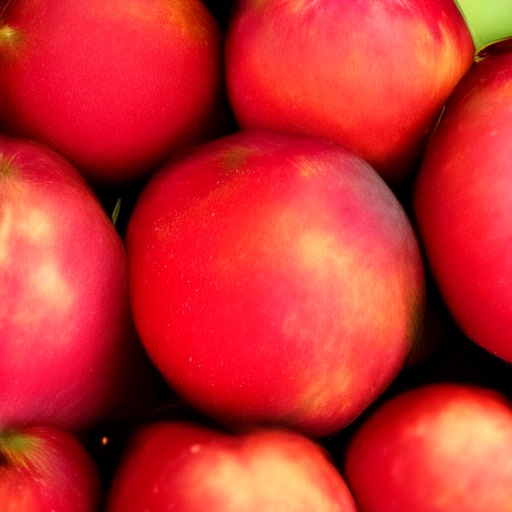}
    \imageWithNote{0.19\textwidth}{\scriptsize \raggedleft \texttt{145M}}{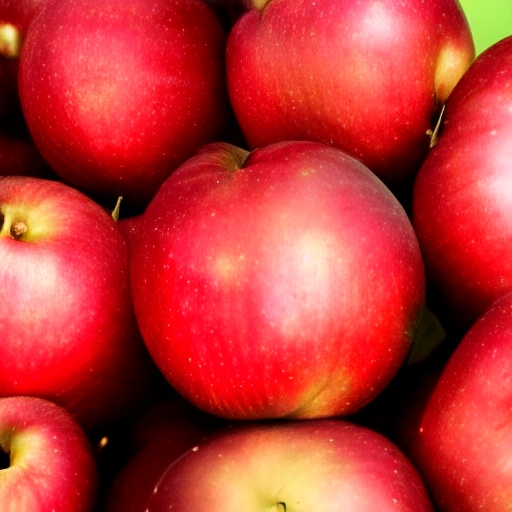}
    \imageWithNote{0.19\textwidth}{\scriptsize \raggedleft \texttt{223M}}{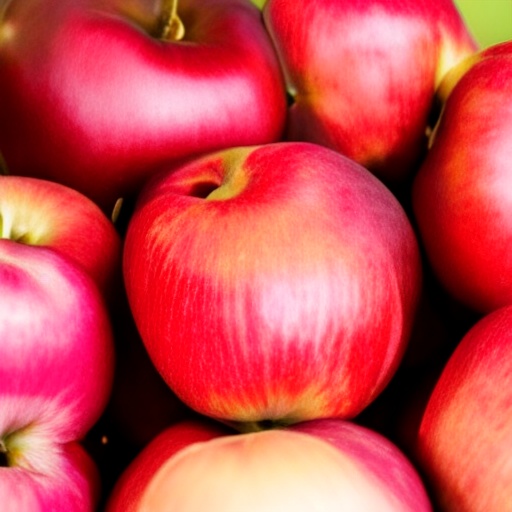}
    \imageWithNote{0.19\textwidth}{\scriptsize \raggedleft \texttt{318M}}{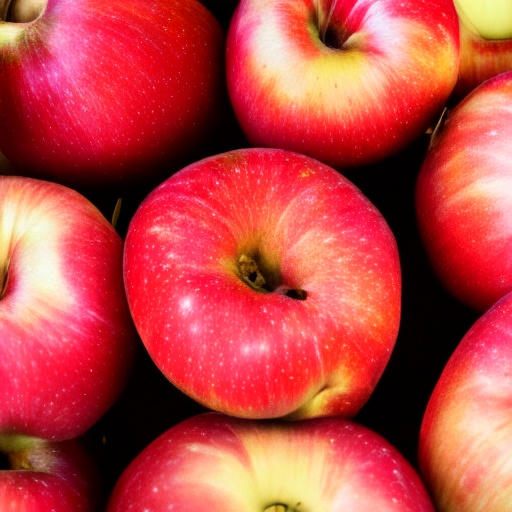}
    \imageWithNote{0.19\textwidth}{\scriptsize \raggedleft \texttt{430M}}{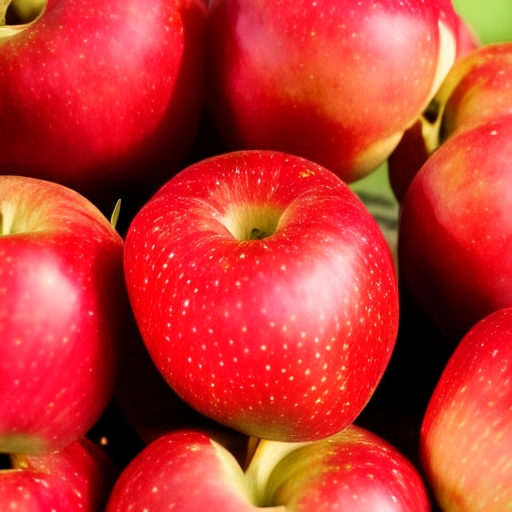}
    \imageWithNote{0.19\textwidth}{\scriptsize \raggedleft \texttt{558M}}{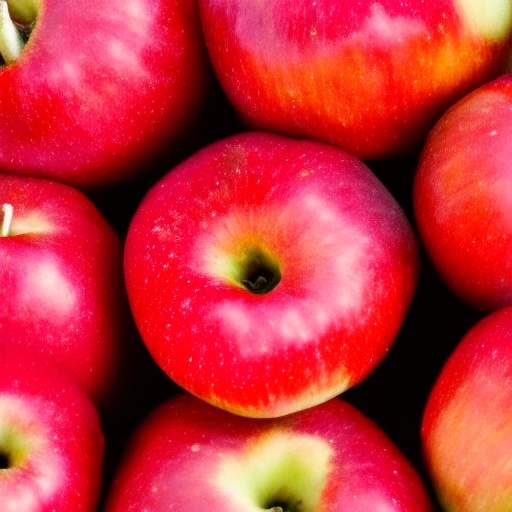}
    \imageWithNote{0.19\textwidth}{\scriptsize \raggedleft \texttt{704M}}{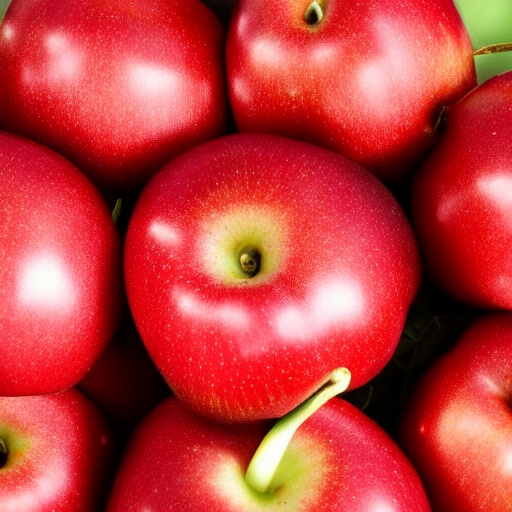}
    \imageWithNote{0.19\textwidth}{\scriptsize \raggedleft \texttt{866M}}{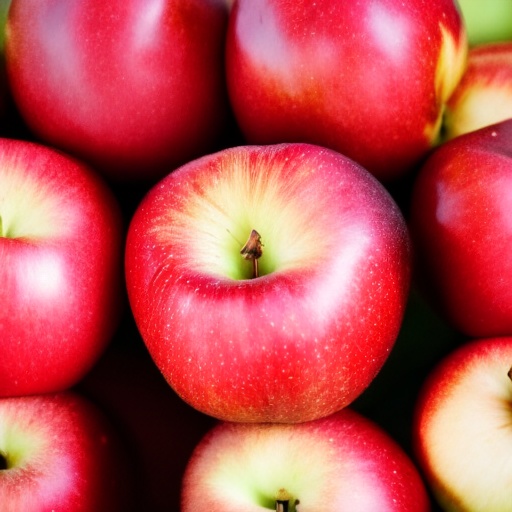}
    \imageWithNote{0.19\textwidth}{\scriptsize \raggedleft \texttt{2B}}{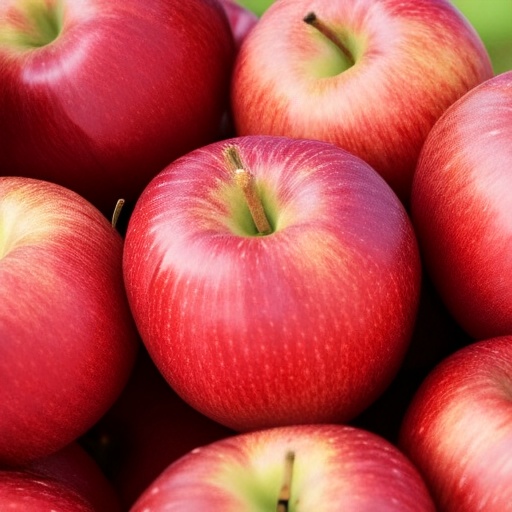}
    \imageWithNote{0.19\textwidth}{\scriptsize \raggedleft \texttt{5B}}{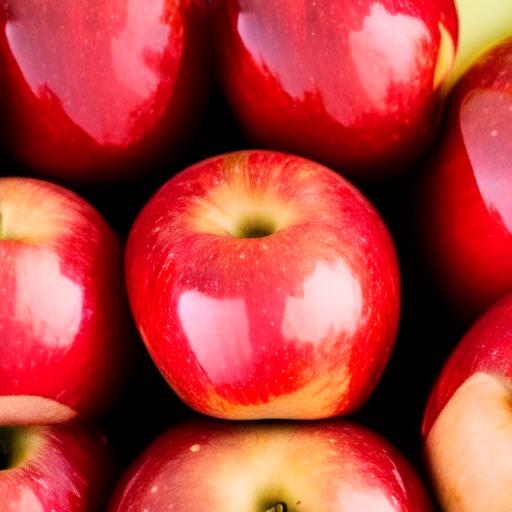}
    \caption{Prompt: \emph{``ten red apples.''}}
    \end{subfigure}
    \caption{Text-to-image results from our scaled LDMs (\texttt{83M} - \texttt{5B}), highlighting the improvement in visual quality with increased model size.}
    \label{fig:suppvisualcompare3}
    \vspace{-2\baselineskip}
\end{figure}

In order to provide detailed visual comparisons for Fig.~\ref{fig:t2i_results} in the main manuscript, Fig.~\ref{fig:suppvisualcompare1}, Fig.~\ref{fig:suppvisualcompare2}, and Fig.~\ref{fig:suppvisualcompare3} show the generated results with the same prompt and the same sampling parameters (\ie, 50-step DDIM sampling and 7.5 CFG rate).

\newpage
\section{Scaling downstream performance}
\label{supp:downstream}
\begin{figure}[!htbp]
    \centering
    \def\xwidth{0.12\linewidth}
    \setlength{\tabcolsep}{1pt}
    \begin{tabular}[t]{c c c c c c}
   &
    \includegraphics[width=\xwidth]{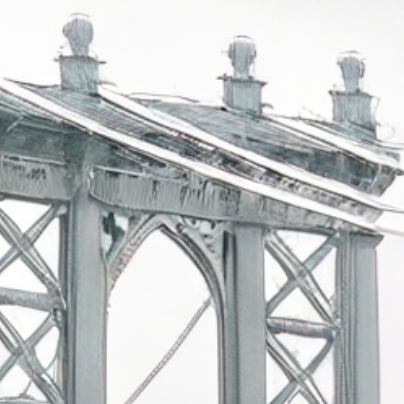} &
    \includegraphics[width=\xwidth]{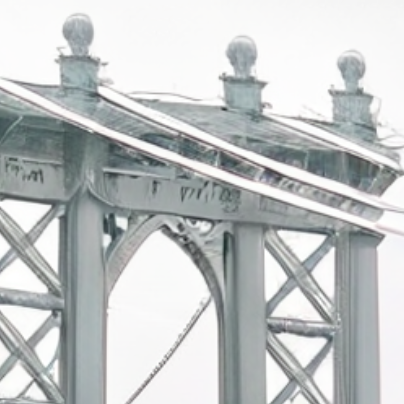} &
    \includegraphics[width=\xwidth]{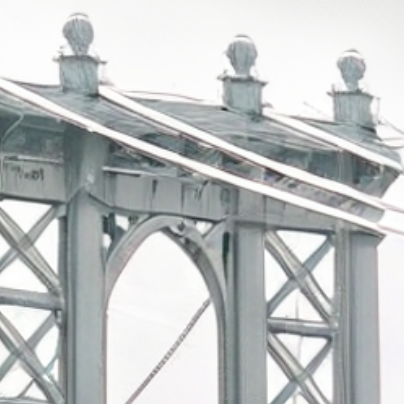} &
    \includegraphics[width=\xwidth]{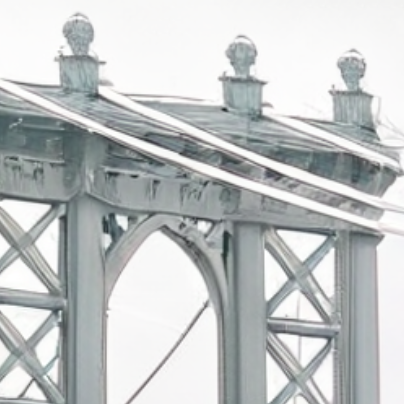} &
    \includegraphics[width=\xwidth]{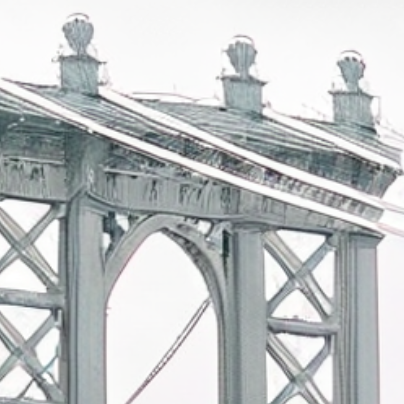}
    \\
    & \small 83M & \small 145M & \small 223M & \small 318M & \small 430M
    \\
    \multirow[t]{3}{*}{
    \includegraphics[width=0.265\linewidth, height=0.27\linewidth]{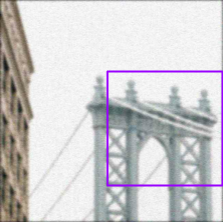}
    } &
    \includegraphics[width=\xwidth]{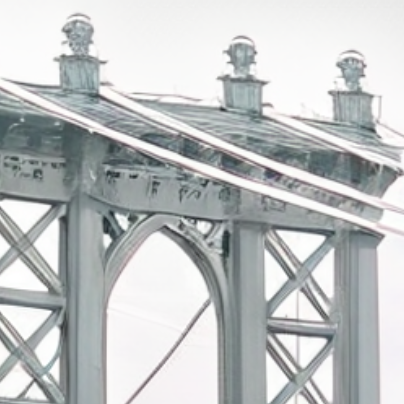} &
    \includegraphics[width=\xwidth]{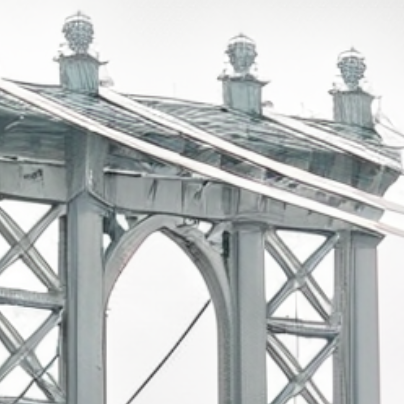} &
    \includegraphics[width=\xwidth]{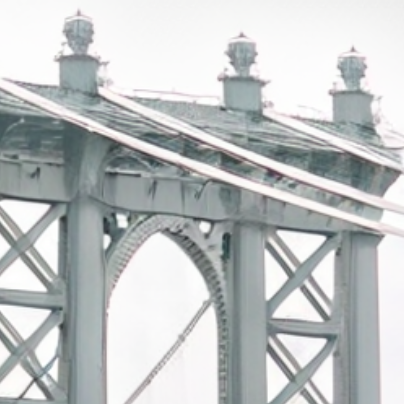} &
    \includegraphics[width=\xwidth]{figures/sr/scaling_sr_c512_3.png} &
    \includegraphics[width=\xwidth]{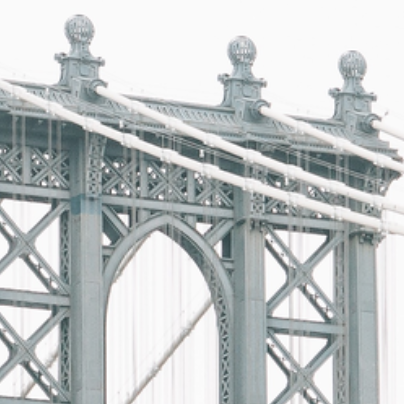}
    \\
    \small LR & \small 558M & \small 704M & \small 866M & \small 2B & \small HR
    \\
    &
    \includegraphics[width=\xwidth]{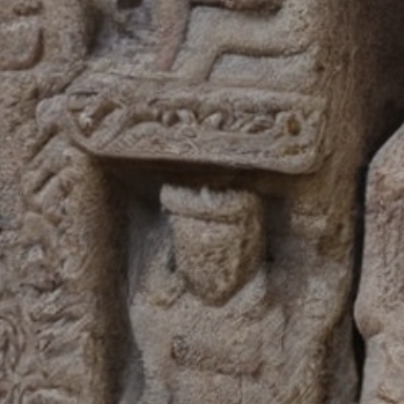} &
    \includegraphics[width=\xwidth]{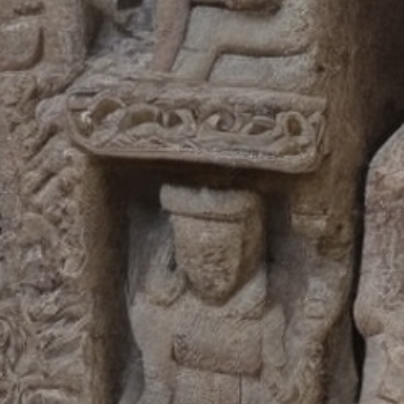} &
    \includegraphics[width=\xwidth]{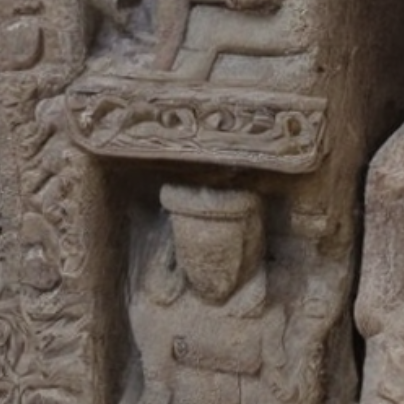} &
    \includegraphics[width=\xwidth]{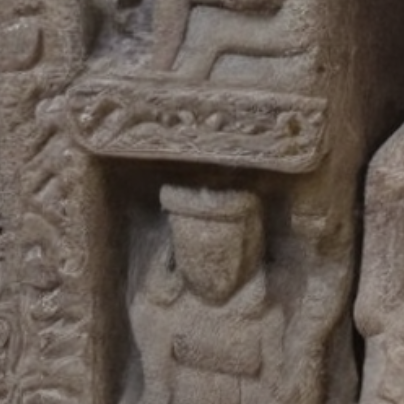} &
    \includegraphics[width=\xwidth]{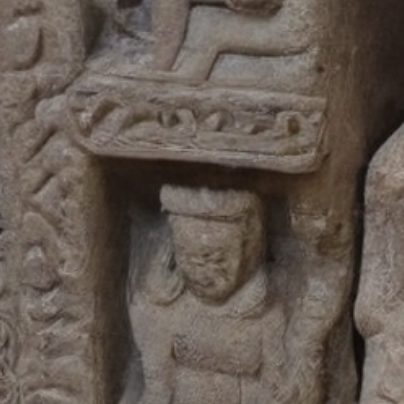}
    \\
    & 83M & 145M & 223M & 318M & 430M
    \\
    \multirow[t]{3}{*}{
    \includegraphics[width=0.265\linewidth, height=0.27\linewidth]{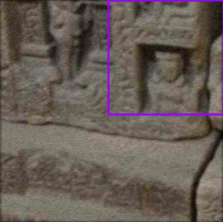}
    } &
    \includegraphics[width=\xwidth]{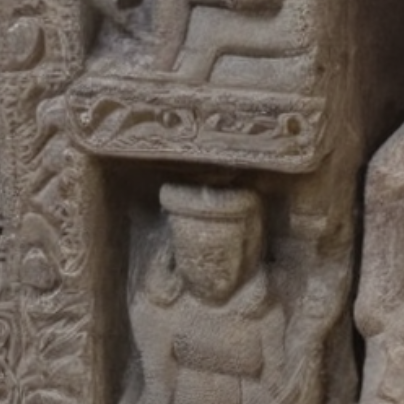} &
    \includegraphics[width=\xwidth]{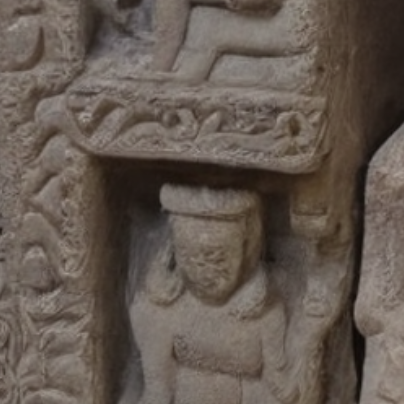} &
    \includegraphics[width=\xwidth]{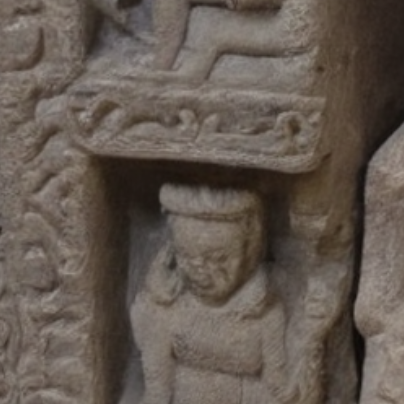} &
    \includegraphics[width=\xwidth]{figures/sr/scaling_sr_c512_4.png} &
    \includegraphics[width=\xwidth]{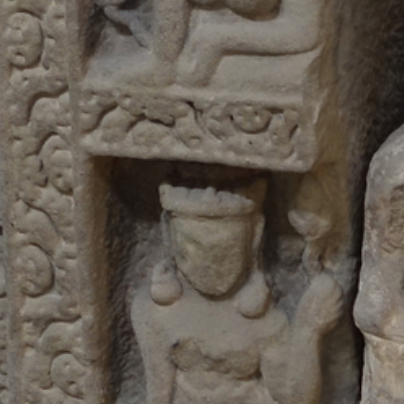}
    \\
    LR &  558M & 704M & 866M & 2B & HR
    \end{tabular}
    \caption{In 4$\times$ super-resolution, visual quality directly improves with increased model size. As these scaled models vary in pretraining performance, the results clearly demonstrate that pretraining boosts super-resolution capabilities.}
    \label{fig:suppsr}
\end{figure}

To provide more metrics for the super-resolution experiments in Fig.~\ref{fig:sr_compute} of the main manuscript, Fig.~\ref{fig:suppsr} shows the generative metric IS for the super-resolution results.
Fig.~\ref{fig:suppsr} shows the visual results of the super-resolution results in order to provide more visual results for the visual comparisons of Fig.~\ref{fig:sr} in the main manuscript.

\begin{figure}[ht]
    \centering
    \def\xwidth{.495\linewidth}
    \includegraphics[width=\xwidth]{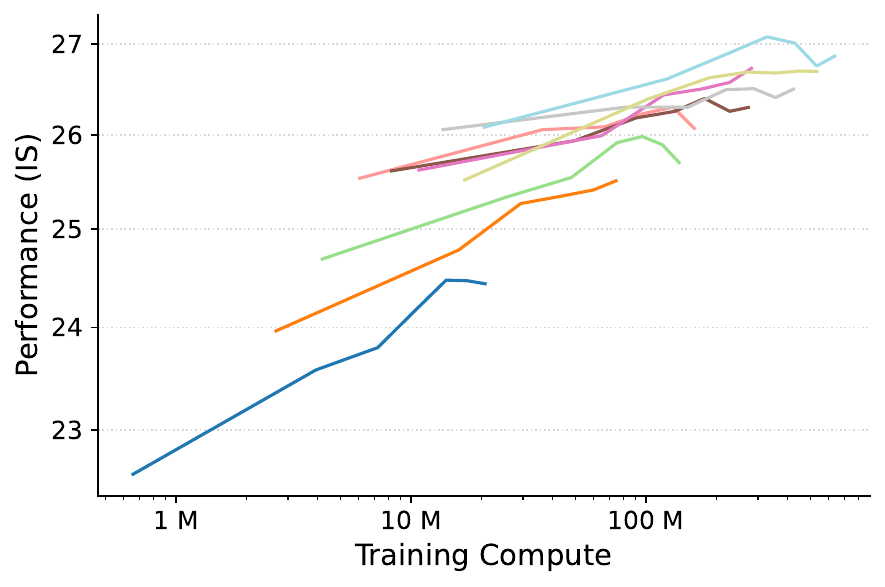}
    \includegraphics[width=\xwidth]{figures/sr_fid_compute.pdf}
    \caption{For super-resolution, we show the trends between the generative metric IS and the training compute still depend on the pretraining, which is similar to the trends between the generative metric FID and the training compute.}
    \label{fig:sr_lpips}
    \vspace{-1\baselineskip}
\end{figure}

\newpage
\section{Scaling sampling-efficiency in distilled LDMs}
\begin{figure}[htbp]
    \centering
    \def\xwidth{.3\linewidth}
    \includegraphics[height=\xwidth]{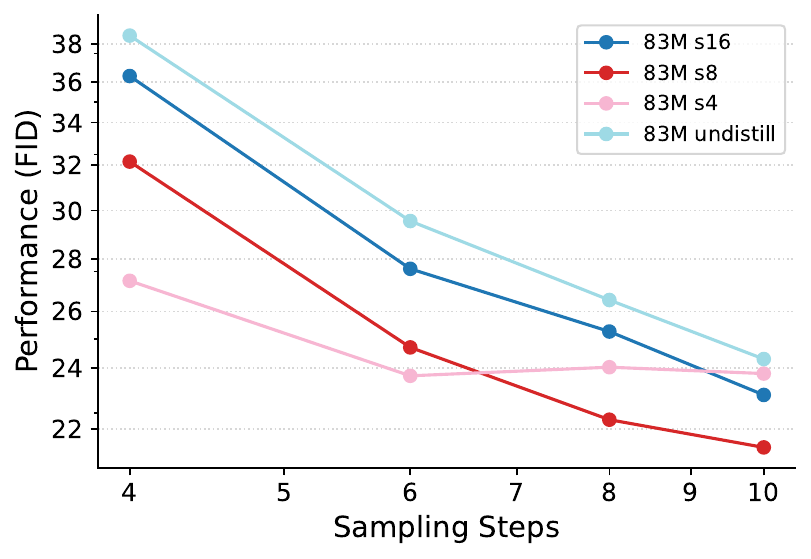}
    \includegraphics[height=\xwidth]{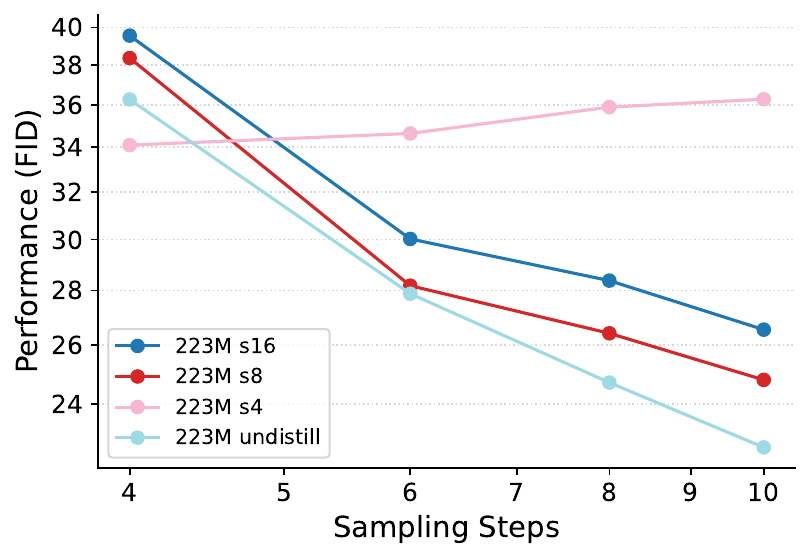}
    \caption{\emph{Left:} Guided distillation on the 83M model for text-to-image generation. \emph{Right:} Guided distillation on the 224M model for text-to-image generation.}
    \vspace{-2\baselineskip}
    \label{fig:suppdistillcost}
\end{figure}

Diffusion distillation methods for accelerating sampling are generally derived from Progressive Distillation (PD)~\cite{salimans2022progressive} and Consistency Models (CM)~\cite{song2023consistency}.
In the main paper, we have shown that CoDi~\cite{mei2023conditional} based on CM is scalable to different model sizes.
Here we show other investigated methods, \ie, guided distillation~\cite{meng2023distillation}, has inconsistent acceleration effects across different model sizes.
Fig.~\ref{fig:suppdistillcost} shows  guided distillation results for the 83M and 223M models respectively, where \texttt{s16} and \texttt{s8} denote different distillation stages.
It is easy to see that the performance improvement of these two models is inconsistent.

\begin{figure}[ht]
    \centering
    \def\xwidth{.25\linewidth}
    \begin{subfigure}[b]{\textwidth}
    \centering
    \imageWithMoreNote{\xwidth}{\scriptsize \raggedleft \texttt{866M 4-step distilled}}{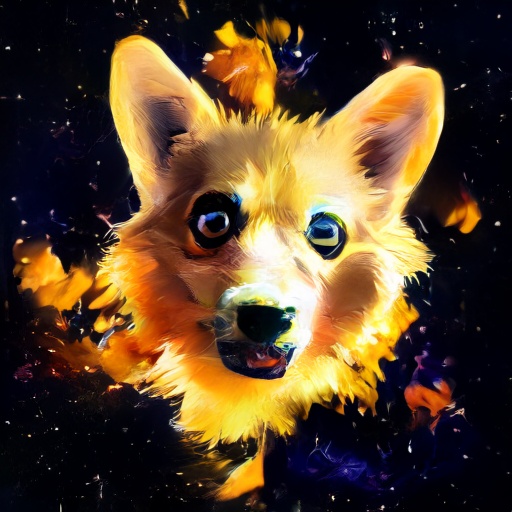}
    \imageWithMoreNote{\xwidth}{\scriptsize \raggedleft \texttt{866M 4-step undistilled}}{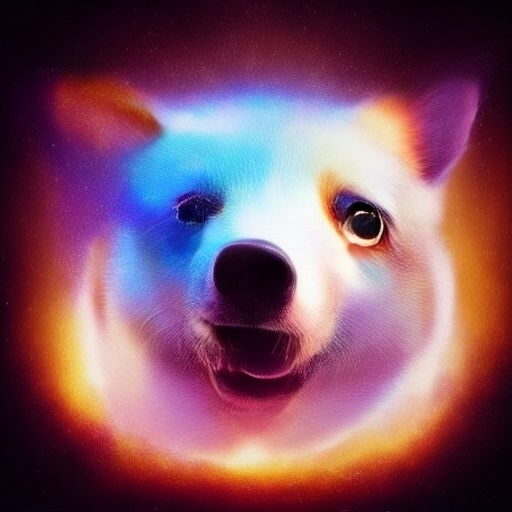}
    \imageWithMoreNote{\xwidth}{\scriptsize \raggedleft \texttt{83M 50-step undistilled}}{figures/t2i-efficiency/0_step_50_6.0.jpg}
    \vspace{.5em}
    \caption{Prompt: \emph{``A corgi's head depicted as a nebula.''}. Sampling Cost $\approx$ 7.}
    \end{subfigure}
    \begin{subfigure}[b]{\textwidth}
    \centering
    \imageWithMoreNote{\xwidth}{\scriptsize \raggedleft \texttt{866M 4-step distilled}}{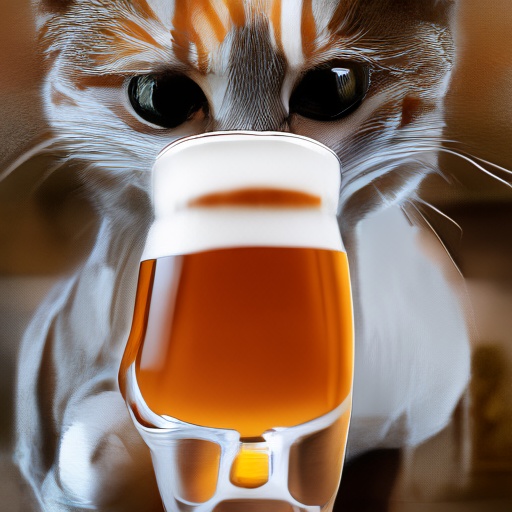}
    \imageWithMoreNote{\xwidth}{\scriptsize \raggedleft \texttt{866M 4-step undistilled}}{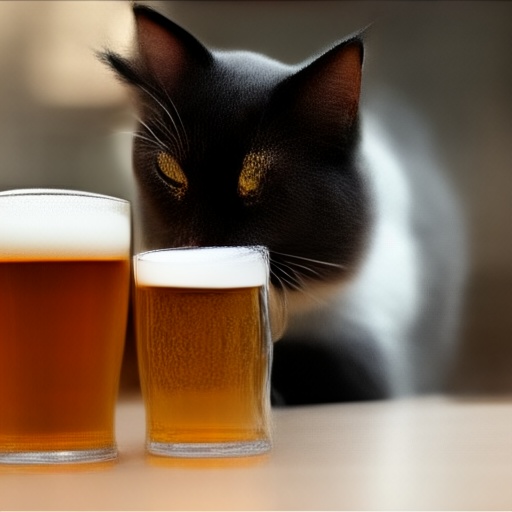}
    \imageWithMoreNote{\xwidth}{\scriptsize \raggedleft \texttt{83M 50-step undistilled}}{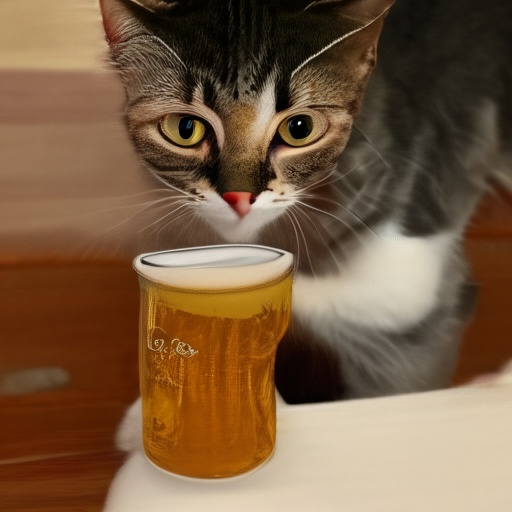}
    \vspace{.5em}
    \caption{Prompt: \emph{``a cat drinking a pint of beer.''}. Sampling Cost $\approx$ 7.}
    \end{subfigure}
    \vspace{-2em}
    \caption{We visualize text-to-image generation results of the tested LDMs under approximately the same inference cost.}
    \label{supp:distillvisual}
    \vspace{-1\baselineskip}
\end{figure}

Fig.~\ref{supp:distillvisual} shows the visual results of the CoDi distilled models and the undistilled models under the same sampling cost to demonstrate the sampling-efficiency.

\section{Scaling the sampling-efficiency}
\begin{figure}[ht]
    \centering
    \def\xwidth{.24\linewidth}
    \begin{subfigure}[b]{\linewidth}
    \imageWithNote{\xwidth}{\scriptsize \raggedleft \texttt{83M\,}}{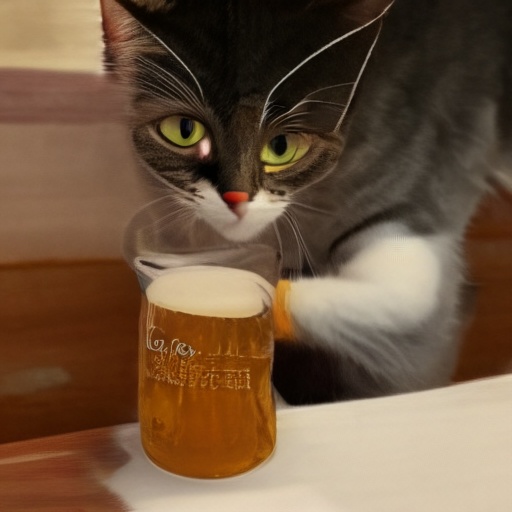}
    \imageWithNote{\xwidth}{\scriptsize \raggedleft \texttt{145M}}{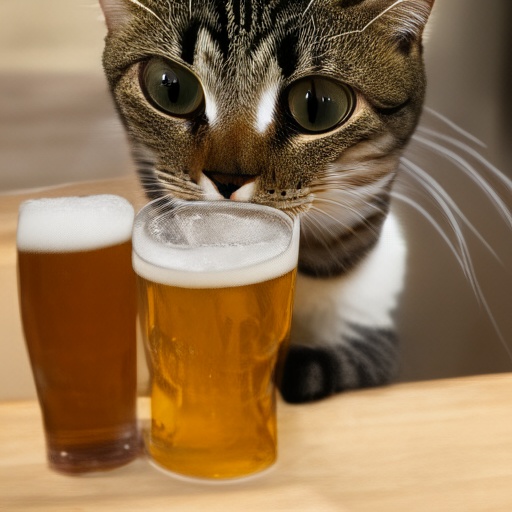}
    \imageWithNote{\xwidth}{\scriptsize \raggedleft \texttt{223M}}{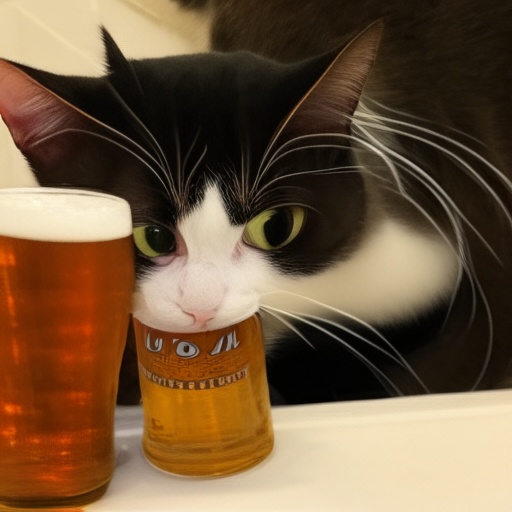}
    \imageWithNote{\xwidth}{\scriptsize \raggedleft \texttt{318M}}{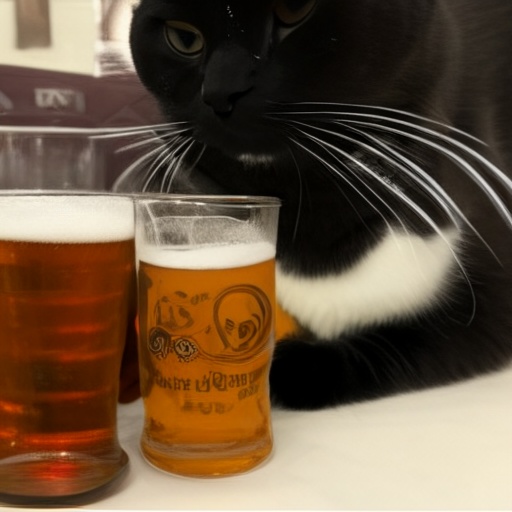}
    \hfill
    
    \imageWithNote{\xwidth}{\scriptsize \raggedleft \texttt{430M}}{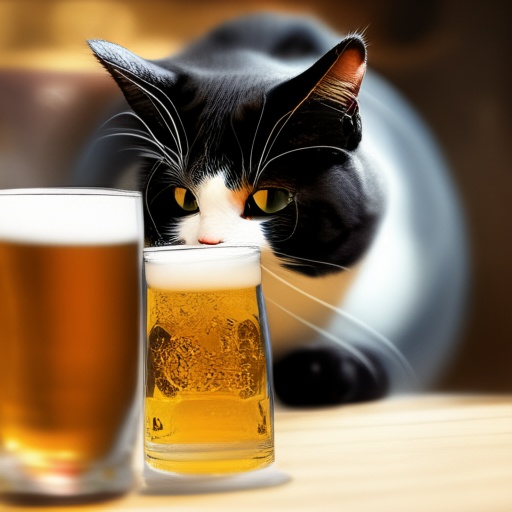}
    \imageWithNote{\xwidth}{\scriptsize \raggedleft \texttt{558M}}{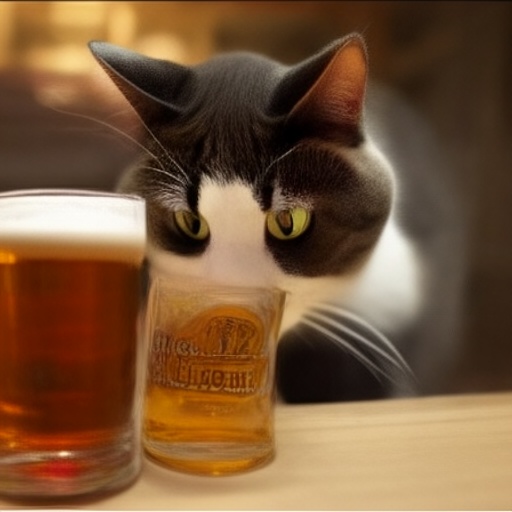}
    \imageWithNote{\xwidth}{\scriptsize \raggedleft \texttt{704M}}{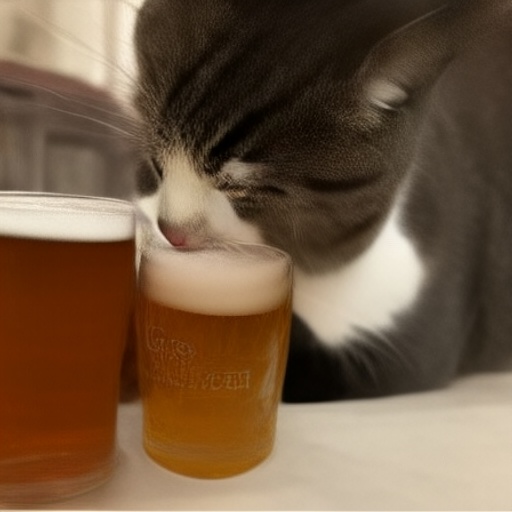}
    \imageWithNote{\xwidth}{\scriptsize \raggedleft \texttt{866M}}{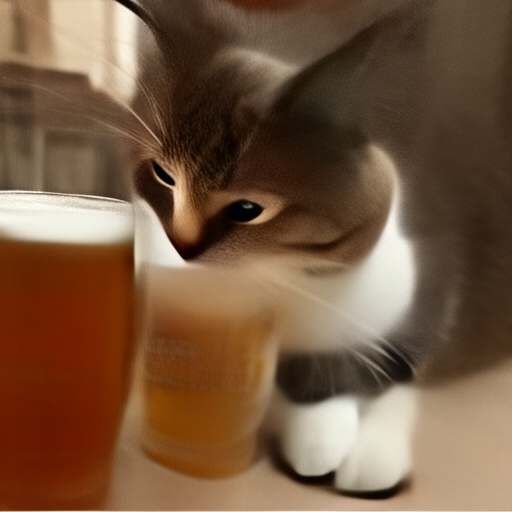}
    \vspace{.5em}
    \caption{Prompt: \emph{``a cat drinking a pint of beer.''}. Sampling Cost $\approx$ 10.}
    \end{subfigure}
    
    \begin{subfigure}[b]{\linewidth}
    
    \imageWithNote{\xwidth}{\scriptsize \raggedleft \texttt{83M\,}}{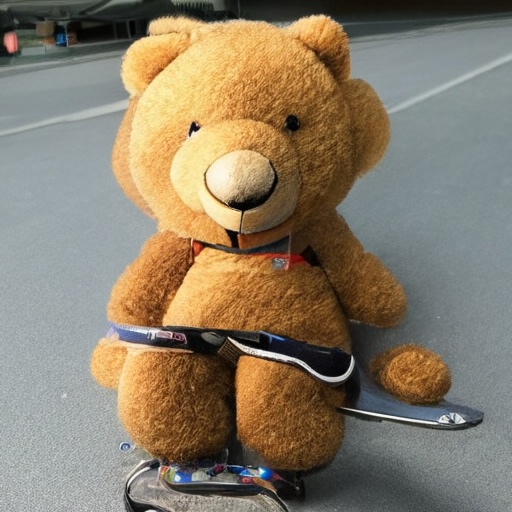}
    \imageWithNote{\xwidth}{\scriptsize \raggedleft \texttt{145M}}{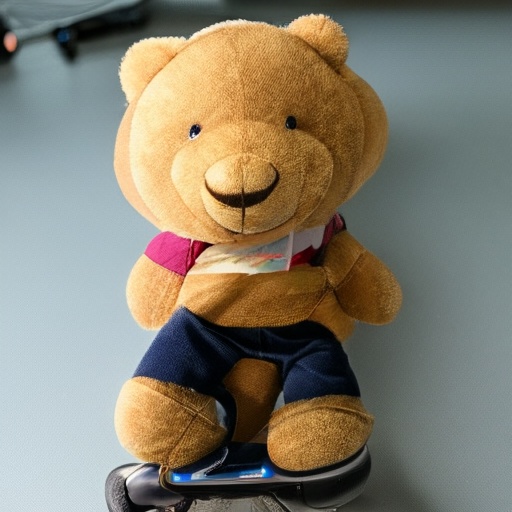}
    \imageWithNote{\xwidth}{\scriptsize \raggedleft \texttt{223M}}{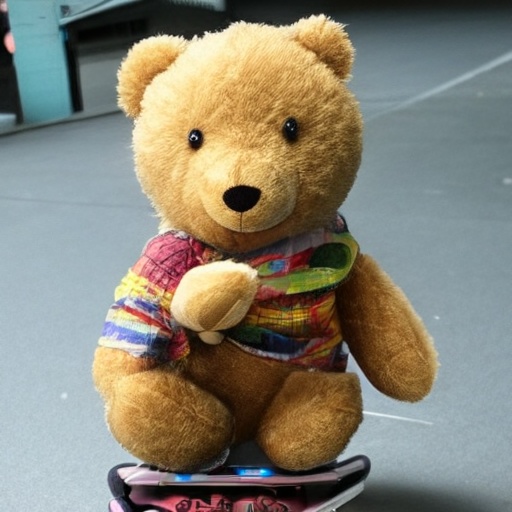}
    \imageWithNote{\xwidth}{\scriptsize \raggedleft \texttt{318M}}{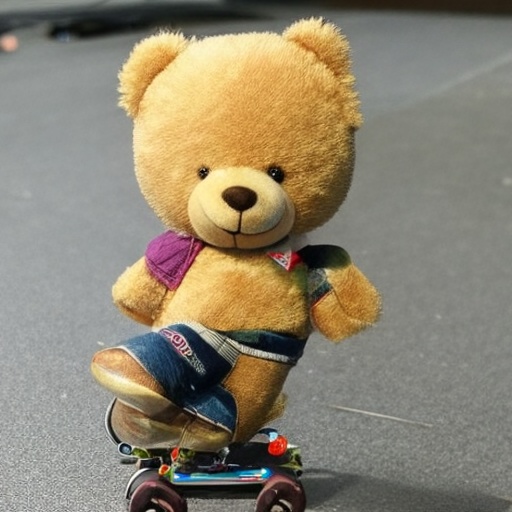}
    \hfill
    
    \imageWithNote{\xwidth}{\scriptsize \raggedleft \texttt{430M}}{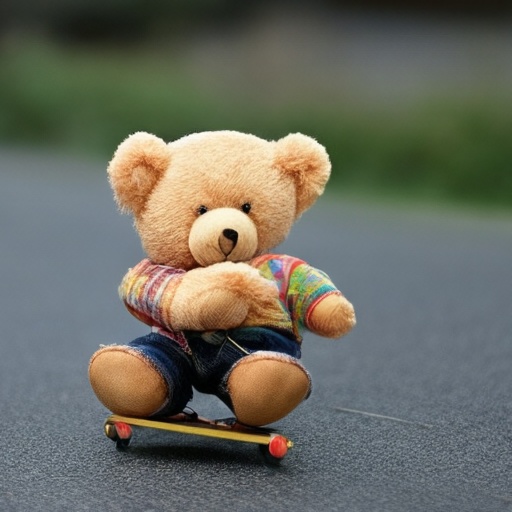}
    \imageWithNote{\xwidth}{\scriptsize \raggedleft \texttt{558M}}{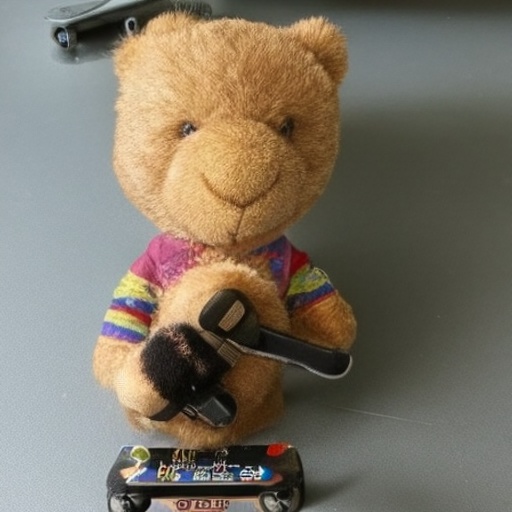}
    \imageWithNote{\xwidth}{\scriptsize \raggedleft \texttt{704M}}{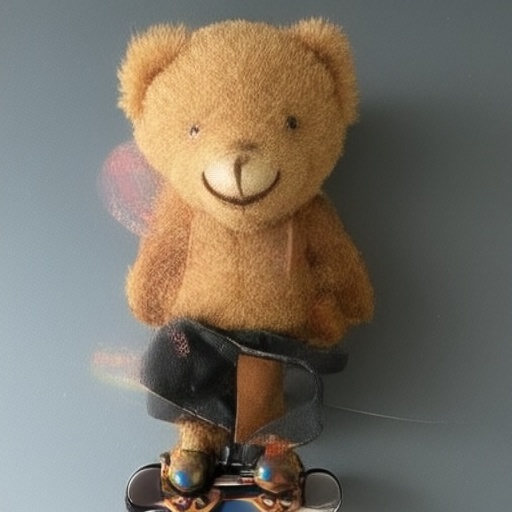}
    \imageWithNote{\xwidth}{\scriptsize \raggedleft \texttt{866M}}{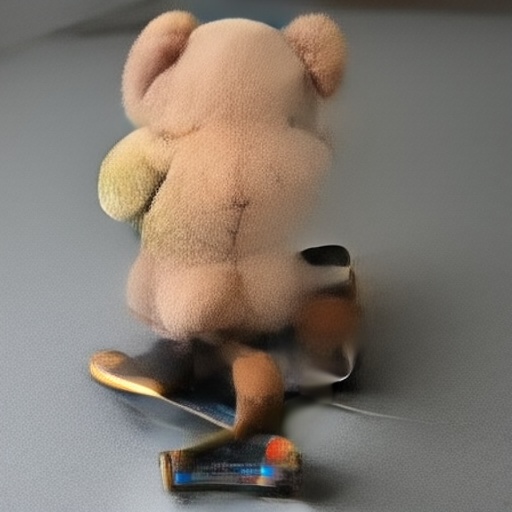}
    \vspace{.5em}
    \caption{Prompt: \emph{``a teddy bear on a skateboard.''}. Sampling Cost $\approx$ 10.}
    \end{subfigure}
    
    \caption{We visualize text-to-image generation results of the tested LDMs under approximately the same inference cost. We observe that smaller models can produce comparable or even better visual results than larger models under similar sampling cost (model GFLOPs $\times$ sampling steps).}
    \label{supp:efficiency1}
\end{figure}

\begin{figure}[ht]
    \centering
    \def\xwidth{.24\linewidth}
    \begin{subfigure}[b]{\linewidth}
    \imageWithNote{\xwidth}{\scriptsize \raggedleft \texttt{83M\,}}{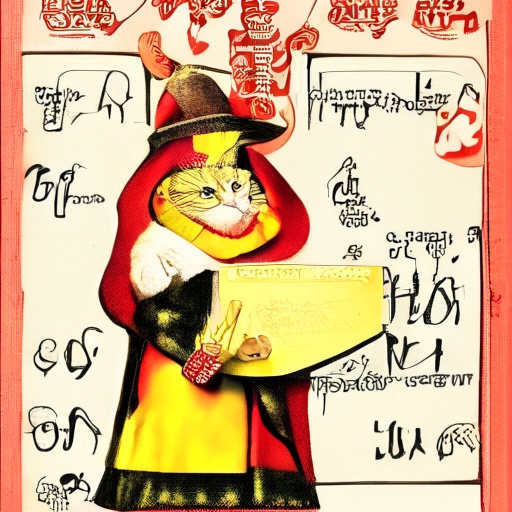}
    \imageWithNote{\xwidth}{\scriptsize \raggedleft \texttt{145M}}{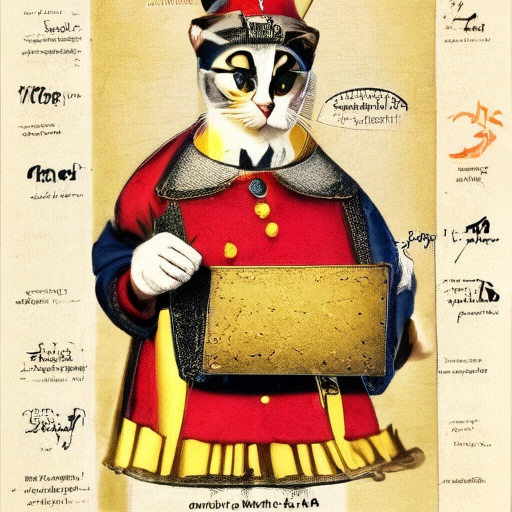}
    \imageWithNote{\xwidth}{\scriptsize \raggedleft \texttt{223M}}{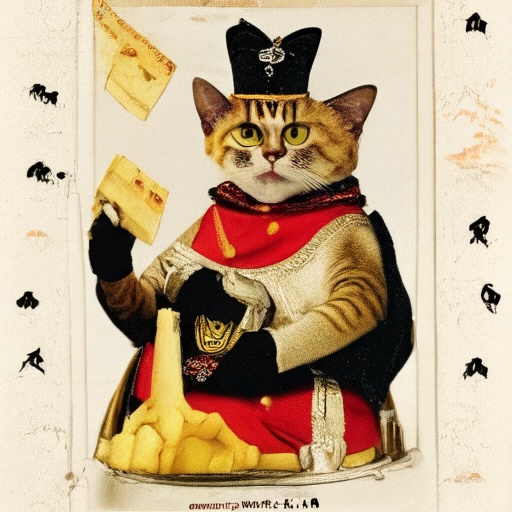}
    \imageWithNote{\xwidth}{\scriptsize \raggedleft \texttt{318M}}{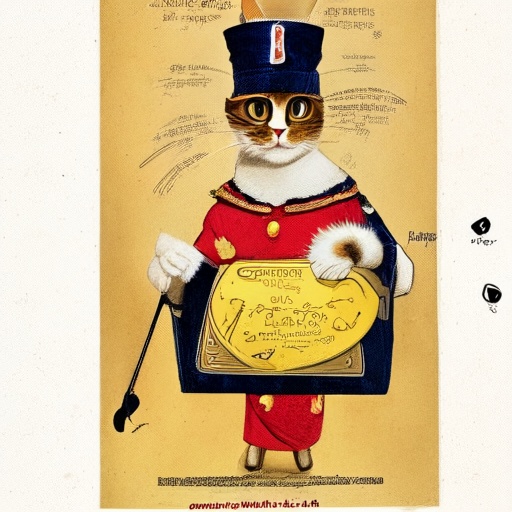}
    \hfill
    
    \imageWithNote{\xwidth}{\scriptsize \raggedleft \texttt{430M}}{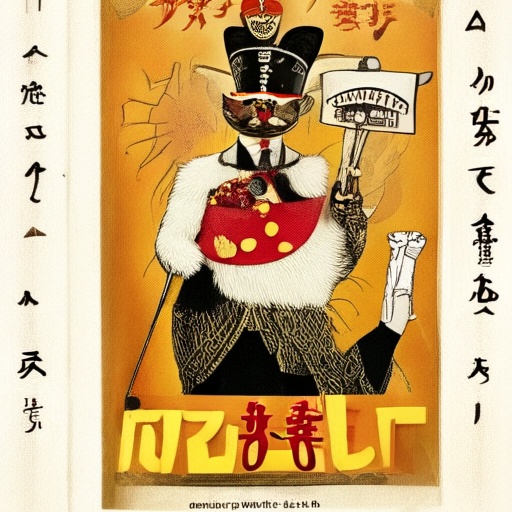}
    \imageWithNote{\xwidth}{\scriptsize \raggedleft \texttt{558M}}{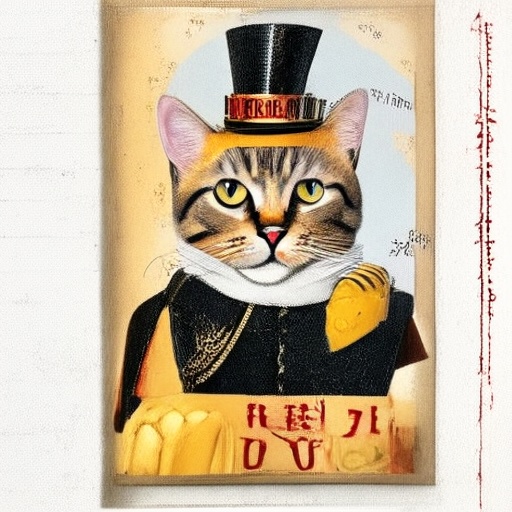}
    \imageWithNote{\xwidth}{\scriptsize \raggedleft \texttt{704M}}{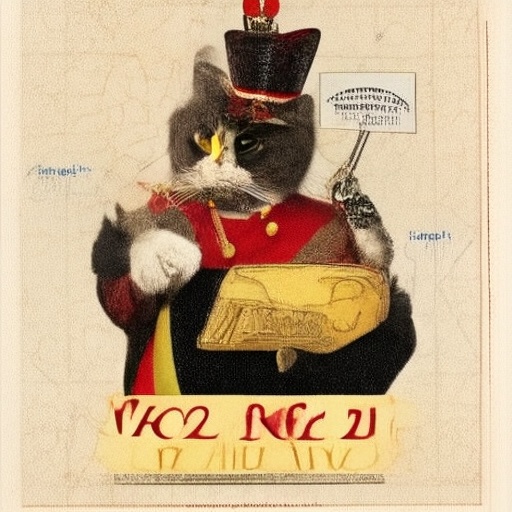}
    \imageWithNote{\xwidth}{\scriptsize \raggedleft \texttt{866M}}{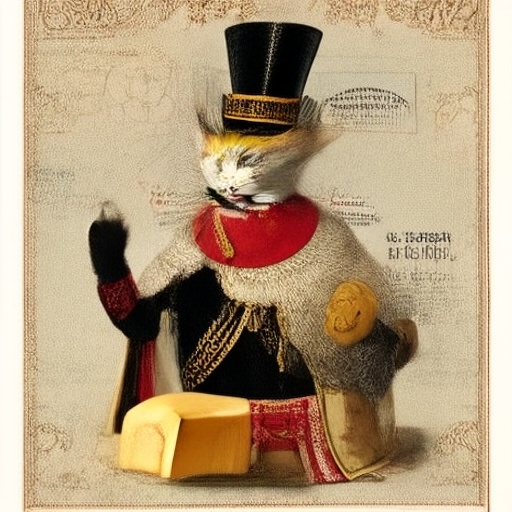}
    \vspace{.5em}
    \caption{Prompt: \emph{``a propaganda poster depicting a cat dressed as french emperor napoleon holding a piece of cheese.''}. Sampling Cost $\approx$ 14.}
    \end{subfigure}

    \vspace{2em}

    \begin{subfigure}[b]{\linewidth}
    
    \imageWithNote{\xwidth}{\scriptsize \raggedleft \texttt{83M\,}}{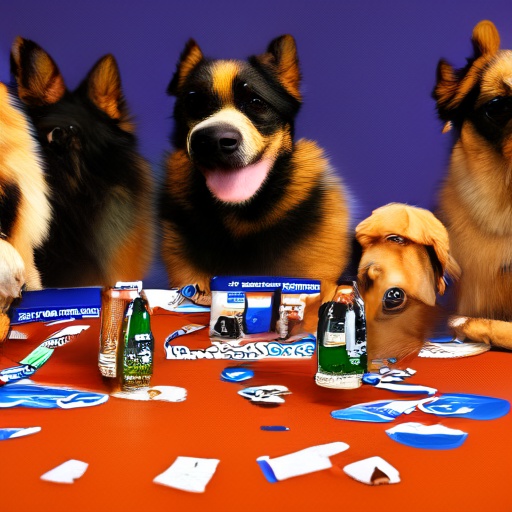}
    \imageWithNote{\xwidth}{\scriptsize \raggedleft \texttt{145M}}{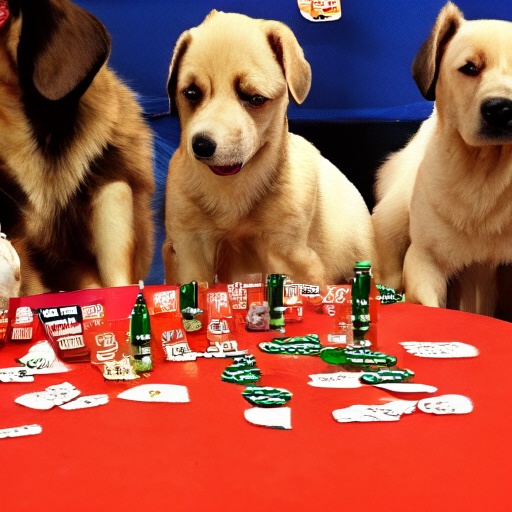}
    \imageWithNote{\xwidth}{\scriptsize \raggedleft \texttt{223M}}{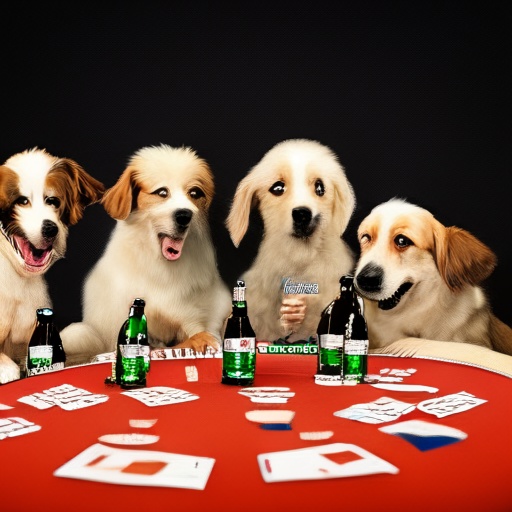}
    \imageWithNote{\xwidth}{\scriptsize \raggedleft \texttt{318M}}{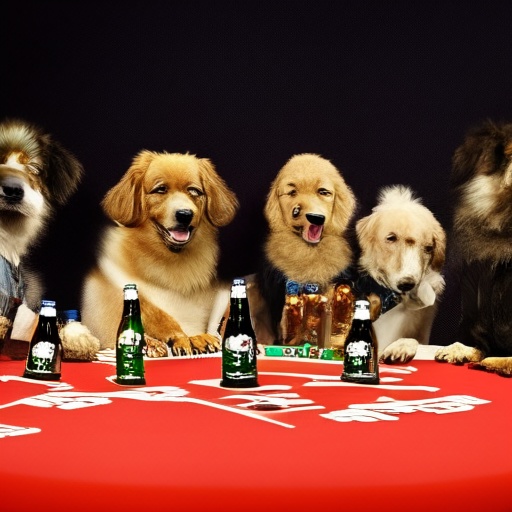}
    \hfill
    
    \imageWithNote{\xwidth}{\scriptsize \raggedleft \texttt{430M}}{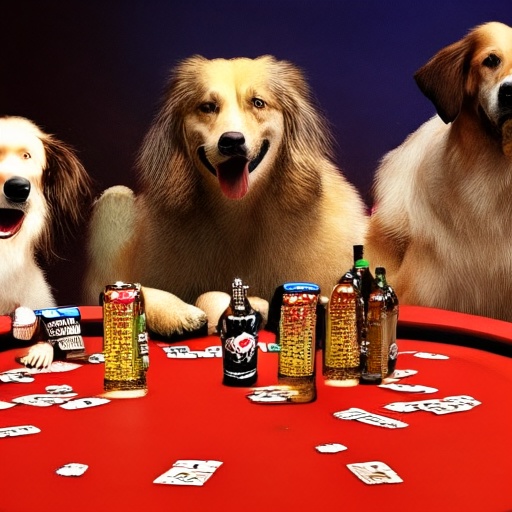}
    \imageWithNote{\xwidth}{\scriptsize \raggedleft \texttt{558M}}{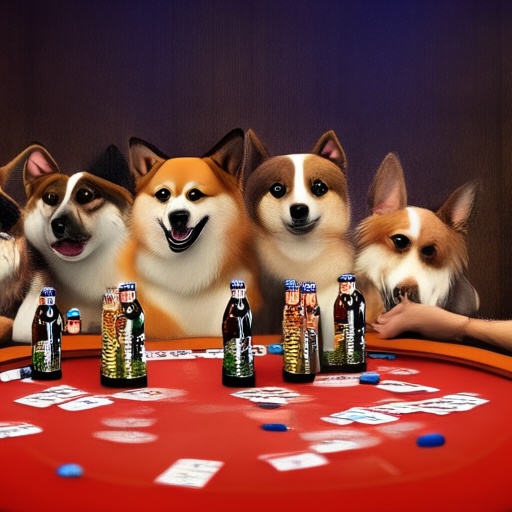}
    \imageWithNote{\xwidth}{\scriptsize \raggedleft \texttt{704M}}{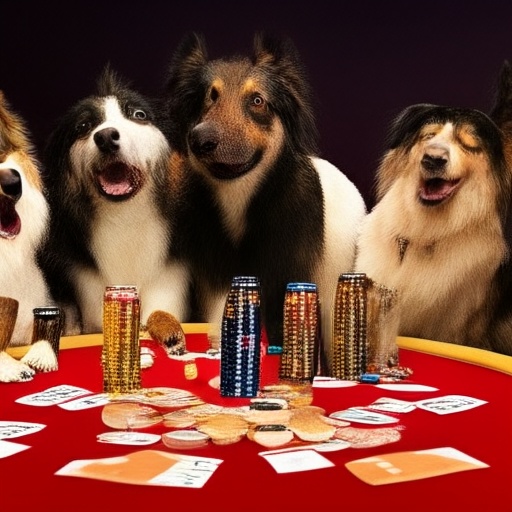}
    \imageWithNote{\xwidth}{\scriptsize \raggedleft \texttt{866M}}{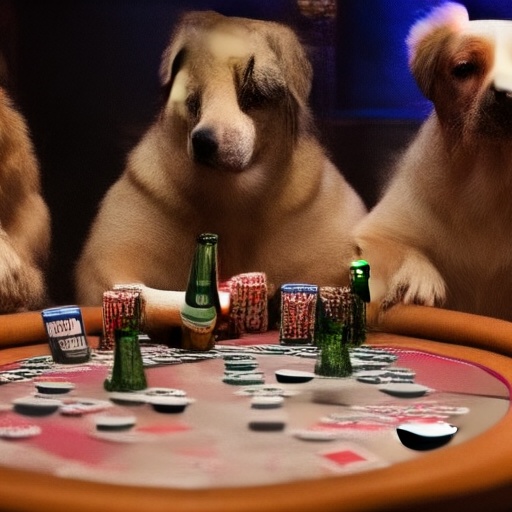}
    \vspace{.5em}
    \caption{Prompt: \emph{``Dogs sitting around a poker table with beer bottles and chips.''}. Sampling Cost $\approx$ 14.}
    \end{subfigure}   

    \caption{We visualize text-to-image generation results of the tested LDMs under approximately the same inference cost. We observe that smaller models can produce comparable or even better visual results than larger models under similar sampling cost (model GFLOPs $\times$ sampling steps).}
    \label{supp:efficiency2}
\end{figure}

\begin{figure}[ht]
    \centering
    \def\xwidth{.24\linewidth}
    \begin{subfigure}[b]{\linewidth}
    \imageWithNote{\xwidth}{\scriptsize \raggedleft \texttt{83M\,}}{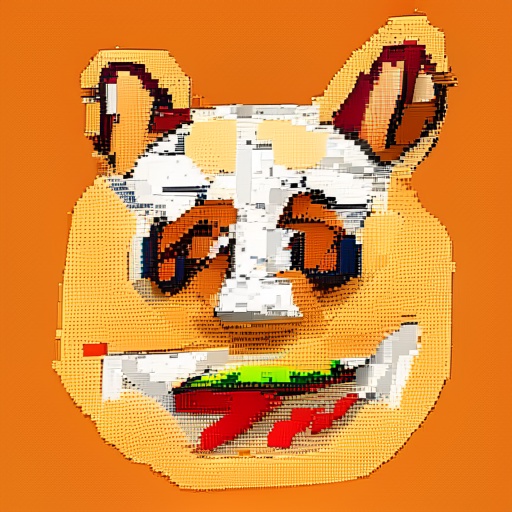}
    \imageWithNote{\xwidth}{\scriptsize \raggedleft \texttt{145M}}{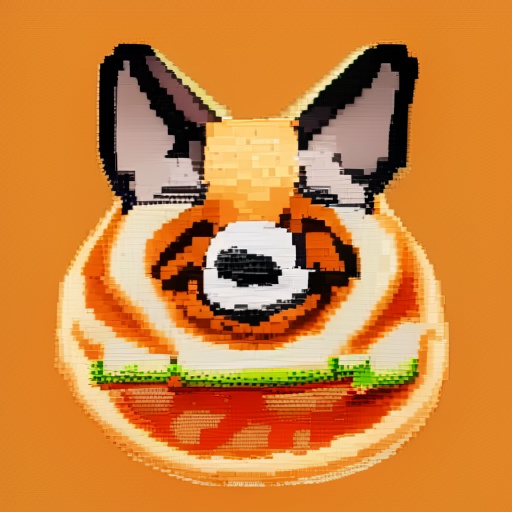}
    \imageWithNote{\xwidth}{\scriptsize \raggedleft \texttt{223M}}{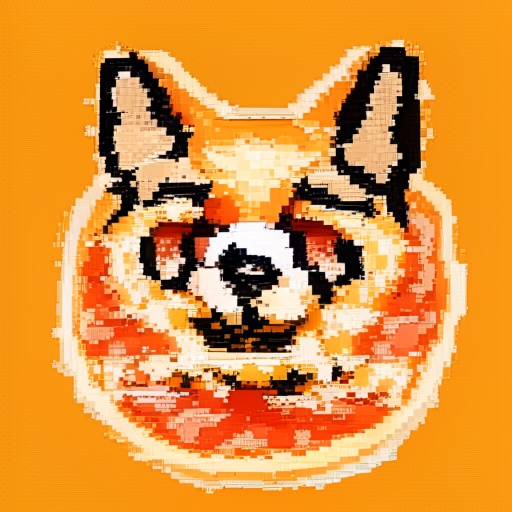}
    \imageWithNote{\xwidth}{\scriptsize \raggedleft \texttt{318M}}{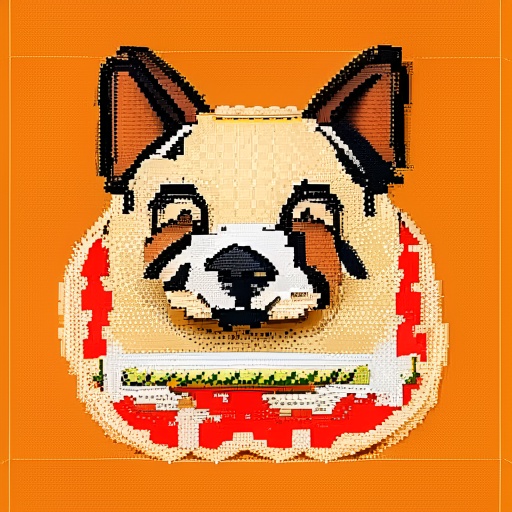}
    \hfill
    
    \imageWithNote{\xwidth}{\scriptsize \raggedleft \texttt{430M}}{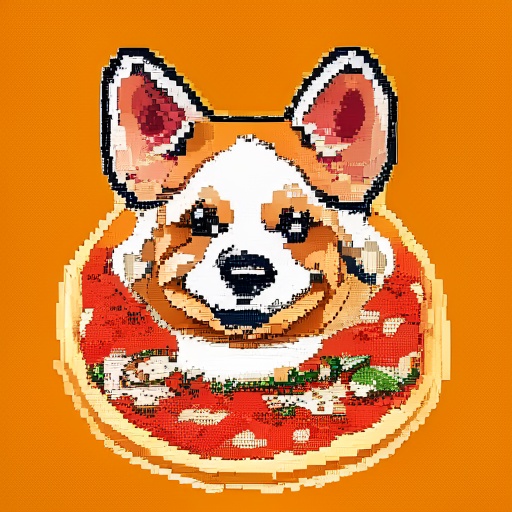}
    \imageWithNote{\xwidth}{\scriptsize \raggedleft \texttt{558M}}{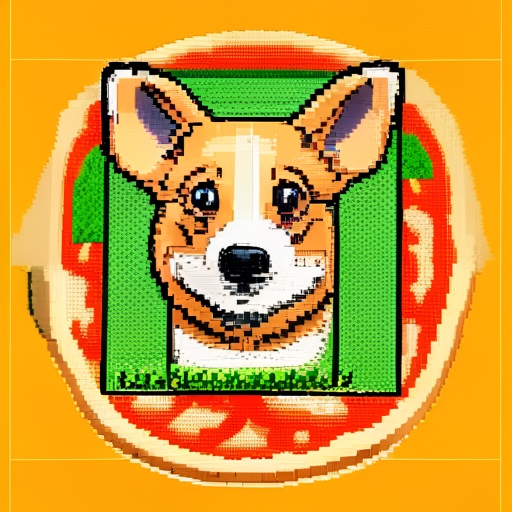}
    \imageWithNote{\xwidth}{\scriptsize \raggedleft \texttt{704M}}{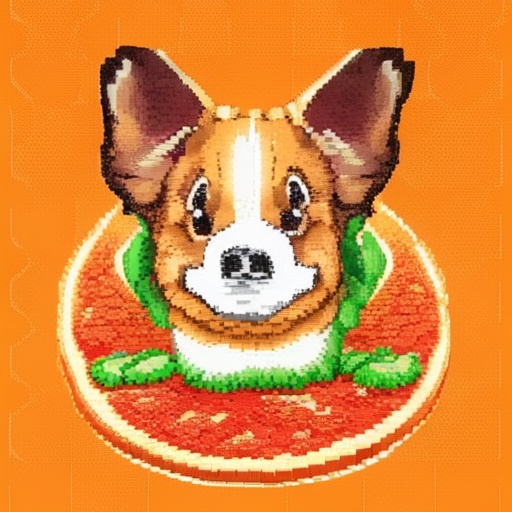}
    \imageWithNote{\xwidth}{\scriptsize \raggedleft \texttt{866M}}{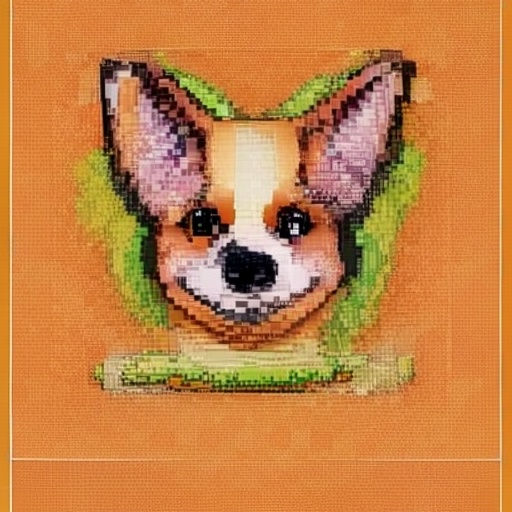}
    \vspace{.5em}
    \caption{Prompt: \emph{``a pixel art corgi pizza.''}. Sampling Cost $\approx$ 18.}
    \end{subfigure}

    \vspace{2em}

    \begin{subfigure}[b]{\linewidth}
    
    \imageWithNote{\xwidth}{\scriptsize \raggedleft \texttt{83M\,}}{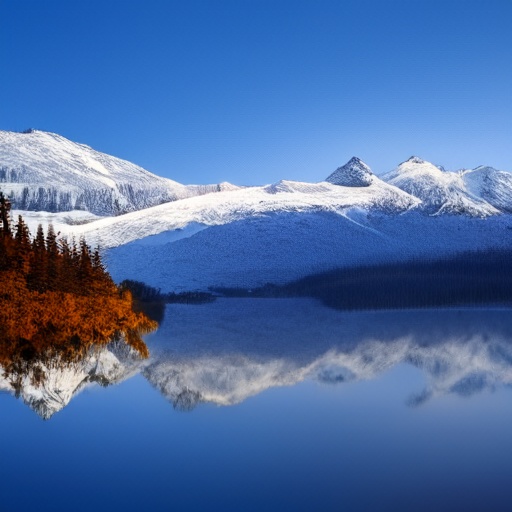}
    \imageWithNote{\xwidth}{\scriptsize \raggedleft \texttt{145M}}{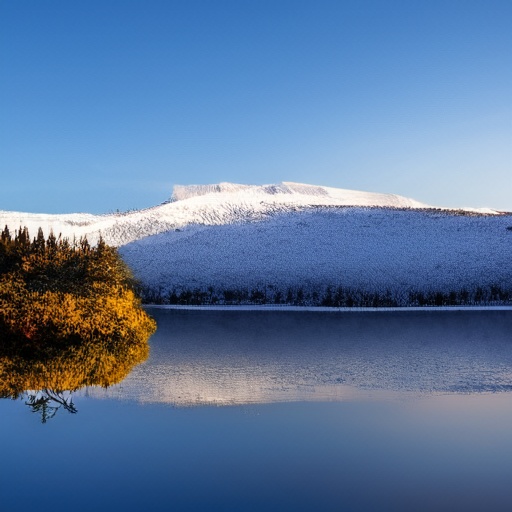}
    \imageWithNote{\xwidth}{\scriptsize \raggedleft \texttt{223M}}{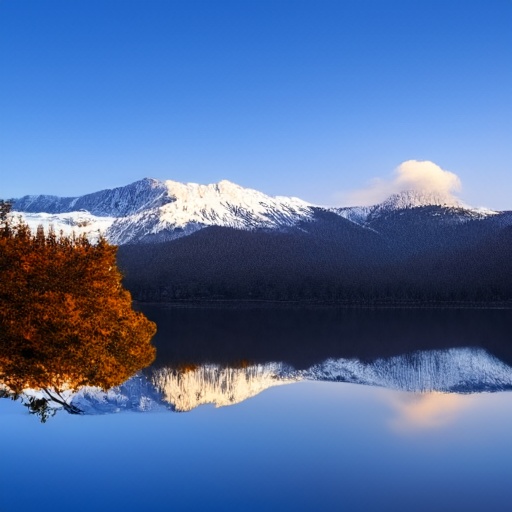}
    \imageWithNote{\xwidth}{\scriptsize \raggedleft \texttt{318M}}{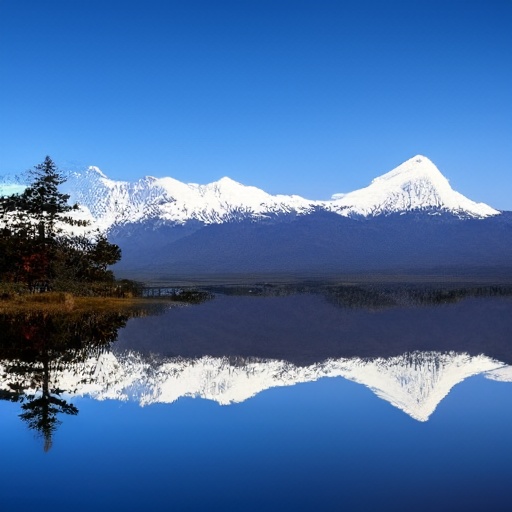}
    \hfill
    
    \imageWithNote{\xwidth}{\scriptsize \raggedleft \texttt{430M}}{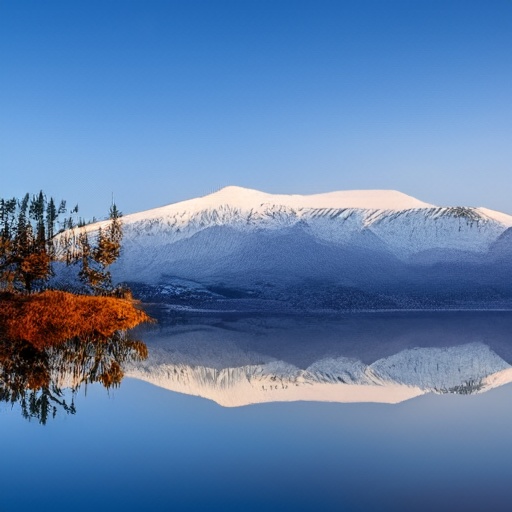}
    \imageWithNote{\xwidth}{\scriptsize \raggedleft \texttt{558M}}{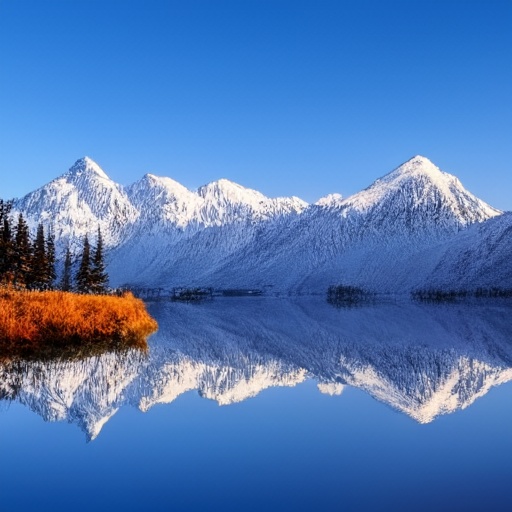}
    \imageWithNote{\xwidth}{\scriptsize \raggedleft \texttt{704M}}{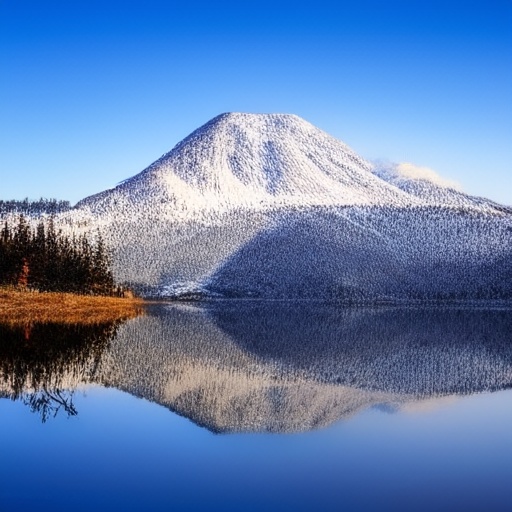}
    \imageWithNote{\xwidth}{\scriptsize \raggedleft \texttt{866M}}{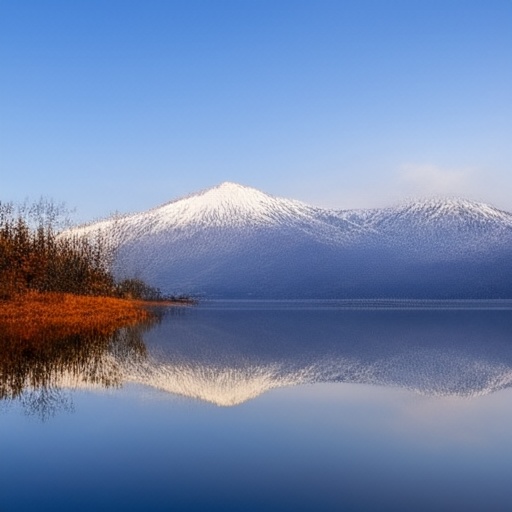}
    \vspace{.5em}
    \caption{Prompt: \emph{``Snow mountain and tree reflection in the lake.''}. Sampling Cost $\approx$ 18.}
    \end{subfigure}   

    \caption{We visualize text-to-image generation results of the tested LDMs under approximately the same inference cost. We observe that smaller models can produce comparable or even better visual results than larger models under similar sampling cost (model GFLOPs $\times$ sampling steps).}
    \label{supp:efficiency3}
\end{figure}

To provide more visual comparisons additional to Fig.~\ref{fig:optimalvisual} in the main paper, Fig.~\ref{supp:efficiency1}, Fig.\ref{supp:efficiency2}, and Fig.~\ref{supp:efficiency3} present visual comparisons between different scaled models under a uniform sampling cost. This highlights that the performance of smaller models can indeed match their larger counterparts under similar sampling cost.

\newpage

\newpage

\newpage
\newpage
\section{Scaling interpretability of text prompt interpolatation}

\begin{figure}[hb]
    \centering
    \def\xwidth{.19\linewidth}
    \begin{subfigure}[b]{\linewidth}
    \imageWithMoreNote{\xwidth}{\scriptsize \raggedleft \texttt{145M $\alpha=1.0$}}{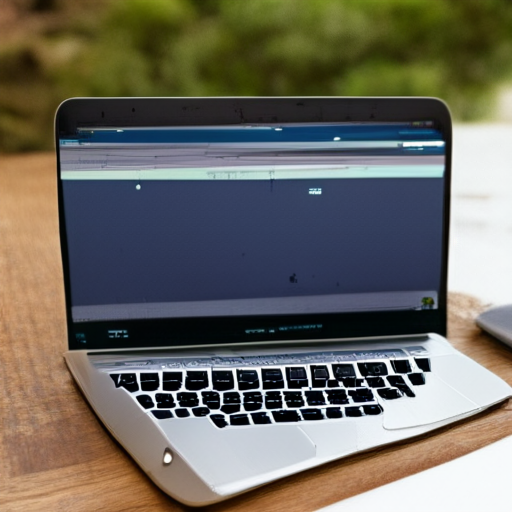}
    \imageWithMoreNote{\xwidth}{\scriptsize \raggedleft \texttt{145M $\alpha=0.8$}}{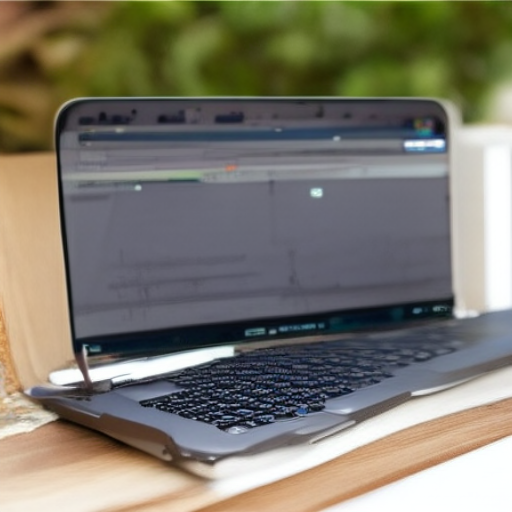}
    \imageWithMoreNote{\xwidth}{\scriptsize \raggedleft \texttt{145M $\alpha=0.5$}}{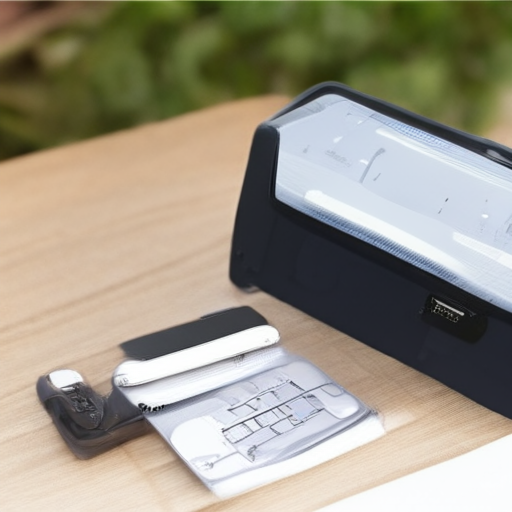}
    \imageWithMoreNote{\xwidth}{\scriptsize \raggedleft \texttt{145M $\alpha=0.2$}}{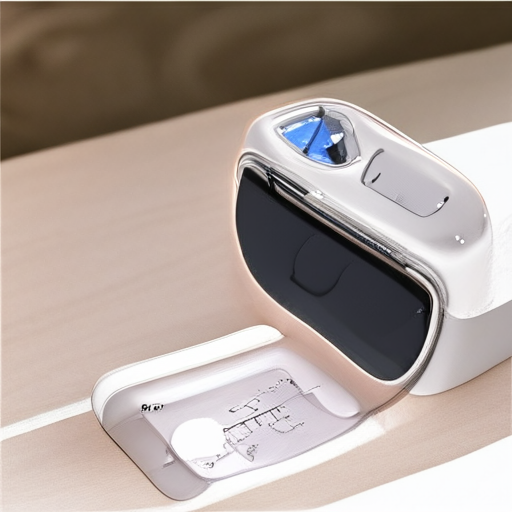}
    \imageWithMoreNote{\xwidth}{\scriptsize \raggedleft \texttt{145M $\alpha=0.0$}}{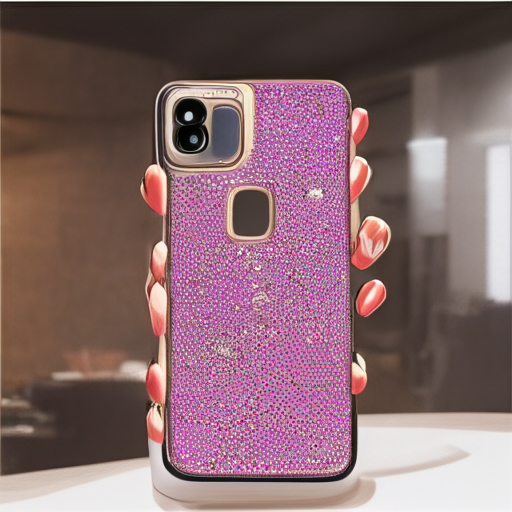}
    \hfill
    
    \imageWithMoreNote{\xwidth}{\scriptsize \raggedleft \texttt{318M $\alpha=1.0$}}{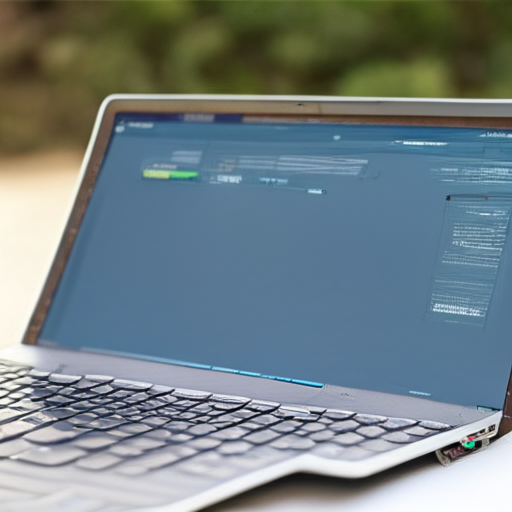}
    \imageWithMoreNote{\xwidth}{\scriptsize \raggedleft \texttt{318M $\alpha=0.8$}}{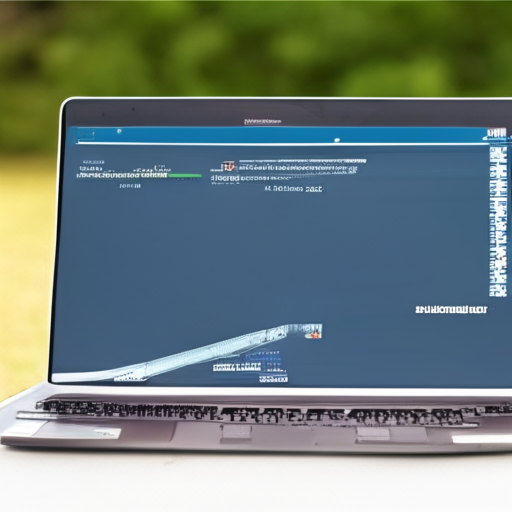}
    \imageWithMoreNote{\xwidth}{\scriptsize \raggedleft \texttt{318M $\alpha=0.5$}}{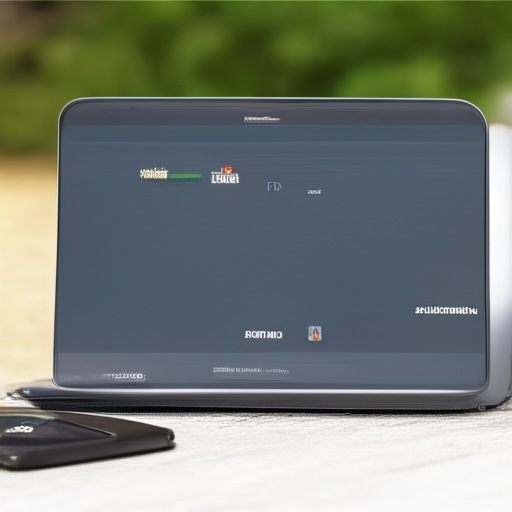}
    \imageWithMoreNote{\xwidth}{\scriptsize \raggedleft \texttt{318M $\alpha=0.2$}}{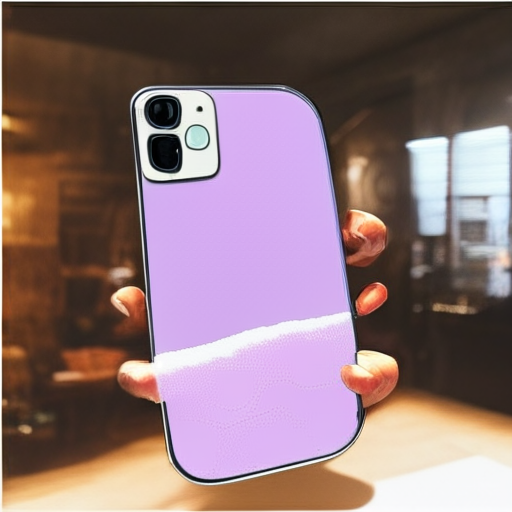}
    \imageWithMoreNote{\xwidth}{\scriptsize \raggedleft \texttt{318M $\alpha=0.0$}}{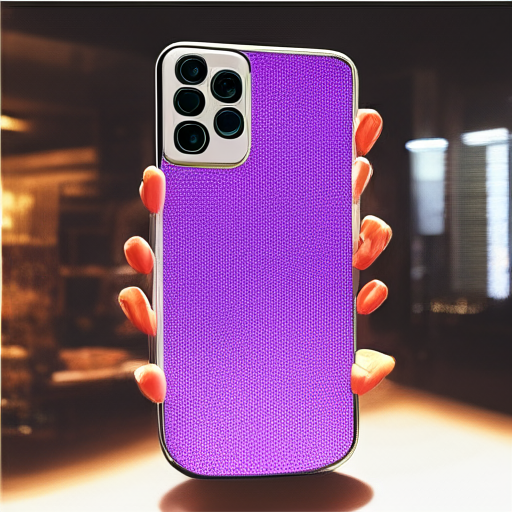}
    \hfill
    
    \imageWithMoreNote{\xwidth}{\scriptsize \raggedleft \texttt{430M $\alpha=1.0$}}{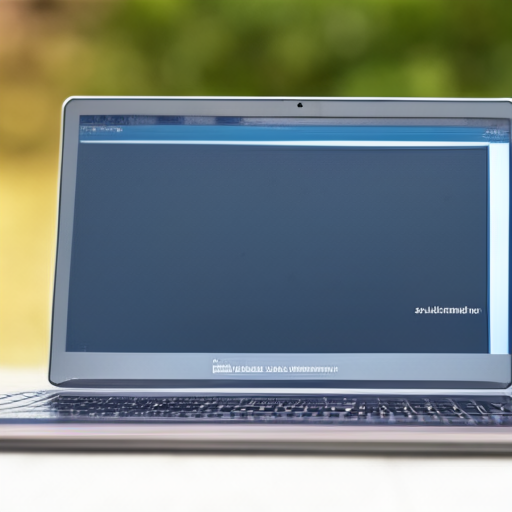}
    \imageWithMoreNote{\xwidth}{\scriptsize \raggedleft \texttt{430M $\alpha=0.8$}}{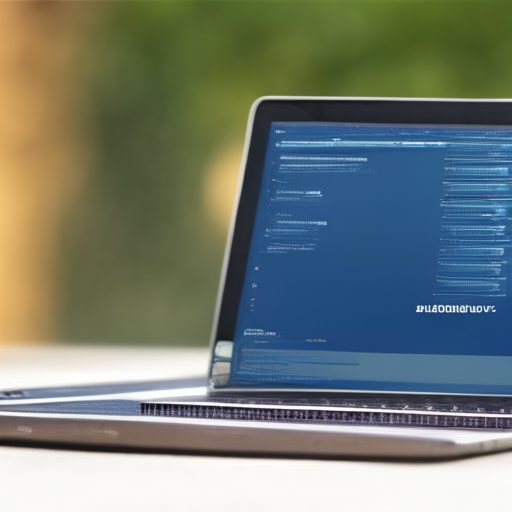}
    \imageWithMoreNote{\xwidth}{\scriptsize \raggedleft \texttt{430M $\alpha=0.5$}}{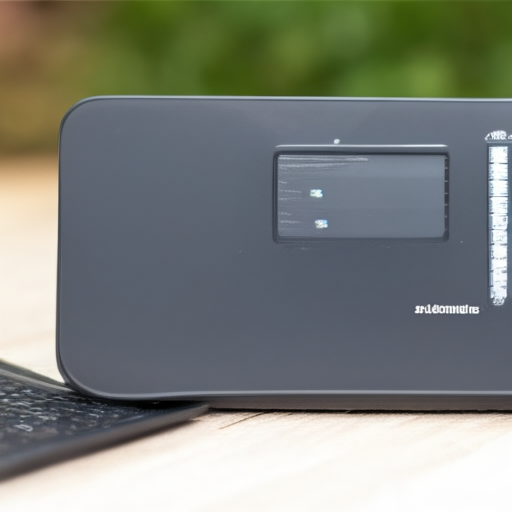}
    \imageWithMoreNote{\xwidth}{\scriptsize \raggedleft \texttt{430M $\alpha=0.2$}}{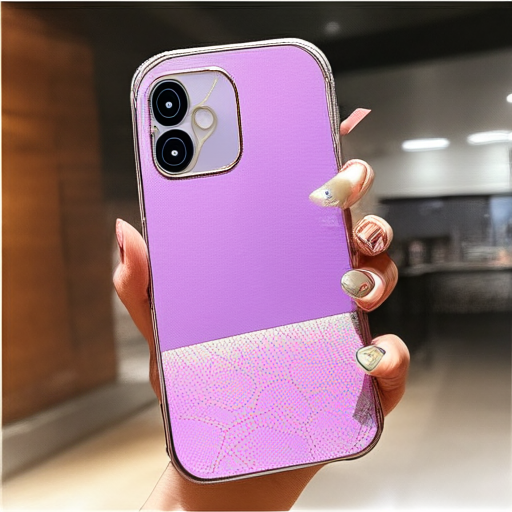}
    \imageWithMoreNote{\xwidth}{\scriptsize \raggedleft \texttt{430M $\alpha=0.0$}}{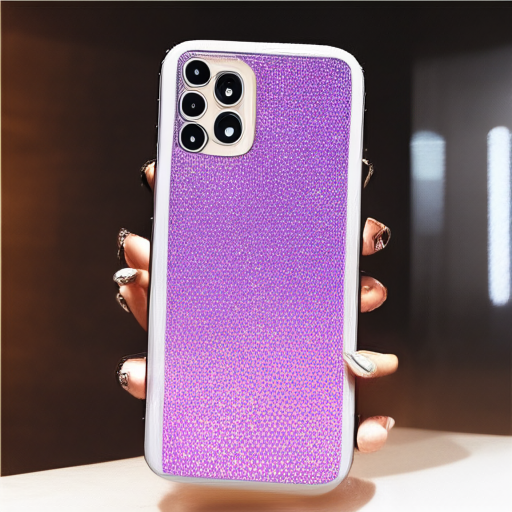}
    \vspace{.5em}
    \hfill
    
    \imageWithMoreNote{\xwidth}{\scriptsize \raggedleft \texttt{866M $\alpha=1.0$}}{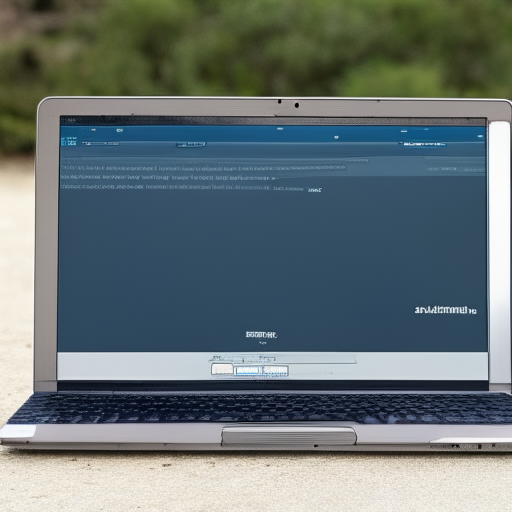}
    \imageWithMoreNote{\xwidth}{\scriptsize \raggedleft \texttt{866M $\alpha=0.8$}}{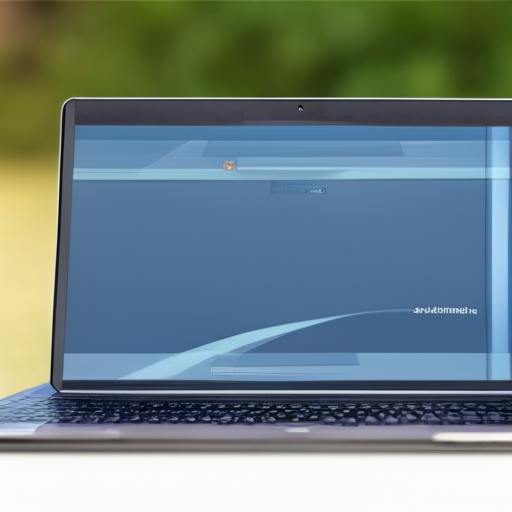}
    \imageWithMoreNote{\xwidth}{\scriptsize \raggedleft \texttt{866M $\alpha=0.5$}}{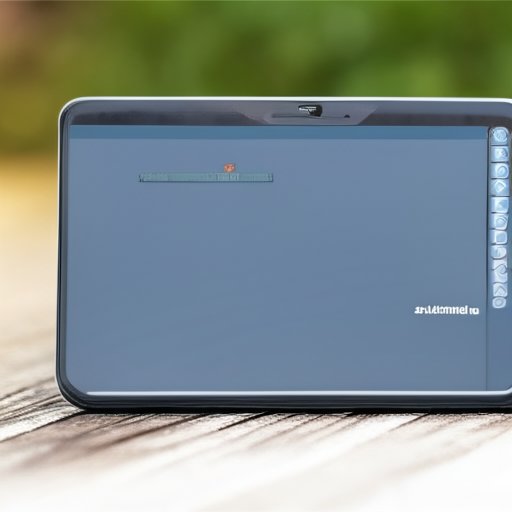}
    \imageWithMoreNote{\xwidth}{\scriptsize \raggedleft \texttt{866M $\alpha=0.2$}}{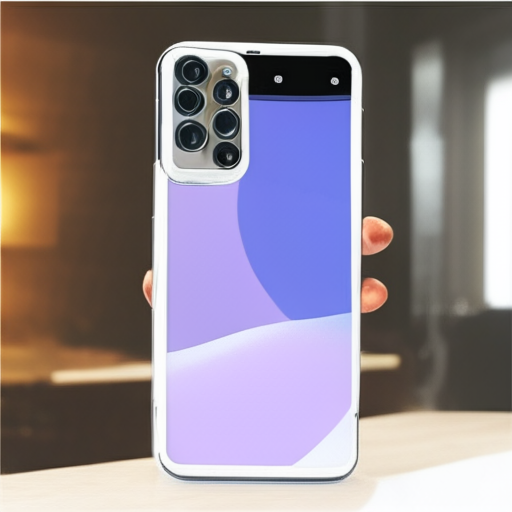}
    \imageWithMoreNote{\xwidth}{\scriptsize \raggedleft \texttt{866M $\alpha=0.0$}}{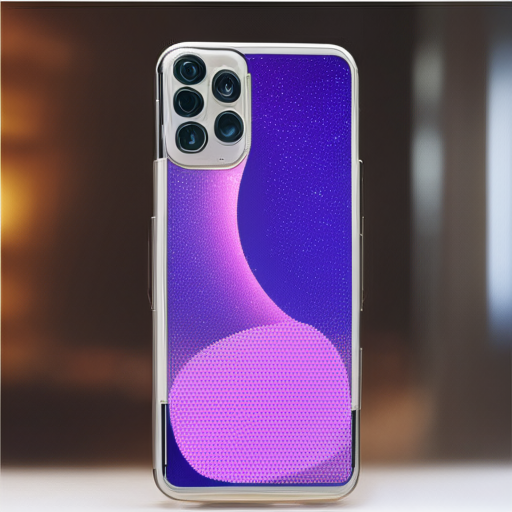}
    \vspace{.5em}
    \hfill

    \imageWithMoreNote{\xwidth}{\scriptsize \raggedleft \texttt{2B $\alpha=1.0$}}{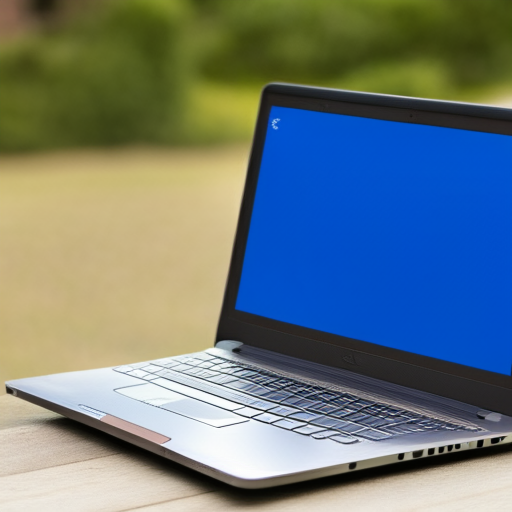}
    \imageWithMoreNote{\xwidth}{\scriptsize \raggedleft \texttt{2B $\alpha=0.8$}}{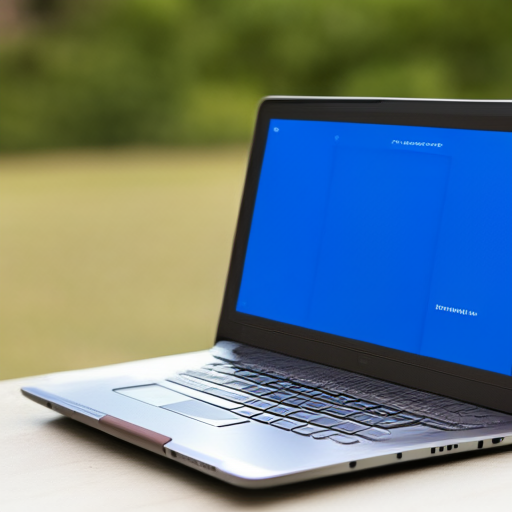}
    \imageWithMoreNote{\xwidth}{\scriptsize \raggedleft \texttt{2B $\alpha=0.5$}}{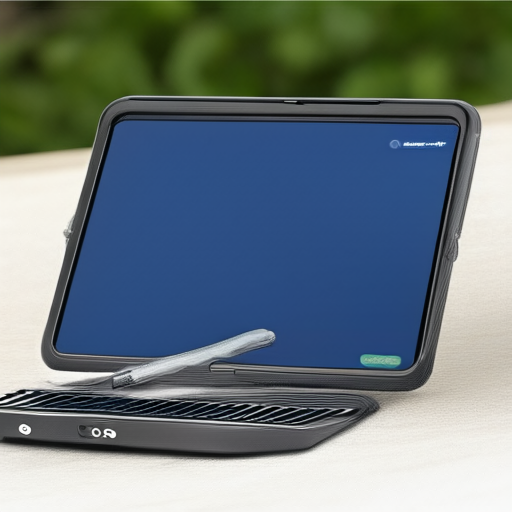}
    \imageWithMoreNote{\xwidth}{\scriptsize \raggedleft \texttt{2B $\alpha=0.2$}}{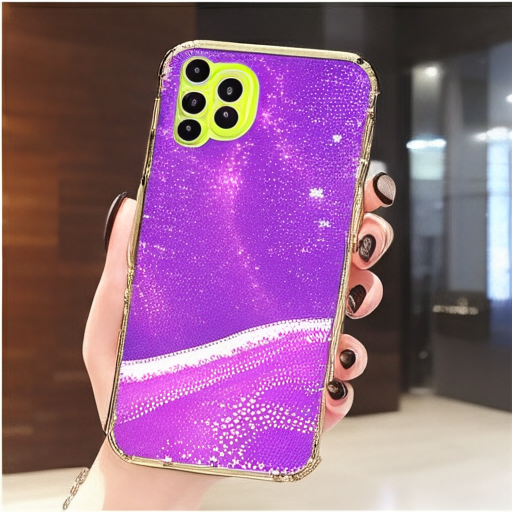}
    \imageWithMoreNote{\xwidth}{\scriptsize \raggedleft \texttt{2B $\alpha=0.0$}}{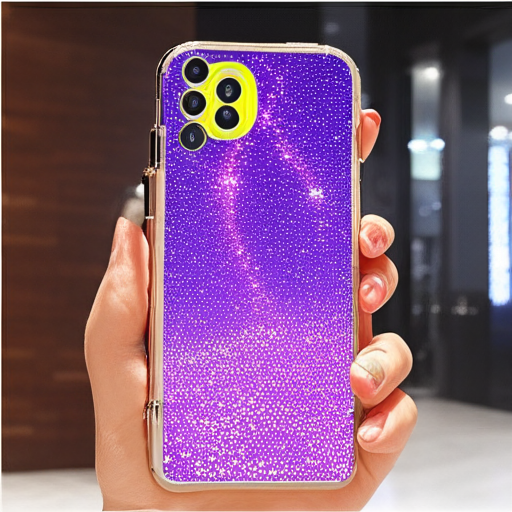}
    \hfill
    
    \vspace{.5em}
    \caption{Prompt $A$: \emph{``a old computer''}. Prompt $B$: \emph{``a fancy mobile phone''}.}
    \end{subfigure}

    \caption{We visualize the text-prompt interpolation results of scaled models in different sizes. Each row shows the results of the same model with different interpolation fraction $\alpha A + (1-\alpha)B$. All results are sampled with the same 20-step DDIM sampler and CFG of 7.5.}
    \label{supp:interp}
    \vspace{-1\baselineskip}
\end{figure}

Text prompt interpolation is widely recognized as a way to evaluate the interpretability of text-to-image models in recent works~\citep{li2024self, park2023understanding}.
In Figure~\ref{supp:interp}, we show the text-prompt interopolation results of models in different sizes and visualize their sampling results.
Specifically, we use two distinct prompts $A$ and $B$ and interpolate their CLIP embeddings as $\alpha A + (1-\alpha) B, \alpha \in [0, 1]$, to generate intermediate text-to-image results. A clear pattern emerges: larger models leads to more semantically coherent and visually plausible interpolations compared to their smaller counterparts. 
The figure demonstrates the 2B model's superior ability to accurately interpret interpolated prompts, as evidenced by its generation of a tablet computer with a touch pen.

